\documentclass[10pt,twocolumn,letterpaper]{article}
\PassOptionsToPackage{table}{xcolor}
\usepackage[pagenumbers]{cvpr}

\usepackage{amsmath,amsfonts,bm}

\def\eqref#1{equation~\ref{#1}}

\def\1{\bm{1}}

\DeclareMathAlphabet{\mathsfit}{\encodingdefault}{\sfdefault}{m}{sl}
\SetMathAlphabet{\mathsfit}{bold}{\encodingdefault}{\sfdefault}{bx}{n}

\usepackage{comment}
\definecolor{ModelBand}{HTML}{E8F2F9}
\definecolor{ModelInk}{HTML}{365B76}
\definecolor{GainInk}{HTML}{406C89}
\definecolor{DropInk}{HTML}{95635E}

\definecolor{MergedBand}{HTML}{E8F2F9}
\definecolor{MergedInk}{HTML}{365B76}

\definecolor{ActionBand}{HTML}{E8F2F9}
\definecolor{ActionInk}{HTML}{365B76}

\definecolor{MemoryBand}{HTML}{E8F2F9}
\definecolor{MemoryInk}{HTML}{365B76}

\definecolor{BenchmarkBand}{HTML}{E8F2F9}
\definecolor{BenchmarkInk}{HTML}{365B76}

\usepackage{longtable}
\usepackage{wrapfig}        %
\usepackage{fix-cm}
\definecolor{citecolor}{HTML}{0071BC}

\usepackage{algorithm}
\usepackage{algpseudocode}
\usepackage{pifont}

\newcommand{\cmark}{\ding{51}}
\newcommand{\xmark}{\ding{55}}
\newcommand{\pmark}{$\circ$}

\definecolor{cvprblue}{rgb}{0.21,0.49,0.74}
\usepackage[pagebackref,breaklinks,colorlinks,allcolors=cvprblue]{hyperref}

\title{MemAct: Recalling the Past and Acquiring the Future for Streaming Spatial Understanding}
\author{%
\parbox{\dimexpr.95\textwidth-2\tabcolsep\relax}{\centering
Yuxi Wei\textsuperscript{1}\quad
Wei Huang\textsuperscript{1}\quad
Qirui Chen\textsuperscript{3}\quad
Lu Hou\textsuperscript{2,*}\quad
Xiaojuan Qi\textsuperscript{1,*}\\[0.6em]
\normalfont\small
\textsuperscript{1}The University of Hong Kong\\
\textsuperscript{2}Yinwang Intelligent Technology Co. Ltd\\
\textsuperscript{3}Shanghai Jiao Tong University\\[0.4em]
\textsuperscript{*}Corresponding authors\\[0.6em]
\href{https://vfishc.github.io/memact/}{Project page} \quad \href{https://github.com/vfishc/memact}{Code}%
}
}
\hypersetup{pdftitle={MemAct: Recalling the Past and Acquiring the Future for Streaming Spatial Understanding}, pdfauthor={Yuxi Wei, Wei Huang, Qirui Chen, Lu Hou, Xiaojuan Qi}}

\begin{document}

\maketitle

\begin{abstract}
Embodied agents perceive the world as a stream and must answer spatial questions while acting. The evidence may lie in the past, in views evicted from the bounded context; in the present view; or in the future, in views reachable only by moving to resolve occlusion, truncation, or scale. Because an agent cannot know in advance which case applies, it must address all three. Existing methods cover only part of it: streaming models lose evicted history and cannot alter viewpoints, whereas active embodied agents typically rely on depth-based maps outside a stream. We introduce MemAct, a streaming vision-language model that unifies memory and action. Operating on RGB alone, it writes memory records during observation, retrieves them with past frames on demand, and takes local camera actions to complete partly visible targets, deciding whether to retrieve, move, or answer. Because camera actions often pay off only in later views and trajectory-level RL such as GRPO assigns credit only through final returns, we train the policy with bandit-based counterfactual optimization, which evaluates every candidate command and an immediate answer using multi-step returns from copies of the same simulator state. We also construct MemAct-Data, with one million training QA turns, and MemAct-Bench, with 18,162 QA turns annotated with evidence locations. On MemAct-Bench, MemAct-9B achieves 70.1\% accuracy and 58.7 mean relative accuracy, surpassing the closed-source GPT-6 Astra and a same-backbone model fine-tuned on the same QA data. Ablations show that retrieval adds 7.9 points on questions requiring evicted frames, while learned camera actions add 10.3 points on questions that need view completion, where bandit-based optimization also outperforms GRPO.
\end{abstract}

\section{Introduction}
\label{sec:intro}

\begin{figure*}[tbp]
\centering
\includegraphics[width=0.8\linewidth]{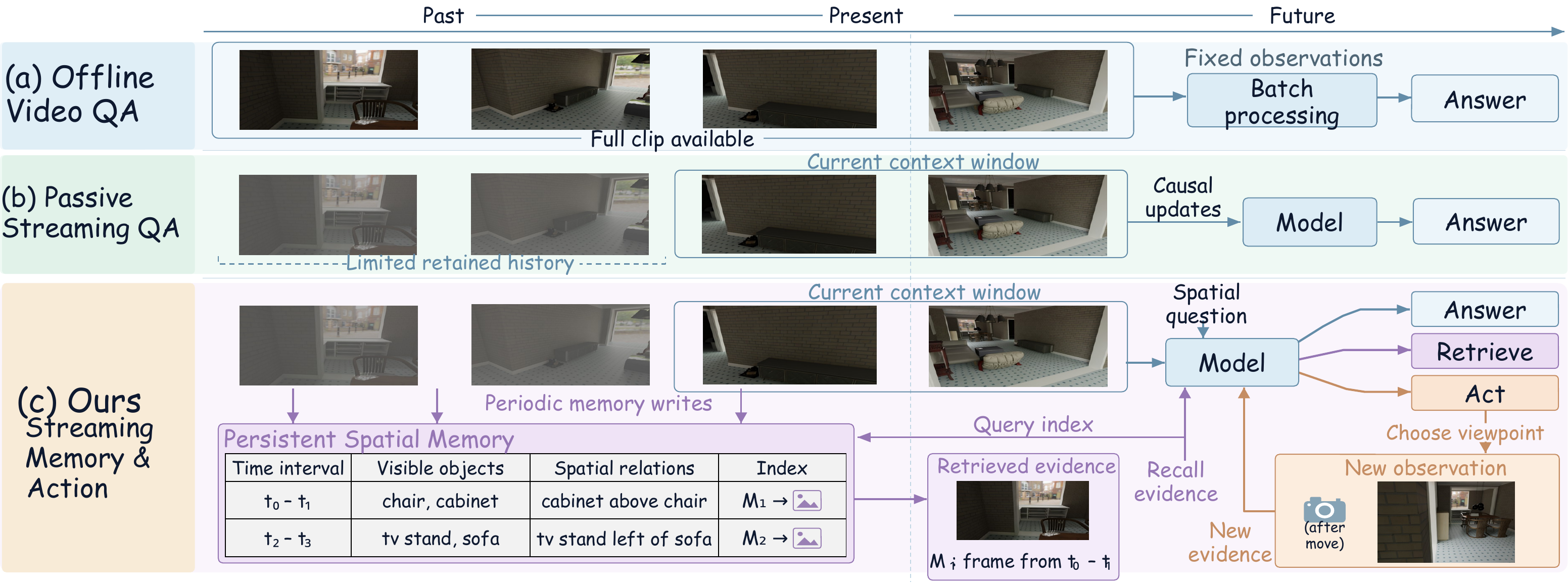}
\vspace{-3mm}
\caption{\small Offline, passive streaming, and MemAct streaming spatial QA. MemAct writes memory while observing; once a question arrives, it answers, retrieves records and past frames, or acquires a better local view, used only for that question before the stream resumes.}
\label{fig:teaser}
\vspace{-5mm}
\end{figure*}

Embodied agents such as household robots must understand space while they move and act, reasoning about objects, their locations and geometry, and their relations~\citep{yang2024thinking,yang2024mem3d,huang2022vlmaps,jatavallabhula2023conceptfusion,gu2023conceptgraphs}. Their observations arrive as a continuous stream~\citep{das2017embodiedqa,gordon2017iqa,yokoyama2023vlfm}, and spatial questions arise during a task rather than after it, so an agent cannot wait for a complete recording and answer offline (Figure~\ref{fig:teaser}a). The evidence a question needs may lie in the past, the present, or the future. When a lamp was last seen depends on earlier views, which may have left the model's bounded context; how far away a visible chair is can be read from the current view; and the size of a cabinet that is occluded, cut off by the image border, or too small to measure becomes answerable only after the agent moves to see it better. A deployed agent meets all three cases in one stream and cannot tell in advance which applies, so one model must recall past evidence, use the present view, and acquire future views as needed.

Existing methods cover only part of this span (Table~\ref{tab:comparison}). Streaming models~\citep{xu2025streamingvlm,yu2026stream3d} process the present as it arrives but neither write nor retrieve memory (Figure~\ref{fig:teaser}b): StreamingVLM evicts old visual tokens and Stream3D-VLM reduces them, so past evidence that has left the context is lost, and because they follow a predetermined camera stream, they cannot obtain a better view. Active embodied QA systems~\citep{yang2024mem3d,zhai2025memoryeqa,ginting2025mindpalace,wang2026lmee} retrieve stored evidence and act, but do not process a stream; they navigate large environments, usually over depth-based 3D maps, and the closest, LMEE~\citep{wang2026lmee}, queries a bank of image features that the model does not write. VG-AVS~\citep{koo2025vgavs} completes the view of a target already in sight, but from a single image, in one step, and without memory. No existing method combines streaming input, memory that the model writes and retrieves, and local view completion, which together span the present, past, and future of a stream (Figure~\ref{fig:teaser}c). Cascading them as separate modules would split coupled decisions: what is recorded during observation constrains what can later be retrieved, and whether to move depends on what the past and present views already provide.

Spanning the past, present, and future in one streaming model raises three challenges. First, past evidence must be kept before the question is known: the active context is bounded, so the model must decide what to record while the stream unfolds. Second, once queried, the model must tell from RGB alone whether the evidence is in the present view, recoverable from the past, or obtainable only from a future view. Third, acquiring future views requires learning when to move and when to stop, although a camera action often pays off only later: turning toward a cabinet may not reveal it but makes the next forward step informative. Trajectory-level RL such as GRPO~\citep{shao2024deepseekmath}, which LMEE uses, gives every decision in a trajectory the same advantage, so credit assignment is hard: such preparatory actions and the decision to stop are credited only through the final return.

We introduce \textbf{MemAct}, a streaming VLM that spans all three in one model and addresses each challenge (Section~\ref{sec:method}). First, MemAct writes timestamped, frame-linked records every five seconds; text stays in the active context, while the oldest frames are evicted but remain retrievable. Second, once queried, it chooses the response that matches where the evidence lies: it answers from its context, requests records or past frames, or issues up to three camera commands to better view a target already in sight, a task we call \emph{local view completion}; a tool runtime without learned parameters executes these requests, and no depth or 3D maps are used. Third, after supervised fine-tuning, we address credit assignment with \emph{bandit-based counterfactual optimization}: at each decision state, every camera command is executed in its own simulator copy, valued by multi-step continuations, and compared with answering immediately, crediting preparatory actions and stopping by estimated downstream returns.

Learning these decisions needs memory and action supervision, and evaluating them needs streaming questions that may require moving; no existing benchmark combines streaming input with actions (Table~\ref{tab:comparison}, top). We therefore build MemAct-Data, one million training QA turns with this supervision from real and simulated streams, and MemAct-Bench, 18,162 answer-balanced QA turns whose Current, Memory, and Action questions target the present, the past, and views that may require moving, each annotated with its evidence location to support stage-wise error attribution (Section~\ref{sec:dataset}).

On MemAct-Bench, MemAct-9B achieves the best overall accuracy and mean relative accuracy, ahead of the closed-source GPT-6 Astra and a same-backbone model fine-tuned on the same QA data. Evidence-location annotations show that its margin over the latter grows where evidence has left the context or must be acquired by moving: on Memory questions it widens with evidence age and is 5.1 points once evidence is evicted, and on Action questions that need view completion it is 9.4. From identical pre-question states, retrieval adds 7.9 points on questions whose needed frames have been evicted. Enabling the camera adds 10.3 points on questions that need view completion, where bandit-based optimization exceeds GRPO by 4.9 under the same answerer and stays ahead with a frozen SFT answerer.

Our contributions are threefold. (1)~\textbf{Model:} MemAct, a streaming VLM that, under a bounded active context, spans the past, present, and future of a stream: it recovers past evidence through memory it writes and retrieves itself and acquires future views through local camera actions, from RGB alone. (2)~\textbf{Training:} bandit-based counterfactual optimization, which addresses credit assignment for preparatory camera actions and stopping by comparing them with all alternatives at the same state through multi-step returns, and outperforms GRPO, especially on questions that need view completion. (3)~\textbf{Data and benchmark:} MemAct-Data and MemAct-Bench, which supervise memory writing, retrieval, and multi-step view acquisition, and evidence-location annotations that show where each ability matters and attribute Memory and Action failures to stages.

\section{Related Work}
\label{sec:related}

\begin{table}[t]
\centering
\caption{\small Related benchmarks (top) and methods (bottom); \cmark{} yes, \pmark{} partial, \xmark{} no. \emph{Stream}: questions see only earlier frames. \emph{Past evidence}: includes questions whose evidence has left the current view. Nav: navigation; Local: completing the view of a target in sight. \emph{Write}: the model writes its own records.}
\label{tab:comparison}
\vspace{-2mm}
\begingroup
\scriptsize
\setlength{\tabcolsep}{2pt}
\renewcommand{\arraystretch}{0.9}
\begin{tabular*}{\linewidth}{@{\extracolsep{\fill}}lcccc@{}}
\toprule
 & & Past & & Active \\
Benchmark & Stream & evidence & Scenes & exploration \\
\midrule
VSI-Bench~\citep{yang2024thinking} & \xmark & \pmark & real video & -- \\
StreamEQA~\citep{wang2025streameqa} & \cmark & \cmark & real video & -- \\
EXPRESS-Bench~\citep{Jiang2025iccv} & \xmark & \xmark & real scans & Nav \\
LMEE-Bench~\citep{wang2026lmee} & \xmark & \cmark & real scans & Nav \\
ESI-Bench~\citep{hong2026esi} & \xmark & \pmark & synthetic & Local \\
AVS-HM3D~\citep{koo2025vgavs} & \xmark & \xmark & real scans & Local \\
MemAct-Bench (ours) & \cmark & \cmark & real + synth. & Local \\
\midrule
 & & \multicolumn{2}{c}{Memory} & Active \\
\cmidrule(lr){3-4}
Method & Stream & Write & Retrieve & exploration \\
\midrule
StreamingVLM~\citep{xu2025streamingvlm} & \cmark & \xmark & \xmark & -- \\
Stream3D-VLM~\citep{yu2026stream3d} & \cmark & \xmark & \xmark & -- \\
Cambrian-S~\citep{yang2025cambrians} & \pmark & \pmark & \cmark & -- \\
3D-Mem~\citep{yang2024mem3d} & \xmark & \xmark & \cmark & Nav \\
MemoryEQA~\citep{zhai2025memoryeqa} & \xmark & \pmark & \pmark & Nav \\
Mind Palace~\citep{ginting2025mindpalace} & \xmark & \cmark & \cmark & Nav \\
LMEE~\citep{wang2026lmee} & \xmark & \xmark & \cmark & Nav \\
VG-AVS~\citep{koo2025vgavs} & \xmark & \xmark & \xmark & Local \\
MemAct (ours) & \cmark & \cmark & \cmark & Local \\
\bottomrule
\end{tabular*}
\endgroup
\vspace{-4mm}
\end{table}

\paragraph{Spatial and streaming understanding.}
Spatial benchmarks~\citep{azuma2021scanqa,ma2022sqa3d,fu2024blink,yang2024thinking} and spatially grounded training data~\citep{chen2024spatialvlm,cheng2024spatialrgpt,song2024robospatial,ray2024sat,yang2025cambrians} strengthen reasoning over supplied visual evidence; Cambrian-S~\citep{yang2025cambrians} also builds VSI-590K and surprise-driven memory. Streaming methods process observations as they arrive~\citep{zhang2024flashvstream,yao2025timechatonline,he2024malmm,lin2024streamingbench,huang2024ovbench,chen2025livecc,xu2025streamingvlm,yu2026stream3d}, and StreamEQA~\citep{wang2025streameqa} evaluates reasoning on recorded videos.

\paragraph{Active embodied QA: exploration versus local view completion.}
\emph{Long-horizon exploration} asks an agent to find information in a large environment: A-EQA~\citep{majumdar2024openeqa} and EXPRESS-Bench~\citep{Jiang2025iccv} evaluate answers together with exploration, and systems~\citep{ren2024exploreconfident,yang2024mem3d,zhai2025memoryeqa,ginting2025mindpalace} plan over frontiers, snapshot memories, or scene graphs, usually built from depth; LMEE~\citep{wang2026lmee} trains an MLLM with GRPO to navigate over a depth-based voxel map and queries a bank of CLIP features once per question. \emph{Local view completion} instead starts from a target already in view: VG-AVS~\citep{koo2025vgavs} predicts a single continuous camera adjustment without memory and introduces AVS-HM3D on real HM3D~\citep{ramakrishnan2021hm3d} scans, and ESI-Bench~\citep{hong2026esi} evaluates the perception--action loop in simulated homes with camera, locomotion, and manipulation actions. MemAct belongs to the second kind but runs on a stream with memory, so we evaluate transfer on AVS-HM3D and ESI-Bench rather than on EXPRESS-Bench, which requires long-horizon navigation.

\paragraph{Credit assignment in RL.}
PPO~\citep{schulman2017ppo} learns a value network for per-step credit, whereas GRPO~\citep{shao2024deepseekmath} normalizes trajectory returns within a group, so every decision in a trajectory shares one advantage and stopping is credited only through the return. VinePPO~\citep{kazemnejad2025vineppo} values sampled intermediate states by Monte Carlo restarts, following TRPO's vine procedure~\citep{schulman2015trpo}. Mean actor-critic and expected policy gradients~\citep{allen2017mac,ciosek2018epg} weight every action's gradient by its estimated value, rollout-based policy improvement~\citep{tesauro1996rollout} estimates action values by simulation, and contextual bandits~\citep{langford2007epochgreedy,li2010linucb} choose one arm per context. Our bandit-based counterfactual optimization (Section~\ref{sec:training_strategy}) combines these ideas to compare all candidates, including an explicit stopping candidate, at the same state.

\section{MemAct-Bench and MemAct-Data}
\label{sec:dataset}

\begin{figure*}[tbp]
\centering
\includegraphics[width=0.85\linewidth]{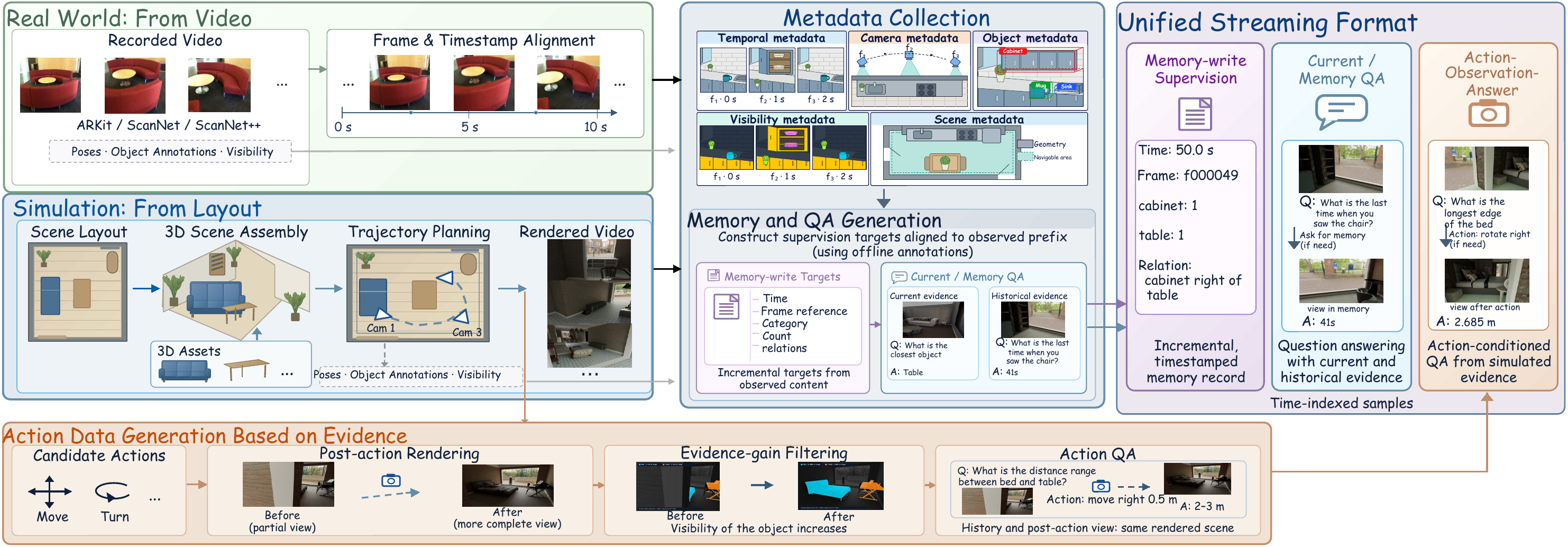}
\vspace{-3mm}
\caption{\small Construction of MemAct-Data and MemAct-Bench. Real and simulated streams are aligned with temporal, camera, object, visibility, and scene metadata, which computes answers, memory and retrieval supervision, and camera-action demonstrations selected by evidence gain, but never enters model inputs.}
\label{fig:benchmark_construction}
\vspace{-5mm}
\end{figure*}

One pipeline produces both MemAct-Data and MemAct-Bench (Figure~\ref{fig:benchmark_construction}). Per-frame metadata computes answers, supervision, and rewards but never enters the model input of RGB frames, timestamps, questions, and tool results.

\subsection{Question Families}
\label{sec:taxonomy}

\textbf{Current QA} uses the current observation or a recent window (five seconds by default): distances, camera motion, visibility changes, relations, heights, and room area. \textbf{Memory QA} asks about evidence before the current window: first and last appearances, appearance order, distinct-instance counts, and historical co-appearance, relations, and nearest neighbors. It is defined by \emph{where its evidence lies}: when the evidence is still in context, the model may answer directly, so we annotate whether it has been evicted (Section~\ref{sec:test_construction}). A Memory template is \emph{record-answerable} (R) if the memory records alone determine the answer, or \emph{frame-required} (F) if it needs metric geometry or instance identity that the records omit. \textbf{Action QA} asks for an object's size or the distance between two objects whose initial view is incomplete; the distance-range template phrases each pair twice, so its 142 questions cover 71 pairs. The families have 18, 10 (7 R, 3 F), and 3 templates and 9,147, 7,964, and 1,051 test questions (Supp.~\ref{app:catalog}; examples: Supp.\ Fig.~\ref{fig:trace_samples}, \ref{app:visualization}).

\subsection{Data Construction and Test Set}
\label{sec:observation_construction}
\label{sec:evidence_alignment}

\paragraph{Sources and alignment.}
Simulated streams follow randomized room visits in 3D-FRONT~\citep{fu2021front} layouts, and real streams use ScanNet~\citep{dai2017scannet}, ScanNet++~\citep{yeshwanth2023scannetpp}, and ARKitScenes~\citep{baruch2021arkitscenes} videos; all are sampled at 1~FPS as $640\times480$ RGB frames and aligned with per-frame metadata (Supp.~\ref{app:alignment}). Independent streams last at most 180~s (Section~\ref{sec:method_loop}), and a question at time $t$ sees only the causal prefix of its stream.

\paragraph{Memory and retrieval supervision.}
\label{sec:memory_qa_construction}
For each five-second window, the memory target has one record per frame with the counts of visible categories and optional image-plane relations on a $3\times3$ grid (Supp.~\ref{app:memory_targets}; examples in Supp.\ Figs.~\ref{fig:memory_write_sample_1}--\ref{fig:memory_write_sample_2}). Records hold no template answers, so distinct-instance counts and historical distances need retrieved frames, and these frame-required templates make up 43.0\% of Memory test questions. Retrieval traces issue \texttt{use\_memory} requests whose results, under the Memory Store's rules (Section~\ref{sec:method_memory}), contain the evidence within budget, and answer directly if it is in context (Supp.~\ref{app:spatial_qa}--\ref{app:question_times}).

\paragraph{Action data.}
\label{sec:action_construction}
In simulated streams, Action QA binds partly visible (visible mask $\geq0.2\%$ of the image), uniquely referable targets that are occluded, truncated, or too small, and asks for a quantity fixed by geometry: an object's longest bounding-box side or the distance between two objects' 3D centers, as a number or a range. It is thus \emph{local view completion} rather than a search for unseen objects. Five commands move forward, left, or right by $0.5$~m or turn left or right by $45^\circ$, with at most three attempts (Supp.~\ref{app:action_data}). A view $x$ has quality $E(x,q)\in[0,1]$, the minimum over targets of the visible-pixel fraction, normalized at 10\% of the image, times framing completeness (Supp.~\ref{app:action_optimization}); it is \emph{adequate} if $E(x,q)\geq0.2$, and $\Phi_{\mathrm{pre}}(q)=\max_{x\in O_{\mathrm{pre}}}E(x,q)$ is the best quality over the observation history $O_{\mathrm{pre}}$ before the question. If $\Phi_{\mathrm{pre}}(q)\leq0.05$, we render up to three commands and keep the earliest adequate endpoint that clearly improves on $\Phi_{\mathrm{pre}}(q)$. Intermediate views need not be adequate, so demonstrations include preparatory actions (worked example in Supp.~\ref{app:action_data}).

\paragraph{Test split and strata.}
\label{sec:test_construction}
\label{sec:assembly_qc}
Training and test sets are disjoint in physical spaces, and answers are balanced within each template by a rule independent of model outputs (Supp.~\ref{app:test_composition}). Each Memory question carries a reference label, fixed under the eviction rule of Section~\ref{sec:method_loop}, that records whether all of its required evidence frames are still in context; 62.0\% are labeled evicted. An Action question \emph{needs no move} if its initial view $x_0$ is adequate, $E(x_0,q)\geq0.2$, and \emph{needs completion} if $\Phi_{\mathrm{pre}}(q)\leq0.05$ but some view reachable within three steps is adequate, split into \emph{one-step} and \emph{multi-step} by the fewest steps to an adequate view; others are excluded. Of the 1,051 Action questions, 769 need completion and 282 need no move. All annotations were fixed before any model was evaluated (composition: Supp.~\ref{app:distribution}, \ref{app:test_composition}).

\section{MemAct}
\label{sec:method}

\begin{figure*}[tbp]
\centering
\includegraphics[width=0.8\linewidth]{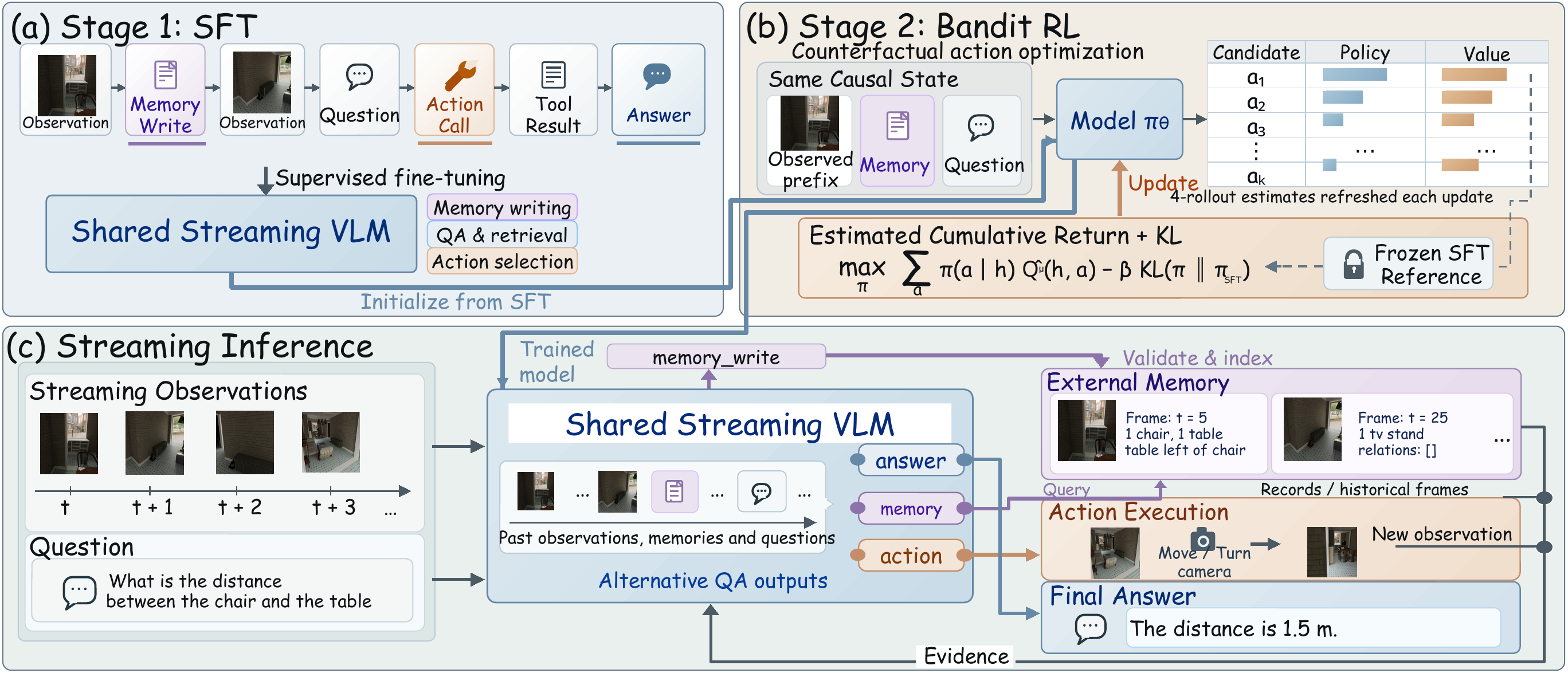}
\vspace{-3mm}
\caption{\small MemAct inference and training. (a) SFT teaches one VLM a shared interface. (b) Bandit-based counterfactual optimization compares the five camera commands and \texttt{answer\_now} from copies of the same state by multi-step returns, acting only on camera and stopping decisions. (c) While streaming, the VLM writes memory every five seconds; once a question arrives, it answers, issues \texttt{use\_memory} requests to the Memory Store, or sends commands to the Camera Executor.}
\label{fig:method}
\vspace{-5mm}
\end{figure*}

MemAct consists of a policy $\pi_\theta$, implemented as a single VLM, and a deterministic \emph{tool runtime} with no learned parameters (Figure~\ref{fig:method}); what to record, request, and look at, and when to stop, are decided by $\pi_\theta$.

\subsection{Problem Formulation}
\label{sec:method_overview}

Observations $o_1,o_2,\dots$ of a stream arrive at 1~FPS, and a question $q$ may arrive at any time $t_q$. The policy outputs (i) memory writes $\Delta M$ during observation and, while answering, (ii) \texttt{use\_memory} requests, (iii) camera commands, or (iv) an answer $\hat y$, with at most four retrieval calls, four historical frames, and three camera attempts per question (Supp.\ Tab.~\ref{tab:budgets}). It never sees future frames, ground-truth geometry, or rewards. The runtime comprises a \emph{Memory Store}, which validates, stores, and queries records and keeps all past frames, and a \emph{Camera Executor}, which executes commands and returns the views. We index stream time by $t$, rounds within a question by $k$, and camera attempts by $j$; $C_t$ is the active context, $M$ the stored records, and $h_k$ the history at round $k$, with request $u_k$ and result $z_k$.

\subsection{Inference Loop}
\label{sec:method_loop}

\paragraph{Streams, context, and memory writing.}
Each independent stream (Section~\ref{sec:observation_construction}) initializes its own context and external memory. $C_t$ is kept in a persistent KV cache, and its budget of $L=16{,}384$ tokens applies to the complete active context, including textual history, visual tokens, and question-time tool interactions. Text, namely the system prompt, questions and answers, memory records, and tool-interaction text, is retained within each stream, while the visual tokens of the oldest frames are evicted, eight frames at a time, as needed to maintain the total budget (Supp.~\ref{app:streaming_context}); evicted frames remain retrievable from the Memory Store. Every five seconds, $\pi_\theta$ writes $\Delta M$, one record per frame of the elapsed window, which enters the Memory Store through $M\leftarrow\operatorname{Update}(M,\Delta M)$ and stays in $C_t$.

\paragraph{Answering.}
When $q$ arrives, the stream pauses and frames arriving meanwhile are buffered. At round $k$, $\pi_\theta$ receives $h_k=(C_{t_q},q,u_0,z_0,\dots,u_{k-1},z_{k-1},b_k)$, managed under the same budget $L$, where $b_k$ holds the remaining per-question budgets, and outputs $u_k$; the runtime returns $z_k$, and $h_{k+1}=\operatorname{Append}(h_k,u_k,z_k)$. Successive commands continue from the state reached, and exhausting the retrieval-call or camera-attempt budget triggers a request for the final answer. After the answer, retrieved frames and acquired views leave the context and are never written to the Memory Store, the question, tool-interaction text, and answer remain as retained text, and the recorded stream resumes (Supp.\ Alg.~\ref{alg:inference}). Views are rendered in simulation.

\subsection{Tool Runtime}
\label{sec:method_memory}
\label{sec:method_action}

$\operatorname{Update}$ appends a valid write with a frame and time index and rejects an invalid one, which $\pi_\theta$ may retry (Supp.~\ref{app:router}). A \texttt{use\_memory} request specifies categories, an optional time range, and whether frames are needed; the Memory Store returns matching records in time order, with at most two representative frames per call. The Camera Executor translates by $0.5$~m in the camera's horizontal frame or rotates by $45^\circ$; a command that fails collision checks is invalid but consumes an attempt (Supp.~\ref{app:evaluation}).

\begin{table*}[!t]
\centering
\caption{\small MemAct-Bench results (ACC / MRA). Data: fine-tuned on our QA turns (QA) or on QA, memory, and action supervision (Full), both with VSI-590K. Action has 142 multiple-choice and 909 numerical questions. Backbones, latency, and reference baselines (uniform, per-template modal answer, training median, text-only) are in Supp.\ Tab.~\ref{tab:main_full}.}
\label{tab:main}
\vspace{-3mm}
\begingroup
\scriptsize
\setlength{\tabcolsep}{4pt}
\renewcommand{\arraystretch}{0.9}
\begin{tabular*}{\textwidth}{@{\extracolsep{\fill}}lcrrrrrrrr@{}}
\toprule
 & & \multicolumn{2}{c}{Current} & \multicolumn{2}{c}{Memory} & \multicolumn{2}{c}{Action} & \multicolumn{2}{c}{Overall} \\
\cmidrule(lr){3-4}\cmidrule(lr){5-6}\cmidrule(lr){7-8}\cmidrule(l){9-10}
Model & Data & ACC & MRA & ACC & MRA & ACC & MRA & ACC & MRA \\
\midrule
GPT-6 Astra & \xmark & \textbf{68.07} & 50.36 & 68.48 & 61.50 & 49.30 & 46.53 & 67.98 & 55.90 \\
Gemini-3.8-Flash & \xmark & 58.48 & 40.16 & 60.96 & 47.95 & 39.44 & 36.62 & 59.17 & 43.92 \\
Qwen3.5-4B & \xmark & 40.61 & 17.65 & 54.66 & 34.11 & 29.58 & 15.09 & 45.75 & 26.25 \\
Qwen3.5-9B & \xmark & 44.45 & 21.83 & 60.49 & 41.41 & 34.51 & 18.21 & 50.35 & 31.98 \\
Qwen3.8-27B & \xmark & 46.32 & 23.29 & 62.74 & 44.55 & 38.03 & 21.07 & 52.39 & 34.54 \\
InternVL3.5-8B & \xmark & 39.65 & 18.11 & 56.05 & 35.69 & 31.69 & 15.48 & 45.71 & 27.30 \\
InternVL3.5-38B & \xmark & 42.24 & 20.31 & 57.58 & 39.06 & 35.92 & 17.35 & 47.92 & 30.10 \\
VLM-3R & \xmark & 44.65 & 23.61 & 63.40 & 44.33 & 35.21 & 19.78 & 51.57 & 34.35 \\
Cambrian-S & \xmark & 44.05 & 22.57 & 61.22 & 43.51 & 35.21 & 19.77 & 50.38 & 33.57 \\
Cambrian-S-FT & QA & 66.36 & 37.31 & 67.35 & 52.08 & 46.48 & 35.67 & 66.48 & 45.11 \\
StreamingVLM (Qwen2.5-VL-7B) & QA & 62.27 & 35.79 & 61.99 & 50.50 & 30.99 & 30.23 & 61.77 & 43.04 \\
Stream3D-VLM & QA & 59.76 & 32.46 & 61.29 & 46.73 & 27.46 & 27.35 & 59.93 & 39.53 \\
\midrule
Same-backbone control (Qwen3.5-9B) & QA & 66.68 & 49.61 & 71.04 & 59.94 & 42.49 & 45.13 & 68.01 & 54.63 \\
Retrieval baseline (Qwen3.5-9B) & QA & 66.67 & 48.91 & 72.21 & 60.17 & 40.85 & 43.37 & 68.43 & 54.29 \\
\midrule
MemAct-4B & Full & 61.06 & 36.06 & 65.19 & 55.98 & 50.70 & 49.08 & 62.48 & 48.61 \\
MemAct (Qwen2.5-VL-7B) & Full & 65.50 & 40.45 & 67.35 & 56.88 & 51.41 & 50.13 & 66.02 & 50.66 \\
MemAct-9B & Full & 68.03 & \textbf{52.33} & \textbf{74.02} & \textbf{64.02} & \textbf{52.35} & \textbf{52.63} & \textbf{70.08} & \textbf{58.72} \\
\bottomrule
\end{tabular*}
\endgroup
\vspace{-3mm}
\end{table*}

\subsection{Training}
\label{sec:training_strategy}
\label{sec:method_sft}
\label{sec:method_rl}

\paragraph{Stage 1: Supervised fine-tuning.}
SFT trains the shared interface on causally ordered MemAct-Data traces, with memory writes, retrievals, direct answers, and camera actions, plus VSI-590K~\citep{yang2025cambrians}. The loss covers assistant tokens only, and each output is trained on its own causal prefix under the inference-time context-management rule (Supp.~\ref{app:training}).

\paragraph{Stage 2: Bandit-based counterfactual optimization.}
A camera action is often useful only through the views it enables later (Section~\ref{sec:intro}). An immediate evidence-gain reward does not capture this later value, and SFT imitates one path per state without alternatives. At a decision state $h$, Stage 2 therefore (1) executes each of the five commands and \texttt{answer\_now} in its own copy of the simulator state; (2) continues each command branch $B=4$ times with a frozen snapshot $\mu$ of the current policy and averages returns, while \texttt{answer\_now} is scored by one greedy answer; and (3) shifts $\pi_\theta$ toward higher-valued candidates. Each state is thus a \emph{contextual-bandit} round~\citep{langford2007epochgreedy} whose six arms are compared at the same state by cumulative multi-step Monte Carlo returns. We call this objective \emph{Bandit}.

\paragraph{Rewards.}
With the view quality $E$ of Section~\ref{sec:action_construction}, $\Phi(h,q)=\max_{x\in\mathcal O(h)}E(x,q)$ is the best quality over the pre-question observations and earlier action results. Writing $\Phi_j$ for $\Phi$ of the history before camera attempt $a_j$, the attempt receives
\begin{equation}
r_j=\alpha(\Phi_{j+1}-\Phi_j)-\lambda_a-\lambda_{\mathrm{inv}}\mathbf 1[\mathrm{invalid}(a_j)],
\end{equation}
with $\alpha=0.10$, $\lambda_a=0.03$, and $\lambda_{\mathrm{inv}}=0.20$, so revisiting evidence already seen earns nothing, and the final answer receives $R_{\mathrm{QA}}$, option correctness or MRA. A continuation that answers after $T$ attempts gives decision $j$ the return $G_j=\sum_{\ell=j}^{T-1}r_\ell+R_{\mathrm{QA}}$. $E$ and $\Phi$ are computed from simulator masks that the model never receives.

\paragraph{Update.}
The value of command $a$ is $\widehat Q^\mu(h,a)=\frac1B\sum_{b=1}^{B}G^{(b)}(h,a)$, and that of \texttt{answer\_now} is $R_{\mathrm{QA}}$ of one greedy answer. With $\pi_\theta(a\mid h)$ obtained by normalizing the summed log probabilities of each candidate's decision tokens over the candidates $\mathcal A(h)$, we maximize
\begin{equation}
\mathbb E_h\Bigl[\sum_{a\in\mathcal A(h)}\pi_\theta(a\mid h)\widehat Q^\mu(h,a)-\beta D_{\mathrm{KL}}(\pi_\theta\Vert\pi_{\mathrm{SFT}})\Bigr],
\end{equation}
with $\beta=0.05$ and a frozen SFT reference. The objective acts only on camera and stopping decisions; writing, retrieval, and answering are trained by interleaved supervised replay (Supp.~\ref{app:action_optimization}), and retrieval is not optimized with RL because its correct target is fixed by the evidence annotation. Simulator copies are needed only for training.

\section{Experiments}
\label{sec:experiments}

\subsection{Experimental Setup}
\label{sec:experimental_setup}

\paragraph{Models and controls.}
MemAct-4B and MemAct-9B fine-tune Qwen3.5-4B and Qwen3.5-9B~\citep{qwen2026qwen35} with LoRA~\citep{hu2021lora} on MemAct-Data and VSI-590K~\citep{yang2025cambrians}; to separate method from backbone, we also train MemAct on Qwen2.5-VL-7B~\citep{bai2025qwen25vl}, the backbone of StreamingVLM. Baselines are closed-source (GPT-6 Astra~\citep{openai2026astra}, Gemini-3.8-Flash~\citep{deepmind2026gemini38}), open-source (Qwen3.5, Qwen3.8~\citep{qwen2026qwen38}, InternVL3.5~\citep{wang2025internvl35}), spatial (VLM-3R~\citep{fan2025vlm3r}, Cambrian-S), and streaming models (StreamingVLM~\citep{xu2025streamingvlm}, Stream3D-VLM~\citep{yu2026stream3d}); the streaming models and Cambrian-S-FT are fine-tuned on our QA data and VSI-590K (Supp.~\ref{app:baselines}). They use the same tool protocol at test time but receive no memory or action supervision, which is specific to MemAct's tool runtime. The \emph{same-backbone control} fine-tunes Qwen3.5-9B with the same configuration, questions, and steps as MemAct-9B, including the QA steps of Stage 2 replay, but without memory and action supervision, so it writes no records and does not retrieve; comparisons with it reflect MemAct's complete training and system scheme. The \emph{retrieval baseline} is trained identically but receives, at each question, the four past frames most similar to it under SigLIP2-so400m~\citep{tschannen2025siglip2}. \emph{Oracle retrieval} is a diagnostic in which MemAct-9B receives the annotated evidence frames.

\paragraph{Protocol and metrics.}
Offline models receive 32 uniformly sampled prefix frames, streaming models observe each independent stream at 1~FPS with $L=16{,}384$, and all models can use the camera through the same text protocol; on real streams, a camera command returns the original frame. We report accuracy (ACC) and mean relative accuracy (MRA) as in VSI-Bench~\citep{yang2024thinking}, and \emph{Score}, the mean of option correctness or MRA, for groups that mix both. MemAct-9B, the same-backbone control, and every policy in Table~\ref{tab:action-ablation} are trained with three seeds, and tables report means; standard deviations and paired 95\% cluster-bootstrap confidence intervals are in Supp.~\ref{app:full_results}. To evaluate views independently of the reward, the \emph{full-visibility rate} is the share of Action questions whose targets are all found in the final view by GroundingDINO~\citep{liu2024groundingdino} with SAM2~\citep{ravi2025sam2}, with masks matching the ground truth and boxes clear of the image border (settings in Supp.~\ref{app:action_optimization}). It agrees with a geometric ground-truth rule on 92.7\% of 300 calibration views and measures complete, detectable views rather than scale. The Action strata and failure attribution use thresholds on $\Phi$ and only locate where accuracy changes. Supp.~\ref{app:timing} separately evaluates when to answer, without tools.

\subsection{Main Results}
\label{sec:exp_main}

\paragraph{Overall comparison.}
MemAct-9B achieves the best Overall ACC (70.08) and MRA (58.72) in Table~\ref{tab:main} and leads in Memory and Action. Among the baselines other than the same-backbone control and the retrieval baseline, the closed-source GPT-6 Astra is the strongest in every family; on Current questions, which need neither memory nor actions, MemAct-9B is on par with it in ACC (68.03 vs.\ 68.07) and ahead in MRA (52.33 vs.\ 50.36), and overall it exceeds GPT-6 Astra by 2.10 ACC (95\% CI [1.26, 2.92]) and 2.82 MRA ([2.29, 3.35]). Compared with StreamingVLM, fine-tuned on the same QA data and processing the same stream, it improves Memory ACC by 12.03 points and Action Score by 22.26 points, and MemAct on StreamingVLM's own backbone improves over it by 4.25 Overall ACC and 7.62 MRA, so the gains are not specific to the Qwen3.5 backbone. Reference baselines stay near chance, so the questions require visual evidence (Supp.\ Tab.~\ref{tab:main_full}).

\paragraph{Where the gains come from.}
The margin over the same-backbone control grows where memory or actions are needed. MemAct-9B exceeds it by 2.07/4.09 Overall ACC/MRA (95\% CIs [1.54, 2.61] and [3.71, 4.49]), by 1.35/2.72 on Current, which needs neither, and by 2.98/4.08 on Memory and 9.86/7.49 on Action. Within these two families, the margin is largest where the evidence is out of reach without them: 5.05 Score points on Memory questions with evicted evidence versus 0.94 when it is still in context, where both models can read it (Table~\ref{tab:memory-grouped}), and 9.41 on Action questions that need completion, above its 7.81 over all Action questions (Table~\ref{tab:action-direct}). MemAct-9B also leads the retrieval baseline by 1.81 Memory ACC, 3.85 Memory MRA, and 9.56 Action Score, where similarity search cannot produce a new view (per-template results: Supp.\ Tab.~\ref{tab:per_template}).

\subsection{Past Evidence: Does Memory Help?}
\label{sec:exp_memory}

\begin{table}[t]
\centering
\caption{\small Memory QA Score by reference evidence location and template type (R / F, Section~\ref{sec:taxonomy}). Results by evidence age are in Supp.\ Tab.~\ref{tab:memory_age}.}
\label{tab:memory-grouped}
\vspace{-2mm}
\begingroup
\scriptsize
\setlength{\tabcolsep}{3pt}
\begin{tabular*}{\linewidth}{@{\extracolsep{\fill}}lcccc@{}}
\toprule
 & \multicolumn{2}{c}{Evidence} & \multicolumn{2}{c}{Template} \\
\cmidrule(lr){2-3}\cmidrule(l){4-5}
Model & In context & Evicted & R & F \\
\midrule
StreamingVLM & 76.03 & 44.76 & 62.13 & 49.37 \\
Same-backbone control & 80.90 & 56.67 & 68.27 & 62.71 \\
Retrieval baseline & 80.39 & 58.16 & 69.93 & 62.21 \\
Oracle retrieval (diagnostic) & 84.94 & 69.60 & 78.82 & 70.94 \\
MemAct-9B & 81.84 & 61.72 & 72.75 & 64.89 \\
\bottomrule
\end{tabular*}
\endgroup
\vspace{-4mm}
\end{table}

As intended, memory matters once evidence leaves the context (Table~\ref{tab:memory-grouped}): the gap to the same-backbone control widens with evidence age, from 0.81 points at most 10~s after the evidence to 5.21 beyond 120~s (Supp.\ Tab.~\ref{tab:memory_age}). MemAct retrieves selectively: on 78.0\% of questions with evicted evidence but 20.0\% with evidence in context.

\begin{table}[t]
\centering
\caption{\small Question-time retrieval ablation on Memory QA (Score) from identical pre-question states; the disabled branch blocks new retrieval for the current question only (protocol in Supp.~\ref{app:retrieval_ablation}). Means over three checkpoints; $\Delta$: paired difference [95\% CI]. Groups use the reference labels of Section~\ref{sec:test_construction}; $N$: unique questions.}
\label{tab:retrieval-ablation}
\vspace{-2mm}
\begingroup
\scriptsize
\setlength{\tabcolsep}{3pt}
\begin{tabular*}{\linewidth}{@{\extracolsep{\fill}}lrccc@{}}
\toprule
 & & \multicolumn{2}{c}{Retrieval} & \\
\cmidrule(lr){3-4}
Group (reference labels) & $N$ & Enabled & Disabled & $\Delta$ [95\% CI] \\
\midrule
All Memory QA & 7,964 & 69.37 & 65.61 & +3.76 [3.48, 4.03] \\
R, retained & 1,999 & 82.99 & 82.76 & +0.23 [0.10, 0.35] \\
R, evicted & 2,540 & 64.68 & 61.09 & +3.59 [3.17, 4.01] \\
F, retained & 1,027 & 79.60 & 78.25 & +1.36 [0.91, 1.79] \\
F, evicted & 2,398 & 58.59 & 50.67 & \textbf{+7.92} [7.22, 8.63] \\
\bottomrule
\end{tabular*}
\endgroup
\vspace{-4mm}
\end{table}

\paragraph{Retrieval beyond the retained context.}
To isolate retrieval, we compare retrieval-enabled and retrieval-disabled answering from identical pre-question states (Table~\ref{tab:retrieval-ablation}), so earlier records, retrieved text, and answers stay in context in both. Enabling retrieval improves Memory QA from 65.61 to 69.37, a gain of 3.76 Score points (95\% CI [3.48, 4.03]). The largest gain occurs on frame-required questions whose visual evidence is labeled as evicted (+7.92), consistent with retrieval supplying historical visual evidence, whereas record-answerable questions labeled as retained, whose records and frames are both still in context, change little (+0.23, [0.10, 0.35]). The disabled branch is a diagnostic for the current question, not a decomposition of MemAct's gain over the same-backbone control (Supp.~\ref{app:retrieval_ablation}).

\paragraph{Memory components and diagnostics.}
Retrained single-seed variants that each change one part of the memory system (Supp.\ Tab.~\ref{tab:memory-ablation}) show that removing historical frames lowers Memory ACC/MRA from 74.02/64.02 to 69.77/60.82; counts, relations, record retention, and write intervals are analyzed in Supp.~\ref{app:full_results}. MemAct-Bench's annotations also locate the remaining headroom: ground-truth records raise Memory ACC/MRA to 77.31/67.64, and oracle retrieval exceeds MemAct-9B by 6.05 points on frame-required templates (70.94 vs.\ 64.89) and by 7.88 on all evicted questions (69.60 vs.\ 61.72).

\subsection{Future Views: Do Actions Help?}
\label{sec:exp_action}

\begin{table}[t]
\centering
\caption{\small Action QA Score with the camera disabled or enabled. $\uparrow$/$\downarrow$: answers turned wrong$\to$right / right$\to$wrong by enabling it (numerical: right if MRA $\geq0.5$; mean over seeds). MemAct-9B answers for fixed policies and the best view.}
\label{tab:action-direct}
\vspace{-2mm}
\begingroup
\scriptsize
\setlength{\tabcolsep}{2.5pt}
\begin{tabular*}{\linewidth}{@{\extracolsep{\fill}}lccccc@{}}
\toprule
 & \multicolumn{2}{c}{All Action} & \multicolumn{2}{c}{Needs completion} & \\
\cmidrule(lr){2-3}\cmidrule(lr){4-5}
 & \multicolumn{4}{c}{Camera} & \\
Model / policy & Disabled & Enabled & Disabled & Enabled & $\uparrow$ / $\downarrow$ \\
\midrule
GPT-6 Astra & 46.23 & 46.91 & 42.98 & 43.71 & 9 / 3 \\
Gemini-3.8-Flash & 36.54 & 37.00 & 36.64 & 36.93 & 10 / 3 \\
Qwen3.5-9B & 20.22 & 20.41 & 19.97 & 20.07 & 2 / 1 \\
Same-backbone control & 44.46 & 44.78 & 41.18 & 41.42 & 6.3 / 2.3 \\
\midrule
Scripted scan (L, R, R) & 45.10 & 44.96 & 40.51 & 41.64 & 24.3 / 22.0 \\
Random commands & 45.10 & 43.83 & 40.51 & 40.63 & 21.0 / 28.0 \\
MemAct-9B & 45.10 & 52.59 & 40.51 & 50.83 & 62.0 / 4.3 \\
Best view within 3 steps & -- & 53.43 & -- & 53.51 & -- \\
\bottomrule
\end{tabular*}
\endgroup
\vspace{-4mm}
\end{table}

\paragraph{Camera enabled versus disabled.}
Disabling the camera removes camera commands but keeps each model's checkpoint, streams, memory and retrieval protocol, and budgets. For MemAct-9B, enabling the camera turns 62.0 wrong answers right and 4.3 right answers wrong (Table~\ref{tab:action-direct}), and raises Score from 45.10 to 52.59 on all Action questions (+7.49, 95\% CI [6.78, 8.23]) and from 40.51 to 50.83 on those that need completion (+10.32, [9.38, 11.27]), where the best view within three steps reaches 53.51. Zero-shot models gain at most 0.73 points because they rarely act, on at most 15.9\% of the questions that need completion, although their outputs follow the protocol (Supp.~\ref{app:baselines}). Among them, GPT-6 Astra has the highest camera-disabled Score (46.23), consistent with strong estimation from existing views, yet gains only 0.68; MemAct-9B gains 7.49 and overtakes it. Moving by itself helps little: a scripted scan and random commands score below the camera-disabled setting on all Action questions (44.96 and 43.83 vs.\ 45.10) and gain only 1.13 and 0.12 on need-completion questions. Views that miss the specific occlusion, truncation, or scale problem can mislead the answer (rollouts: Supp.\ Fig.~\ref{fig:action_sample_4}).

\begin{table}[t]
\centering
\caption{\small Stage 2 objectives on Action QA. Top: policies with the final MemAct-9B as answerer; Score overall and per stratum; Full vis.: full-visibility rate; Unnec.: \% of no-move-needed questions acted on; Att.: mean attempts. Bottom: paired Score differences [95\% CI] with the frozen Stage 1 model as answerer. More: Supp.\ Tabs.~\ref{tab:action_full} and~\ref{tab:answerer}.}
\label{tab:action-ablation}
\vspace{-2mm}
\begingroup
\scriptsize
\setlength{\tabcolsep}{1.5pt}
\begin{tabular*}{\linewidth}{@{\extracolsep{\fill}}lccccccc@{}}
\toprule
Setting & Score & No move & 1-step & Multi & Full vis. & Unnec. & Att. \\
\midrule
Camera disabled & 45.10 & 57.60 & 47.96 & 28.81 & 32.64 & 0.00 & 0.00 \\
SFT only & 48.07 & 57.13 & 52.25 & 32.94 & 66.03 & 69.62 & 2.05 \\
SFT + GRPO & 48.91 & 56.90 & 53.35 & 34.37 & 66.41 & 60.05 & 1.96 \\
SFT + Bandit (MemAct-9B) & \textbf{52.59} & 57.39 & \textbf{57.60} & \textbf{40.19} & \textbf{77.45} & 25.89 & 1.89 \\
\midrule
\emph{Frozen SFT answerer} & \multicolumn{3}{c}{$\Delta$ Score, all} & \multicolumn{4}{c}{$\Delta$ Score, needs compl.} \\
\midrule
Bandit $-$ SFT only & \multicolumn{3}{c}{+2.07 [1.78, 2.37]} & \multicolumn{4}{c}{+2.20 [1.83, 2.59]} \\
Bandit $-$ GRPO & \multicolumn{3}{c}{+1.75 [1.43, 2.08]} & \multicolumn{4}{c}{+1.90 [1.52, 2.30]} \\
\bottomrule
\end{tabular*}
\endgroup
\vspace{-4mm}
\end{table}

\paragraph{Bandit versus GRPO.}
Bandit outperforms GRPO where views must be completed and stays on par where no move is needed. All policies in Table~\ref{tab:action-ablation} (top) share MemAct-9B's pre-question states and answerer, so these rows differ only in the acquired views. GRPO starts from the same SFT checkpoint and uses the same reward, replay mixture, and update schedule as Bandit (Supp.~\ref{app:action_optimization}). Bandit exceeds GRPO by 3.68 points ([3.12, 4.25]) on all Action questions and by 4.85 ([4.13, 5.59]) on those that need completion, most on multi-step ones (40.19 vs.\ 34.37), and is on par on those that need no move (+0.49). It reaches a higher full-visibility rate (77.45\% vs.\ 66.41\%) with fewer attempts, acts on fewer no-move-needed questions (25.89\% vs.\ 60.05\%), and stops prematurely less often, in line with \texttt{answer\_now} competing with every command at the same state. Because this answerer is itself trained with Bandit, we also let the frozen Stage 1 model answer: Bandit's advantage persists, exceeding GRPO by 1.75 and 1.90 points on all and need-completion ones (Table~\ref{tab:action-ablation}, bottom).

\paragraph{Reward and action budget.}
In the cumulative reward ablation, the last step, which measures gains against the best historical view, adds 2.83 Score points ([2.31, 3.36]); with the same policy, raising the attempt limit from one to three adds 3.47 ([2.96, 3.98]), and three attempts suffice: five or eight add at most 0.23, since an adequate view is reachable within three steps by construction (Supp.~\ref{app:full_results}).

\subsection{Failure Analysis}
\label{sec:exp_failure}

\begin{table}[t]
\centering
\caption{\small Failure attribution (\% of wrong answers, pooled over seeds; definitions in Supp.~\ref{app:attribution}). Memory (first matching rule): M1 write error, M2 not retrieved, M3 wrong although retrieved, M4 wrong although in context. Action: A1 premature stop, A2 ineffective exploration, A3 wrong although an adequate view was reached.}
\label{tab:failure}
\vspace{-2mm}
\begingroup
\scriptsize
\setlength{\tabcolsep}{3pt}
\begin{tabular*}{\linewidth}{@{\extracolsep{\fill}}lcccc@{}}
\toprule
Memory errors & M1 & M2 & M3 & M4 \\
\midrule
MemAct-4B & 37.10 & 32.27 & 16.31 & 14.32 \\
MemAct-9B & 28.95 & 32.58 & 19.17 & 19.29 \\
\midrule
Action errors (MemAct-9B) & A1 & A2 & A3 & \\
\midrule
SFT only & 3.99 & 5.77 & 90.25 & \\
SFT + GRPO & 3.91 & 5.09 & 90.99 & \\
SFT + Bandit & 1.90 & 3.80 & 94.30 & \\
\bottomrule
\end{tabular*}
\endgroup
\vspace{-4mm}
\end{table}

Fixed rules assign each error to a stage from the logged writes, retrievals, commands, and views (Table~\ref{tab:failure}); a manual check agrees with 148 of 150 automatic labels, showing consistent rules though not a unique cause per error. The share of writing errors (M1) falls from 37.10\% for MemAct-4B to 28.95\% for MemAct-9B, whose largest share is M2 (32.58\%), evidence written but not returned. On Action, Bandit has the smallest share of premature stops and ineffective explorations (A1+A2: 5.70\% vs.\ 9.00\% for GRPO and 9.76\% for SFT only); the rest fall under A3.

\subsection{Transfer to Other Benchmarks}
\label{sec:exp_external}

View completion and memory also help beyond MemAct-Bench. Without training on them, MemAct-9B reaches 68.39 on AVS-HM3D~\citep{koo2025vgavs}, 9.74 points above answering from the query view, and 45.65 on ESI-Bench~\citep{hong2026esi}, whose tasks also require locomotion and manipulation. On VSI-Bench~\citep{yang2024thinking} and ReVSI~\citep{zhang2026revsi}, memory adds 3.41 and 4.06 points. Protocols and published references: Supp.~\ref{app:external}.

\section{Limitations and Conclusion}
\label{sec:limitations}
\label{sec:conclusion}

\textbf{Limitations.} MemAct targets local view completion of objects already in sight; long-horizon navigation and search for unseen objects lie outside this scope and are complementary directions~\citep{ginting2025mindpalace,wang2026lmee,yang2024mem3d}. Camera actions are learned in 3D-FRONT simulation, whose resets provide the counterfactual branches of Stage 2; transfer to AVS-HM3D scans (Section~\ref{sec:exp_external}) suggests that the policy carries over beyond these scenes. Acquired views serve only the question that requested them, and text is retained within streams of up to 180~s; writing acquired views into memory and managing text over longer streams are natural extensions.

\textbf{Conclusion.} MemAct answers spatial questions whose evidence lies in the past, present, or future of a stream: under a bounded active context, it decides from RGB alone when to retrieve its self-written memory and when to take local camera actions. Bandit-based counterfactual optimization credits preparatory actions and stopping by comparing alternatives at the same state, and outperforms GRPO, especially on questions that need view completion. On MemAct-Bench, whose annotations locate each question's evidence, retrieval helps most when the needed frames have been evicted, and camera actions help where views need completion.

\section*{LLM Usage Statement}
AI tools assisted discussions of research ideas, code implementation, and manuscript preparation, including literature search and checking, language revision, figure editing, and experiment evaluation. Tools and models used included Codex, GPT-5.6-sol, GPT-6 Astra, Gemini-3.8-Flash, and Claude. Dataset construction also used LLM-assisted QA wording refinement while preserving grounded referents and geometry-derived answers. Separately, the benchmark evaluates language and vision--language models as experimental subjects. The authors directed these uses, reviewed and revised the assisted content, and take responsibility for the final paper and its scientific claims.

{
\small
\bibliographystyle{ieeenat_fullname}
\bibliography{iclr2027_conference}

@inproceedings{yang2024thinking,
  title     = {{Thinking in Space: How Multimodal Large Language Models See, Remember, and Recall Spaces}},
  author    = {Yang, Jihan and Yang, Shusheng and Gupta, Anjali W. and Han, Rilyn and Fei-Fei, Li and Xie, Saining},
  booktitle = {Proceedings of the IEEE/CVF Conference on Computer Vision and Pattern Recognition (CVPR)},
  pages     = {10632--10643},
  year      = {2025},
  url       = {https://openaccess.thecvf.com/content/CVPR2025/html/Yang_Thinking_in_Space_How_Multimodal_Large_Language_Models_See_Remember_CVPR_2025_paper.html}
}

@inproceedings{ray2024sat,
  title     = {{SAT: Dynamic Spatial Aptitude Training for Multimodal Language Models}},
  author    = {Ray, Arijit and Duan, Jiafei and Brown, Ellis and Tan, Reuben and Bashkirova, Dina and Hendrix, Rose and Ehsani, Kiana and Kembhavi, Aniruddha and Plummer, Bryan A. and Krishna, Ranjay and Zeng, Kuo-Hao and Saenko, Kate},
  booktitle = {Conference on Language Modeling (COLM)},
  year      = {2025},
  url       = {https://openreview.net/forum?id=DW8U8ZWa1U}
}

@inproceedings{yang2025cambrians,
  title     = {{Cambrian-S: Towards Spatial Supersensing in Video}},
  author    = {Yang, Shusheng and Yang, Jihan and Huang, Pinzhi and Brown, Ellis and Yang, Zihao and Yu, Yue and Tong, Shengbang and Zheng, Zihan and Xu, Yifan and Wang, Muhan and Fergus, Rob and LeCun, Yann and Fei-Fei, Li and Xie, Saining},
  booktitle = {International Conference on Learning Representations (ICLR)},
  year      = {2026},
  url       = {https://openreview.net/forum?id=rBFDvZu6pb}
}

@inproceedings{xu2025streamingvlm,
  title     = {{StreamingVLM: Real-Time Understanding for Infinite Video Streams}},
  author    = {Xu, Ruyi and Xiao, Guangxuan and Chen, Yukang and He, Liuning and Peng, Kelly and Lu, Yao and Han, Song},
  booktitle = {International Conference on Learning Representations (ICLR)},
  year      = {2026},
  url       = {https://openreview.net/forum?id=gVbPWbA97s}
}

@inproceedings{yu2026stream3d,
  title     = {{Stream3D-VLM: Online 3D Spatial Understanding with Incremental Geometry Priors}},
  author    = {Yu, Hanxun and Qu, Xuan and Ke, Lei and Zhang, Boqiang and Wang, Yuxin and Zhu, Jianke and Yu, Dong},
  booktitle = {European Conference on Computer Vision (ECCV)},
  pages     = {130--148},
  year      = {2026},
  doi       = {10.1007/978-3-032-37211-6_8}
}

@inproceedings{wang2025streameqa,
  title     = {{StreamEQA: Towards Streaming Video Understanding for Embodied Scenarios}},
  author    = {Wang, Yifei and Li, Zhenkai and Qian, Tianwen and Zheng, Huanran and Wang, Zheng and Fu, Yuqian and Wang, Xiaoling},
  booktitle = {Proceedings of the IEEE/CVF Conference on Computer Vision and Pattern Recognition (CVPR) Findings},
  pages     = {9422--9432},
  year      = {2026},
  url       = {https://openaccess.thecvf.com/content/CVPR2026F/html/Wang_StreamEQA_Towards_Streaming_Video_Understanding_for_Embodied_Scenarios_CVPRF_2026_paper.html}
}

@inproceedings{yang2024mem3d,
  title     = {{3D-Mem: 3D Scene Memory for Embodied Exploration and Reasoning}},
  author    = {Yang, Yuncong and Yang, Han and Zhou, Jiachen and Chen, Peihao and Zhang, Hongxin and Du, Yilun and Gan, Chuang},
  booktitle = {Proceedings of the IEEE/CVF Conference on Computer Vision and Pattern Recognition (CVPR)},
  pages     = {17294--17303},
  year      = {2025},
  url       = {https://openaccess.thecvf.com/content/CVPR2025/html/Yang_3D-Mem_3D_Scene_Memory_for_Embodied_Exploration_and_Reasoning_CVPR_2025_paper.html}
}

@article{hong2026esi,
  title = {{ESI-Bench: Towards Embodied Spatial Intelligence that Closes the Perception-Action Loop}},
  author = {Hong, Yining and Liu, Jiageng and Yin, Han and Li, Manling and Guibas, Leonidas and Fei-Fei, Li and Wu, Jiajun and Choi, Yejin},
  journal = {arXiv preprint arXiv:2605.18746},
  year = {2026},
  doi = {10.48550/arXiv.2605.18746},
  url = {https://arxiv.org/abs/2605.18746}
}

@inproceedings{fu2021front,
  title = {{3D-FRONT: 3D Furnished Rooms With layOuts and semaNTics}},
  author = {Fu, Huan and Cai, Bowen and Gao, Lin and Zhang, Ling-Xiao and Wang, Jiaming and Li, Cao and Zeng, Qixun and Sun, Chengyue and Jia, Rongfei and Zhao, Binqiang and Zhang, Hao},
  booktitle = {Proceedings of the IEEE/CVF International Conference on Computer Vision (ICCV)},
  pages = {10933--10942},
  year = {2021},
  url = {https://openaccess.thecvf.com/content/ICCV2021/html/Fu_3D-FRONT_3D_Furnished_Rooms_With_layOuts_and_semaNTics_ICCV_2021_paper.html}
}

@inproceedings{dai2017scannet,
  title     = {{ScanNet: Richly-Annotated 3D Reconstructions of Indoor Scenes}},
  author    = {Dai, Angela and Chang, Angel X. and Savva, Manolis and Halber, Maciej and Funkhouser, Thomas and Nie{\ss}ner, Matthias},
  booktitle = {Proceedings of the IEEE Conference on Computer Vision and Pattern Recognition (CVPR)},
  pages     = {2432--2443},
  year      = {2017},
  doi       = {10.1109/CVPR.2017.261},
  url       = {https://openaccess.thecvf.com/content_cvpr_2017/html/Dai_ScanNet_Richly-Annotated_3D_CVPR_2017_paper.html}
}

@inproceedings{yeshwanth2023scannetpp,
  title     = {{ScanNet++: A High-Fidelity Dataset of 3D Indoor Scenes}},
  author    = {Yeshwanth, Chandan and Liu, Yueh-Cheng and Nie{\ss}ner, Matthias and Dai, Angela},
  booktitle = {Proceedings of the IEEE/CVF International Conference on Computer Vision (ICCV)},
  pages     = {12--22},
  year      = {2023},
  url       = {https://openaccess.thecvf.com/content/ICCV2023/html/Yeshwanth_ScanNet_A_High-Fidelity_Dataset_of_3D_Indoor_Scenes_ICCV_2023_paper.html}
}

@inproceedings{baruch2021arkitscenes,
  title     = {{ARKitScenes: A Diverse Real-World Dataset For 3D Indoor Scene Understanding Using Mobile RGB-D Data}},
  author    = {Baruch, Gilad and Chen, Zhuoyuan and Dehghan, Afshin and Dimry, Tal and Feigin, Yuri and Fu, Peter and Gebauer, Thomas and Joffe, Brandon and Kurz, Daniel and Schwartz, Arik and Shulman, Elad},
  booktitle = {Proceedings of the Neural Information Processing Systems Track on Datasets and Benchmarks},
  volume    = {1},
  year      = {2021},
  url       = {https://datasets-benchmarks-proceedings.neurips.cc/paper_files/paper/2021/hash/66f041e16a60928b05a7e228a89c3799-Abstract-round1.html}
}

@inproceedings{zhang2026revsi,
  title     = {{ReVSI: Rebuilding Visual Spatial Intelligence Evaluation for Accurate Assessment of VLM 3D Reasoning}},
  author    = {Zhang, Yiming and Chen, Jiacheng and Tan, Jiaqi and Mao, Yongsen and Chen, Wenhu and Chang, Angel X.},
  booktitle = {Proceedings of the 43rd International Conference on Machine Learning (ICML)},
  series    = {Proceedings of Machine Learning Research},
  volume    = {306},
  pages     = {154497--154518},
  year      = {2026},
  url       = {https://proceedings.mlr.press/v306/zhang26o.html}
}

@inproceedings{yang2025mmsi,
  title     = {{MMSI-Bench: A Benchmark for Multi-Image Spatial Intelligence}},
  author    = {Yang, Sihan and Xu, Runsen and Xie, Yiman and Yang, Sizhe and Li, Mo and Lin, Jingli and Zhu, Chenming and Chen, Xiaochen and Duan, Haodong and Yue, Xiangyu and Lin, Dahua and Wang, Tai and Pang, Jiangmiao},
  booktitle = {International Conference on Learning Representations (ICLR)},
  year      = {2026},
  url       = {https://openreview.net/forum?id=gHRoX4vXm3}
}

@inproceedings{li2025viewspatial,
  title     = {{ViewSpatial-Bench: Evaluating Multi-perspective Spatial Localization in Vision-Language Models}},
  author    = {Li, Dingming and Li, Hongxing and Wang, Zixuan and Yan, Yuchen and Zhang, Hang and Chen, Siqi and Hou, Guiyang and Jiang, Shengpei and Zhang, Wenqi and Shen, Yongliang and Lu, Weiming and Zhuang, Yueting},
  booktitle = {European Conference on Computer Vision (ECCV)},
  pages     = {95--111},
  year      = {2026},
  doi       = {10.1007/978-3-032-37098-3_6}
}

@inproceedings{fan2025vlm3r,
  title     = {{VLM-3R: Vision-Language Models Augmented with Instruction-Aligned 3D Reconstruction}},
  author    = {Fan, Zhiwen and Zhang, Jian and Li, Renjie and Zhang, Junge and Chen, Runjin and Hu, Hezhen and Wang, Kevin and Wang, Peihao and Qu, Huaizhi and Zhou, Shijie and Wang, Dilin and Yan, Zhicheng and Xu, Hongyu and Theiss, Justin and Chen, Tianlong and Li, Jiachen and Tu, Zhengzhong and Wang, Zhangyang and Ranjan, Rakesh},
  booktitle = {Proceedings of the IEEE/CVF Conference on Computer Vision and Pattern Recognition (CVPR)},
  pages     = {31054--31065},
  year      = {2026},
  url       = {https://openaccess.thecvf.com/content/CVPR2026/html/Fan_VLM-3R_Vision-Language_Models_Augmented_with_Instruction-Aligned_3D_Reconstruction_CVPR_2026_paper.html}
}

@article{wang2025internvl35,
  title = {{InternVL3.5: Advancing Open-Source Multimodal Models in Versatility, Reasoning, and Efficiency}},
  author = {Wang, Weiyun and Gao, Zhangwei and Gu, Lixin and Pu, Hengjun and Cui, Long and Wei, Xingguang and Liu, Zhaoyang and Jing, Linglin and Ye, Shenglong and Shao, Jie and Wang, Zhaokai and Chen, Zhe and Zhang, Hongjie and Yang, Ganlin and Wang, Haomin and Wei, Qi and Yin, Jinhui and Li, Wenhao and Cui, Erfei and Chen, Guanzhou and Ding, Zichen and Tian, Changyao and Wu, Zhenyu and Xie, Jingjing and Li, Zehao and Yang, Bowen and Duan, Yuchen and Wang, Xuehui and Hou, Zhi and Hao, Haoran and Zhang, Tianyi and Li, Songze and Zhao, Xiangyu and Duan, Haodong and Deng, Nianchen and Fu, Bin and He, Yinan and Wang, Yi and He, Conghui and Shi, Botian and He, Junjun and Xiong, Yingtong and Lv, Han and Wu, Lijun and Shao, Wenqi and Zhang, Kaipeng and Deng, Huipeng and Qi, Biqing and Ge, Jiaye and Guo, Qipeng and Zhang, Wenwei and Zhang, Songyang and Cao, Maosong and Lin, Junyao and Tang, Kexian and Gao, Jianfei and Huang, Haian and Gu, Yuzhe and Lyu, Chengqi and Tang, Huanze and Wang, Rui and Lv, Haijun and Ouyang, Wanli and Wang, Limin and Dou, Min and Zhu, Xizhou and Lu, Tong and Lin, Dahua and Dai, Jifeng and Su, Weijie and Zhou, Bowen and Chen, Kai and Qiao, Yu and Wang, Wenhai and Luo, Gen},
  journal = {arXiv preprint arXiv:2508.18265},
  year = {2025},
  doi = {10.48550/arXiv.2508.18265},
  url = {https://arxiv.org/abs/2508.18265}
}

@inproceedings{hu2021lora,
  title     = {{LoRA: Low-Rank Adaptation of Large Language Models}},
  author    = {Hu, Edward J. and Shen, Yelong and Wallis, Phillip and Allen-Zhu, Zeyuan and Li, Yuanzhi and Wang, Shean and Wang, Lu and Chen, Weizhu},
  booktitle = {International Conference on Learning Representations (ICLR)},
  year      = {2022},
  url       = {https://openreview.net/forum?id=nZeVKeeFYf9}
}

@inproceedings{loshchilov2017adamw,
  title     = {{Decoupled Weight Decay Regularization}},
  author    = {Loshchilov, Ilya and Hutter, Frank},
  booktitle = {International Conference on Learning Representations (ICLR)},
  year      = {2019},
  url       = {https://openreview.net/forum?id=Bkg6RiCqY7}
}

@article{shao2024deepseekmath,
  title = {{DeepSeekMath: Pushing the Limits of Mathematical Reasoning in Open Language Models}},
  author = {Shao, Zhihong and Wang, Peiyi and Zhu, Qihao and Xu, Runxin and Song, Junxiao and Bi, Xiao and Zhang, Haowei and Zhang, Mingchuan and Li, Y. K. and Wu, Y. and Guo, Daya},
  journal = {arXiv preprint arXiv:2402.03300},
  year = {2024},
  doi = {10.48550/arXiv.2402.03300},
  url = {https://arxiv.org/abs/2402.03300}
}

@article{hart1968astar,
  title = {A Formal Basis for the Heuristic Determination of Minimum Cost Paths},
  author = {Hart, Peter E. and Nilsson, Nils J. and Raphael, Bertram},
  journal = {IEEE Transactions on Systems Science and Cybernetics},
  volume = {4},
  number = {2},
  pages = {100--107},
  year = {1968},
  doi = {10.1109/TSSC.1968.300136},
  url = {https://doi.org/10.1109/TSSC.1968.300136}
}

@misc{qwen2026qwen35,
  title        = {{Qwen3.5: Towards Native Multimodal Agents}},
  author       = {{Qwen Team}},
  year         = {2026},
  howpublished = {Blog post},
  url          = {https://qwen.ai/blog?id=qwen3.5},
  note         = {Accessed 2026-09-24}
}

@misc{qwen2026qwen38,
  title = {{Qwen3.8-27B}},
  author = {{Qwen Team}},
  year = {2026},
  howpublished = {Official model card},
  url = {https://huggingface.co/Qwen/Qwen3.8-27B},
  note = {Accessed 2026-09-24}
}

@misc{openai2026astra,
  title = {{GPT-6 Astra System Card}},
  author = {{OpenAI}},
  year = {2026},
  howpublished = {System card},
  url = {https://deploymentsafety.openai.com/gpt-6-astra},
  note = {Accessed 2026-09-24}
}

@misc{deepmind2026gemini38,
  title = {{Gemini 3.8 Flash Model Card}},
  author = {{Google DeepMind}},
  year = {2026},
  howpublished = {Model card},
  url = {https://deepmind.google/models/model-cards/gemini-3-8-flash/},
  note = {Accessed 2026-09-24}
}

@misc{blenderCycles,
  title = {{Cycles}},
  author = {{Blender Foundation}},
  year = {2026},
  howpublished = {Official renderer documentation},
  url = {https://www.cycles-renderer.org/},
  note = {Accessed 2026-09-24}
}

@article{li2026orarl,
  title   = {{Annotations as Rollouts: Efficient and Scalable Reinforcement Learning for Video MLLMs}},
  author  = {Li, Yunheng and Mu, Guohong and Li, Hao and Qian, Shengsheng and Zhang, Dingwen and Hou, Qibin and Cheng, Ming-Ming},
  journal = {arXiv preprint arXiv:2608.20492},
  year    = {2026},
  doi     = {10.48550/arXiv.2608.20492},
  url     = {https://arxiv.org/abs/2608.20492}
}

@article{dai2026sagent,
  title   = {{S-Agent: Spatial Tool-Use Elicits Reasoning for Spatial Intelligence}},
  author  = {Dai, Yalun and Li, Hao and Tian, Shulin and Yao, Runmao and Dong, Yuhao and Hong, Fangzhou and Chen, Zhaoxi and Liu, Fangfu and Wang, Tao and Yap, Kim-Hui and Liu, Ziwei},
  journal = {arXiv preprint arXiv:2606.20515},
  year    = {2026},
  doi     = {10.48550/arXiv.2606.20515},
  url     = {https://arxiv.org/abs/2606.20515}
}

@article{wang2026guide,
  title   = {{Let Geometry GUIDE: Layer-wise Unrolling of Geometric Priors in Multimodal LLMs}},
  author  = {Wang, Chongyu and Huang, Ting and Sun, Chunyu and Ning, Xinyu and Wang, Di and Tang, Hao},
  journal = {arXiv preprint arXiv:2604.05695},
  year    = {2026},
  doi     = {10.48550/arXiv.2604.05695},
  url     = {https://arxiv.org/abs/2604.05695}
}

@misc{taichu2026results,
  title        = {{ZDTaichu5.0-9B: A Multimodal Foundation Model for Visual and Spatial Reasoning, Agents, and Embodied AI}},
  author       = {{ZDTaichu5.0-9B Contributors}},
  year         = {2026},
  howpublished = {Hugging Face model card},
  url          = {https://huggingface.co/TaichuAI/ZDTaichu5.0-9B},
  note         = {Accessed 2026-09-24}
}

@misc{spatialaxiom2026results,
  title        = {{SpatialAxiom: An Open Spatial Intelligence Model for General Spatial Reasoning}},
  author       = {Lou, Yujing and Chen, Pingyi and Cao, Shen and Gu, Jiaqi and Guo, Jinhui and Tong, Jintao and Hao, Yunzhuo and Liu, Yao and Wu, Yue and Fan, Lubin and Ye, Jieping},
  year         = {2026},
  howpublished = {Hugging Face model card},
  url          = {https://huggingface.co/SpatialAxiom/SpatialAxiom-9B},
  note         = {Accessed 2026-09-24}
}

@inproceedings{azuma2021scanqa,
  title     = {{ScanQA: 3D Question Answering for Spatial Scene Understanding}},
  author    = {Azuma, Daichi and Miyanishi, Taiki and Kurita, Shuhei and Kawanabe, Motoaki},
  booktitle = {Proceedings of the IEEE/CVF Conference on Computer Vision and Pattern Recognition (CVPR)},
  pages     = {19129--19139},
  year      = {2022},
  url       = {https://openaccess.thecvf.com/content/CVPR2022/html/Azuma_ScanQA_3D_Question_Answering_for_Spatial_Scene_Understanding_CVPR_2022_paper.html}
}

@inproceedings{ma2022sqa3d,
  title     = {{SQA3D: Situated Question Answering in 3D Scenes}},
  author    = {Ma, Xiaojian and Yong, Silong and Zheng, Zilong and Li, Qing and Liang, Yitao and Zhu, Song-Chun and Huang, Siyuan},
  booktitle = {International Conference on Learning Representations (ICLR)},
  year      = {2023},
  url       = {https://openreview.net/forum?id=IDJx97BC38}
}

@inproceedings{chen2024spatialvlm,
  title     = {{SpatialVLM: Endowing Vision-Language Models with Spatial Reasoning Capabilities}},
  author    = {Chen, Boyuan and Xu, Zhuo and Kirmani, Sean and Ichter, Brian and Sadigh, Dorsa and Guibas, Leonidas and Xia, Fei},
  booktitle = {Proceedings of the IEEE/CVF Conference on Computer Vision and Pattern Recognition (CVPR)},
  pages     = {14455--14465},
  year      = {2024},
  doi       = {10.1109/CVPR52733.2024.01370},
  url       = {https://openaccess.thecvf.com/content/CVPR2024/html/Chen_SpatialVLM_Endowing_Vision-Language_Models_with_Spatial_Reasoning_Capabilities_CVPR_2024_paper.html}
}

@inproceedings{cheng2024spatialrgpt,
  title     = {{SpatialRGPT: Grounded Spatial Reasoning in Vision-Language Models}},
  author    = {Cheng, An-Chieh and Yin, Hongxu and Fu, Yang and Guo, Qiushan and Yang, Ruihan and Kautz, Jan and Wang, Xiaolong and Liu, Sifei},
  booktitle = {Advances in Neural Information Processing Systems (NeurIPS)},
  volume    = {37},
  pages     = {135062--135093},
  year      = {2024},
  url       = {https://proceedings.neurips.cc/paper_files/paper/2024/hash/f38cb4cf9a5eaa92b3cfa481832719c6-Abstract-Conference.html}
}

@inproceedings{fu2024blink,
  title     = {{BLINK: Multimodal Large Language Models Can See but Not Perceive}},
  author    = {Fu, Xingyu and Hu, Yushi and Li, Bangzheng and Feng, Yu and Wang, Haoyu and Lin, Xudong and Roth, Dan and Smith, Noah A. and Ma, Wei-Chiu and Krishna, Ranjay},
  booktitle = {European Conference on Computer Vision (ECCV)},
  pages     = {148--166},
  year      = {2024},
  doi       = {10.1007/978-3-031-73337-6_9}
}

@inproceedings{song2024robospatial,
  title     = {{RoboSpatial: Teaching Spatial Understanding to 2D and 3D Vision-Language Models for Robotics}},
  author    = {Song, Chan Hee and Blukis, Valts and Tremblay, Jonathan and Tyree, Stephen and Su, Yu and Birchfield, Stan},
  booktitle = {Proceedings of the IEEE/CVF Conference on Computer Vision and Pattern Recognition (CVPR)},
  pages     = {15768--15780},
  year      = {2025},
  url       = {https://openaccess.thecvf.com/content/CVPR2025/html/Song_RoboSpatial_Teaching_Spatial_Understanding_to_2D_and_3D_Vision-Language_Models_CVPR_2025_paper.html}
}

@inproceedings{zhang2024flashvstream,
  title     = {{Flash-VStream: Efficient Real-Time Understanding for Long Video Streams}},
  author    = {Zhang, Haoji and Wang, Yiqin and Tang, Yansong and Liu, Yong and Feng, Jiashi and Jin, Xiaojie},
  booktitle = {Proceedings of the IEEE/CVF International Conference on Computer Vision (ICCV)},
  pages     = {21059--21069},
  year      = {2025},
  url       = {https://openaccess.thecvf.com/content/ICCV2025/html/Zhang_Flash-VStream_Efficient_Real-Time_Understanding_for_Long_Video_Streams_ICCV_2025_paper.html}
}

@inproceedings{lin2024streamingbench,
  title     = {{StreamingBench: Assessing the Gap for MLLMs to Achieve Streaming Video Understanding}},
  author    = {Lin, Junming and Fang, Zheng and Chen, Chi and Cheng, Haoxuan and Wan, Zihao and Luo, Fuwen and Wang, Ziyue and Li, Peng and Liu, Yang and Sun, Maosong},
  booktitle = {IEEE International Conference on Acoustics, Speech and Signal Processing (ICASSP)},
  pages     = {12147--12151},
  year      = {2026},
  doi       = {10.1109/ICASSP55912.2026.11463959}
}

@inproceedings{yao2025timechatonline,
  title     = {{TimeChat-Online: 80\% Visual Tokens are Naturally Redundant in Streaming Videos}},
  author    = {Yao, Linli and Li, Yicheng and Wei, Yuancheng and Li, Lei and Ren, Shuhuai and Liu, Yuanxin and Ouyang, Kun and Wang, Lean and Li, Shicheng and Li, Sida and Kong, Lingpeng and Liu, Qi and Zhang, Yuanxing and Sun, Xu},
  booktitle = {Proceedings of the 33rd ACM International Conference on Multimedia (MM)},
  pages     = {10807--10816},
  year      = {2025},
  doi       = {10.1145/3746027.3754839}
}

@inproceedings{chen2025livecc,
  title     = {{LiveCC: Learning Video LLM with Streaming Speech Transcription at Scale}},
  author    = {Chen, Joya and Zeng, Ziyun and Lin, Yiqi and Li, Wei and Ma, Zejun and Shou, Mike Zheng},
  booktitle = {Proceedings of the IEEE/CVF Conference on Computer Vision and Pattern Recognition (CVPR)},
  pages     = {29083--29095},
  year      = {2025},
  url       = {https://openaccess.thecvf.com/content/CVPR2025/html/Chen_LiveCC_Learning_Video_LLM_with_Streaming_Speech_Transcription_at_Scale_CVPR_2025_paper.html}
}

@inproceedings{gu2023conceptgraphs,
  title     = {{ConceptGraphs: Open-Vocabulary 3D Scene Graphs for Perception and Planning}},
  author    = {Gu, Qiao and Kuwajerwala, Alihusein and Morin, Sacha and Jatavallabhula, Krishna Murthy and Sen, Bipasha and Agarwal, Aditya and Rivera, Corban and Paul, William and Ellis, Kirsty and Chellappa, Rama and Gan, Chuang and de Melo, Celso Miguel and Tenenbaum, Joshua B. and Torralba, Antonio and Shkurti, Florian and Paull, Liam},
  booktitle = {IEEE International Conference on Robotics and Automation (ICRA)},
  pages     = {5021--5028},
  year      = {2024},
  doi       = {10.1109/ICRA57147.2024.10610243}
}

@inproceedings{jatavallabhula2023conceptfusion,
  title     = {{ConceptFusion: Open-set Multimodal 3D Mapping}},
  author    = {Jatavallabhula, Krishna Murthy and Kuwajerwala, Alihusein and Gu, Qiao and Omama, Mohd and Iyer, Ganesh and Saryazdi, Soroush and Chen, Tao and Maalouf, Alaa and Li, Shuang and Keetha, Nikhil Varma and Tewari, Ayush and Tenenbaum, Joshua B. and de Melo, Celso Miguel and Krishna, Madhava and Paull, Liam and Shkurti, Florian and Torralba, Antonio},
  booktitle = {Robotics: Science and Systems (RSS)},
  year      = {2023},
  doi       = {10.15607/RSS.2023.XIX.066},
  url       = {https://www.roboticsproceedings.org/rss19/p066.html}
}

@article{zhai2025memoryeqa,
  title = {{Memory-Centric Embodied Question Answering}},
  author = {Zhai, Mingliang and Gao, Zhi and Wu, Yuwei and Jia, Yunde},
  journal = {arXiv preprint arXiv:2505.13948},
  year = {2025},
  doi = {10.48550/arXiv.2505.13948},
  url = {https://arxiv.org/abs/2505.13948}
}

@inproceedings{ginting2025mindpalace,
  title     = {{Enter the Mind Palace: Reasoning and Planning for Long-term Active Embodied Question Answering}},
  author    = {Ginting, Muhammad Fadhil and Kim, Dong-Ki and Meng, Xiangyun and Reinke, Andrzej and Krishna, Bandi Jai and Kayhani, Navid and Peltzer, Oriana and Fan, David D. and Shaban, Amirreza and Kim, Sung-Kyun and Kochenderfer, Mykel J. and Agha-mohammadi, Ali-akbar and Omidshafiei, Shayegan},
  booktitle = {Proceedings of the 9th Conference on Robot Learning (CoRL)},
  series    = {Proceedings of Machine Learning Research},
  volume    = {305},
  pages     = {5072--5106},
  year      = {2025},
  url       = {https://proceedings.mlr.press/v305/ginting25a.html}
}

@inproceedings{ren2024exploreconfident,
  title     = {{Explore until Confident: Efficient Exploration for Embodied Question Answering}},
  author    = {Ren, Allen Z. and Clark, Jaden and Dixit, Anushri and Itkina, Masha and Majumdar, Anirudha and Sadigh, Dorsa},
  booktitle = {Robotics: Science and Systems (RSS)},
  year      = {2024},
  doi       = {10.15607/RSS.2024.XX.089},
  url       = {https://www.roboticsproceedings.org/rss20/p089.html}
}

@inproceedings{das2017embodiedqa,
  title     = {{Embodied Question Answering}},
  author    = {Das, Abhishek and Datta, Samyak and Gkioxari, Georgia and Lee, Stefan and Parikh, Devi and Batra, Dhruv},
  booktitle = {Proceedings of the IEEE Conference on Computer Vision and Pattern Recognition (CVPR)},
  pages     = {1--10},
  year      = {2018},
  doi       = {10.1109/CVPR.2018.00008},
  url       = {https://openaccess.thecvf.com/content_cvpr_2018/html/Das_Embodied_Question_Answering_CVPR_2018_paper.html}
}

@inproceedings{gordon2017iqa,
  title     = {{IQA: Visual Question Answering in Interactive Environments}},
  author    = {Gordon, Daniel and Kembhavi, Aniruddha and Rastegari, Mohammad and Redmon, Joseph and Fox, Dieter and Farhadi, Ali},
  booktitle = {Proceedings of the IEEE Conference on Computer Vision and Pattern Recognition (CVPR)},
  pages     = {4089--4098},
  year      = {2018},
  doi       = {10.1109/CVPR.2018.00430},
  url       = {https://openaccess.thecvf.com/content_cvpr_2018/html/Gordon_IQA_Visual_Question_CVPR_2018_paper.html}
}

@inproceedings{yokoyama2023vlfm,
  title     = {{VLFM: Vision-Language Frontier Maps for Zero-Shot Semantic Navigation}},
  author    = {Yokoyama, Naoki and Ha, Sehoon and Batra, Dhruv and Wang, Jiuguang and Bucher, Bernadette},
  booktitle = {IEEE International Conference on Robotics and Automation (ICRA)},
  pages     = {42--48},
  year      = {2024},
  doi       = {10.1109/ICRA57147.2024.10610712}
}

@inproceedings{huang2022vlmaps,
  title     = {{Visual Language Maps for Robot Navigation}},
  author    = {Huang, Chenguang and Mees, Oier and Zeng, Andy and Burgard, Wolfram},
  booktitle = {IEEE International Conference on Robotics and Automation (ICRA)},
  pages     = {10608--10615},
  year      = {2023},
  doi       = {10.1109/ICRA48891.2023.10160969}
}

@inproceedings{he2024malmm,
  title     = {{MA-LMM: Memory-Augmented Large Multimodal Model for Long-Term Video Understanding}},
  author    = {He, Bo and Li, Hengduo and Jang, Young Kyun and Jia, Menglin and Cao, Xuefei and Shah, Ashish and Shrivastava, Abhinav and Lim, Ser-Nam},
  booktitle = {Proceedings of the IEEE/CVF Conference on Computer Vision and Pattern Recognition (CVPR)},
  pages     = {13504--13514},
  year      = {2024},
  url       = {https://openaccess.thecvf.com/content/CVPR2024/html/He_MA-LMM_Memory-Augmented_Large_Multimodal_Model_for_Long-Term_Video_Understanding_CVPR_2024_paper.html}
}

@inproceedings{huang2024ovbench,
  title     = {{Online Video Understanding: OVBench and VideoChat-Online}},
  author    = {Huang, Zhenpeng and Li, Xinhao and Li, Jiaqi and Wang, Jing and Zeng, Xiangyu and Liang, Cheng and Wu, Tao and Chen, Xi and Li, Liang and Wang, Limin},
  booktitle = {Proceedings of the IEEE/CVF Conference on Computer Vision and Pattern Recognition (CVPR)},
  pages     = {3328--3338},
  year      = {2025},
  url       = {https://openaccess.thecvf.com/content/CVPR2025/html/Huang_Online_Video_Understanding_OVBench_and_VideoChat-Online_CVPR_2025_paper.html}
}

@InProceedings{majumdar2024openeqa,
    author    = {Majumdar, Arjun and Ajay, Anurag and Zhang, Xiaohan and Putta, Pranav and Yenamandra, Sriram and Henaff, Mikael and Silwal, Sneha and Mcvay, Paul and Maksymets, Oleksandr and Arnaud, Sergio and Yadav, Karmesh and Li, Qiyang and Newman, Ben and Sharma, Mohit and Berges, Vincent and Zhang, Shiqi and Agrawal, Pulkit and Bisk, Yonatan and Batra, Dhruv and Kalakrishnan, Mrinal and Meier, Franziska and Paxton, Chris and Sax, Alexander and Rajeswaran, Aravind},
    title     = {{OpenEQA: Embodied Question Answering in the Era of Foundation Models}},
    booktitle = {Proceedings of the IEEE/CVF Conference on Computer Vision and Pattern Recognition (CVPR)},
    month     = {June},
    year      = {2024},
    pages     = {16488--16498},
    url       = {https://openaccess.thecvf.com/content/CVPR2024/html/Majumdar_OpenEQA_Embodied_Question_Answering_in_the_Era_of_Foundation_Models_CVPR_2024_paper.html}
}

@inproceedings{Jiang2025iccv,
  author    = {Jiang, Kaixuan and Liu, Yang and Chen, Weixing and Luo, Jingzhou and Chen, Ziliang and Pan, Ling and Li, Guanbin and Lin, Liang},
  title     = {{Beyond the Destination: A Novel Benchmark for Exploration-Aware Embodied Question Answering}},
  booktitle = {Proceedings of the IEEE/CVF International Conference on Computer Vision (ICCV)},
  pages     = {9091--9101},
  year      = {2025},
  doi       = {10.1109/ICCV51701.2025.00850}
}

@inproceedings{wang2026lmee,
  title     = {Explore with Long-term Memory: A Benchmark and Multimodal {LLM}-based Reinforcement Learning Framework for Embodied Exploration},
  author    = {Wang, Sen and Liu, Bangwei and Gao, Zhenkun and Ma, Lizhuang and Wang, Xuhong and Xie, Yuan and Tan, Xin},
  booktitle = {Proceedings of the IEEE/CVF Conference on Computer Vision and Pattern Recognition (CVPR)},
  pages     = {37098--37108},
  year      = {2026}
}

@article{koo2025vgavs,
  title   = {Toward Ambulatory Vision: Learning Visually-Grounded Active View Selection},
  author  = {Koo, Juil and Choi, Daehyeon and Youn, Sangwoo and Lee, Phillip Y. and Sung, Minhyuk},
  journal = {arXiv preprint arXiv:2512.13250},
  year    = {2025}
}

@inproceedings{kazemnejad2025vineppo,
  title     = {{VinePPO}: Refining Credit Assignment in {RL} Training of {LLMs}},
  author    = {Kazemnejad, Amirhossein and Aghajohari, Milad and Portelance, Eva and Sordoni, Alessandro and Reddy, Siva and Courville, Aaron and Le Roux, Nicolas},
  booktitle = {International Conference on Machine Learning (ICML)},
  year      = {2025}
}

@article{schulman2017ppo,
  title   = {Proximal Policy Optimization Algorithms},
  author  = {Schulman, John and Wolski, Filip and Dhariwal, Prafulla and Radford, Alec and Klimov, Oleg},
  journal = {arXiv preprint arXiv:1707.06347},
  year    = {2017}
}

@inproceedings{schulman2015trpo,
  title     = {Trust Region Policy Optimization},
  author    = {Schulman, John and Levine, Sergey and Abbeel, Pieter and Jordan, Michael and Moritz, Philipp},
  booktitle = {International Conference on Machine Learning (ICML)},
  pages     = {1889--1897},
  year      = {2015}
}

@inproceedings{langford2007epochgreedy,
  title     = {The Epoch-Greedy Algorithm for Multi-armed Bandits with Side Information},
  author    = {Langford, John and Zhang, Tong},
  booktitle = {Advances in Neural Information Processing Systems (NeurIPS)},
  year      = {2007}
}

@inproceedings{li2010linucb,
  title     = {A Contextual-Bandit Approach to Personalized News Article Recommendation},
  author    = {Li, Lihong and Chu, Wei and Langford, John and Schapire, Robert E.},
  booktitle = {Proceedings of the International Conference on World Wide Web (WWW)},
  pages     = {661--670},
  year      = {2010}
}

@article{bai2025qwen25vl,
  title   = {{Qwen2.5-VL} Technical Report},
  author  = {Bai, Shuai and Chen, Keqin and Liu, Xuejing and Wang, Jialin and Ge, Wenbin and Song, Sibo and Dang, Kai and Wang, Peng and Wang, Shijie and Tang, Jun and Zhong, Humen and Zhu, Yuanzhi and Yang, Mingkun and Li, Zhaohai and Wan, Jianqiang and Wang, Pengfei and Ding, Wei and Fu, Zheren and Xu, Yiheng and Ye, Jiabo and Zhang, Xi and Xie, Tianbao and Cheng, Zesen and Zhang, Hang and Yang, Zhibo and Xu, Haiyang and Lin, Junyang},
  journal = {arXiv preprint arXiv:2502.13923},
  year    = {2025}
}

@inproceedings{liu2024groundingdino,
  title     = {Grounding {DINO}: Marrying {DINO} with Grounded Pre-Training for Open-Set Object Detection},
  author    = {Liu, Shilong and Zeng, Zhaoyang and Ren, Tianhe and Li, Feng and Zhang, Hao and Yang, Jie and Jiang, Qing and Li, Chunyuan and Yang, Jianwei and Su, Hang and Zhu, Jun and Zhang, Lei},
  booktitle = {European Conference on Computer Vision (ECCV)},
  year      = {2024}
}

@inproceedings{ravi2025sam2,
  title     = {{SAM} 2: Segment Anything in Images and Videos},
  author    = {Ravi, Nikhila and Gabeur, Valentin and Hu, Yuan-Ting and Hu, Ronghang and Ryali, Chaitanya and Ma, Tengyu and Khedr, Haitham and R{\"a}dle, Roman and Rolland, Chloe and Gustafson, Laura and Mintun, Eric and Pan, Junting and Alwala, Kalyan Vasudev and Carion, Nicolas and Wu, Chao-Yuan and Girshick, Ross and Doll{\'a}r, Piotr and Feichtenhofer, Christoph},
  booktitle = {International Conference on Learning Representations (ICLR)},
  year      = {2025}
}

@article{tschannen2025siglip2,
  title   = {{SigLIP} 2: Multilingual Vision-Language Encoders with Improved Semantic Understanding, Localization, and Dense Features},
  author  = {Tschannen, Michael and Gritsenko, Alexey and Wang, Xiao and Naeem, Muhammad Ferjad and Alabdulmohsin, Ibrahim and Parthasarathy, Nikhil and Evans, Talfan and Beyer, Lucas and Xia, Ye and Mustafa, Basil and H{\'e}naff, Olivier and Harmsen, Jeremiah and Steiner, Andreas and Zhai, Xiaohua},
  journal = {arXiv preprint arXiv:2502.14786},
  year    = {2025}
}

@article{allen2017mac,
  title   = {Mean Actor Critic},
  author  = {Allen, Cameron and Asadi, Kavosh and Roderick, Melrose and Mohamed, Abdel-rahman and Konidaris, George and Littman, Michael},
  journal = {arXiv preprint arXiv:1709.00503},
  year    = {2017}
}

@inproceedings{ciosek2018epg,
  title     = {Expected Policy Gradients},
  author    = {Ciosek, Kamil and Whiteson, Shimon},
  booktitle = {AAAI Conference on Artificial Intelligence (AAAI)},
  year      = {2018}
}

@inproceedings{tesauro1996rollout,
  title     = {On-line Policy Improvement using {Monte-Carlo} Search},
  author    = {Tesauro, Gerald and Galperin, Gregory R.},
  booktitle = {Advances in Neural Information Processing Systems (NIPS)},
  year      = {1996}
}

@inproceedings{ramakrishnan2021hm3d,
  title     = {{Habitat-Matterport 3D Dataset (HM3D)}: 1000 Large-scale {3D} Environments for Embodied {AI}},
  author    = {Ramakrishnan, Santhosh K. and Gokaslan, Aaron and Wijmans, Erik and Maksymets, Oleksandr and Clegg, Alexander and Turner, John and Undersander, Eric and Galuba, Wojciech and Westbury, Andrew and Chang, Angel X. and Savva, Manolis and Zhao, Yili and Batra, Dhruv},
  booktitle = {NeurIPS Datasets and Benchmarks Track},
  year      = {2021}
}

@inproceedings{savva2019habitat,
  title     = {Habitat: A Platform for Embodied {AI} Research},
  author    = {Savva, Manolis and Kadian, Abhishek and Maksymets, Oleksandr and Zhao, Yili and Wijmans, Erik and Jain, Bhavana and Straub, Julian and Liu, Jia and Koltun, Vladlen and Malik, Jitendra and Parikh, Devi and Batra, Dhruv},
  booktitle = {IEEE/CVF International Conference on Computer Vision (ICCV)},
  year      = {2019}
}
}

\clearpage
\onecolumn
\setcounter{table}{0}
\setcounter{figure}{0}
\setcounter{equation}{0}
\setcounter{algorithm}{0}
\renewcommand{\thetable}{S\arabic{table}}
\renewcommand{\thefigure}{S\arabic{figure}}
\renewcommand{\theequation}{S\arabic{equation}}
\renewcommand{\thealgorithm}{S\arabic{algorithm}}
\renewcommand{\theHtable}{supp.\arabic{table}}
\renewcommand{\theHfigure}{supp.\arabic{figure}}
\renewcommand{\theHequation}{supp.\arabic{equation}}
\providecommand{\theHalgorithm}{}
\renewcommand{\theHalgorithm}{supp.\arabic{algorithm}}
\pdfbookmark[0]{Supplementary Material}{supplement}
\begin{center}
{\Large\bfseries \thetitle\par}
\vspace{0.5em}
{\large Supplementary Material\par}
\end{center}
\appendix

\section*{Appendix Guide}
\phantomsection
\pdfbookmark[0]{Appendix Guide}{appendix-guide}
The linked guide below summarizes the supplementary material.
\begin{itemize}
    \item \hyperref[app:construction]{\textbf{Dataset Construction Details}} (Appendix~\ref*{app:construction}): observation alignment, evidence measurements, memory and QA targets, action trajectories, and question templates.
    \item \hyperref[app:distribution]{\textbf{Training Data Composition and Sampling}} (Appendix~\ref*{app:distribution}): episode counts, training-data sources and composition, and action-length statistics.
    \item \hyperref[app:router]{\textbf{Memory Store Details}} (Appendix~\ref*{app:router}): memory writes, category/time queries, pagination, and image retrieval.
    \item \hyperref[app:training]{\textbf{Training and Evaluation Details}} (Appendix~\ref*{app:training}): optimization, \hyperref[app:action_optimization]{action rewards and rollouts}, \hyperref[app:evaluation]{inference loop, QA scoring, and latency}, and \hyperref[app:timing]{timing evaluation}.
    \item \hyperref[app:test_composition]{\textbf{Test-Set Construction and Composition}} (Appendix~\ref*{app:test_composition}): split rules, answer balancing, evidence-location annotations, failure attribution, and per-template composition.
    \item \hyperref[app:baselines]{\textbf{Baselines and Prompts}} (Appendix~\ref*{app:baselines}): baseline settings, alignment, and the tool-protocol prompt.
    \item \hyperref[app:external]{\textbf{External Benchmarks}} (Appendix~\ref*{app:external}): official protocols and the streaming evaluation on video spatial benchmarks.
    \item \hyperref[app:full_results]{\textbf{Additional Results}} (Appendix~\ref*{app:full_results}): backbones, latency, and reference baselines, the protocol of the question-time retrieval ablation, Memory and Action ablations, seed variation and confidence intervals, answerer robustness, image spatial benchmarks, Memory accuracy by evidence age, and per-template results.
    \item \hyperref[app:visualization]{\textbf{Qualitative Examples}} (Appendix~\ref*{app:visualization}): current-observation, historical-evidence, active-view, and memory-write examples.
\end{itemize}

\section{Dataset Construction Details}
\label{app:construction}
This section details how observations are converted into supervision for memory writing, spatial question answering, and camera actions. We first establish a shared temporal representation, then describe target construction and filtering, and finally specify the action space and multi-step trajectories. The question catalog provides examples and construction rules for individual templates. Figure~\ref{fig:benchmark_construction} of the main paper gives an overview of the pipeline.

\subsection{Observation Preparation and Alignment}
\label{app:alignment}
Simulation episodes use existing 3D-FRONT~\citep{fu2021front} layouts and assets. We remove objects with conspicuous interpenetration or erroneous scale while preserving the remaining scene structure. Scanning trajectories connect geometric centers and selected centers of maximal inscribed circles in randomized order, with the camera oriented toward the room interior. A*~\citep{hart1968astar} plans paths on a mesh-derived occupancy map over the camera's relevant height range, and B\'ezier curves smooth the resulting paths. Multiple trajectories provide different visitation orders and observation histories for each layout. Floor and ceiling textures and randomly selected HDRI environments vary appearance, and Cycles~\citep{blenderCycles} renders the RGB observations and auxiliary visibility annotations.

Real observations originate from ScanNet~\citep{dai2017scannet}, ScanNet++~\citep{yeshwanth2023scannetpp}, and ARKitScenes~\citep{baruch2021arkitscenes} videos. Frames are aligned with camera poses and source 3D annotations through physical timestamps. Coordinate transformations and projection associate annotated objects with individual observations. Both branches provide time-ordered RGB frames, category labels, instance correspondences, geometry, and per-frame visibility. We sample observations at 1~Hz and express geometric distances in meters. Per-frame visibility follows the source annotations. Pixel-level measurements are used where rendered masks are available; projected bounding-box area is separately identified when used as an evidence proxy.

For a question at time $t$, historical evidence is selected from the observation prefix $O_{\leq t}$. An action-generated observation becomes available only after the corresponding command is executed. Offline geometry and instance identifiers determine supervision, whereas the model receives RGB observations, timestamps, retained context, and returned tool content. Each example records its question time, evidence-frame references, answer, and construction provenance.

\subsection{Metadata and Evidence Measurements}
\label{app:evidence_measurements}
Temporal metadata maps frame references to timestamps. Camera metadata contains intrinsics and poses; object metadata supplies persistent identities, categories, 3D centers, dimensions, and available meshes or boxes. Scene metadata supplies geometry, occupancy, and floor polygons. Visibility metadata links these records to individual observations. Distances use meters, image fractions are dimensionless, and each evidence measurement records its provenance.

\paragraph{Visibility and image occupancy.}
Let $I$ be the image rectangle, $N=WH$ its pixel count, $b_o(x)$ the source visibility indicator, and $n_o(x)$ the object's visible mask-pixel count. Pixel visibility is $p_o(x)=n_o(x)/N$. For observations using projected-box evidence, let $R_o(x)$ be the image-plane bounding rectangle of the projected world-axis-aligned 3D box. The occupancy proxy is
\begin{equation}
u_o(x)=b_o(x)\frac{\operatorname{area}(R_o(x)\cap I)}{WH}.
\end{equation}
The indicator gates the proxy, but box area does not measure the unoccluded object surface. Missing source visibility is distinct from an explicit negative observation.

\paragraph{Framing completeness.}
Transform the corners of the mesh-local bounding box to the world and camera frames, project them using camera intrinsics, and form their convex hull $H_o(x)$. Framing completeness is
\begin{equation}
f_o(x)=\frac{\operatorname{area}(H_o(x)\cap I)}{\operatorname{area}(H_o(x))}.
\end{equation}
A valid ratio requires positive projected area and a box entirely beyond the $0.01$~m near plane. Near-plane intersections are retained only as diagnostics, not valid framing scores; fully behind-camera and degenerate projections are unavailable. Framing describes image-boundary clipping, not occlusion by other objects.

\paragraph{Occlusion support.}
An isolated-object render uses the same camera and target geometry while hiding other scene objects. If its target mask contains $n_o^{\mathrm{iso}}(x)>0$ pixels, define
\begin{equation}
v_o^{\mathrm{occ}}(x)=\min\{1,n_o(x)/n_o^{\mathrm{iso}}(x)\}.
\end{equation}
The ratio is unavailable when the isolated mask is missing or empty, or the unbounded ratio exceeds $1.01$. Values up to $1.01$ are capped at one to accommodate small rasterization differences. This measures visible support relative to the isolated in-frame silhouette; framing is checked separately. Required unavailable measurements fail the corresponding quality check rather than being replaced with zeros.

\subsection{Incremental Memory Targets}
\label{app:memory_targets}
Memory targets are generated over consecutive five-second windows. Each target contains one observation record per sampled frame, including frames with no visible objects. A record specifies the frame reference and visible category counts. Optional relations describe image-plane left, right, above, or below relationships, with subjects and objects localized on a $3\times3$ image grid. The final incomplete window is also retained.

Let $V_{f,c}$ denote the set of visible instances of category $c$ in frame $f$. The per-frame count is $|V_{f,c}|$, whereas the historical distinct-instance count is
\begin{equation}
 N_c(t)=\left|\bigcup_{f:\,t_f\leq t}V_{f,c}\right|.
\end{equation}
The latter is computed using offline instance correspondences. For example, observing the same two chairs in three frames yields per-frame counts of two and a historical count of two. Memory targets preserve frame references so that detailed appearance and instance evidence can be recovered from historical images.

\subsection{Spatial QA Construction}
\label{app:spatial_qa}
Each QA example is constructed by selecting a causal observation prefix, binding its referents, computing an answer from aligned annotations, checking evidence and ambiguity conditions, and instantiating a question template. Multiple-choice questions use distinct semantic alternatives and retain the mapping between option letters and answers. Numeric questions specify their units. All QA text undergoes one LLM polishing pass to diversify phrasing while preserving referents, temporal constraints, units, option meanings, and annotation-derived answers.

\paragraph{Current QA.}
Current questions use the latest observation or an explicitly specified recent window, with a default duration of five seconds. Camera-to-object and object-to-object distances use 3D centers. Trajectory length sums consecutive camera translations, endpoint displacement measures the distance between the initial and final positions, and orientation change measures the net endpoint rotation. Referents must be uniquely identifiable under the question wording. Nearest-camera queries require a minimum $0.15$~m separation between the two closest candidates; designated-object nearest-neighbor queries require $0.40$~m. Appearance and disappearance questions require a unique category change between the specified observations.

\paragraph{Memory QA.}
Memory questions require evidence preceding the current window. First and last appearances are computed over the full causal prefix, historical counts use instance-set unions, and historical relationships are evaluated at the referenced frame. Appearance-order questions require successive first-observation times to differ by at least two seconds. Historical-neighbor questions bind a uniquely observed target that is absent from the current view and select neighbors among objects co-visible at the specified historical time. These evidence conditions define the question family; the model can answer directly when its retained context already contains the required information.

\paragraph{Chained questions and geometric examples.}
For linked questions, the first turn establishes an object or category and subsequent turns retain that referent. These linked QA turns are distinct from multiple camera actions taken to answer a single question. Each turn is associated with its own arrival time and evidence prefix. Height questions compare vertical extent rather than elevation above the floor, with a $0.30$~m margin between the two tallest candidates. Room-area questions use the room's floor polygon, including floor beneath furniture, and require at least eight observations, $90\%$ boundary-angular coverage, and a viewing-direction span of at least $150^\circ$.

\paragraph{A construction example.}
Consider a nearest-object question with visible object-center distances of $1.0$, $1.4$, and $2.1$~m. The first object supplies the answer because its $0.4$~m lead exceeds the $0.15$~m ambiguity margin. Replacing the second distance with $1.08$~m rejects the example. For a first-appearance question, the answer is the earliest visible frame in the complete prefix, and the example is assigned to Memory QA when that evidence precedes the current window.

\subsection{Selecting Question Times and Object Referents}
\label{app:question_times}
Construction operates on aligned frame records rather than RGB alone. Each record links a timestamp and image to its camera pose, visible-instance set, and category labels. Object records supply persistent instance identifiers, 3D centers, and the geometric fields required by each template; image-plane relations additionally use projected positions, and room-area questions use floor polygons. These annotations select candidates and compute labels offline. Public questions identify objects through category names, explicit nearest-instance qualifiers, or image regions, without exposing instance identifiers or geometric answers.

\paragraph{Candidate times and evidence windows.}
We enumerate candidate times on the sampled observation sequence and apply template-specific temporal conditions before constructing a question. Single-view questions use the question frame; motion and visibility-change questions require the specified earlier endpoint. Nearest-neighbor, fixed-reference relation, and height-comparison templates enumerate interior frames. Last-seen, historical coappearance, historical direction, and appearance-order templates are proposed every five sampled frames starting at index ten. Memory evidence must belong to the available prefix and be covered by completed memory writes. Room-area candidates are proposed when a continuous within-room observation run meets the coverage conditions. A frame that does not satisfy a template's conditions produces no example for that template.

\paragraph{Object eligibility and combinations.}
For nearest-neighbor, fixed-reference relation, and height-comparison templates, candidate objects are visible at the question frame and its two adjacent frames, and each selected category has exactly one visible instance at the question frame. The adjacent-frame test selects stable observations offline; the following frame is not included in the question's model input or used to compute its geometric answer. Category-conflict checks remove semantically overlapping options. We then enumerate reference--target pairs for relations and nearest-neighbor questions, or four distinct-category objects for height comparison. Missing geometric fields remove the affected combination. Other templates retain their own bindings: a nearest-instance qualifier can resolve multiple objects of one category, whereas historical single-instance questions require a unique previously observed instance.

\paragraph{Answer computation and acceptance.}
For a bound combination, we first compute the geometric or temporal quantity, then check the template's uniqueness and evidence conditions. A designated-object nearest-neighbor question requires a $0.40$~m gap between its two closest eligible neighbors. Height comparison uses world-vertical bounding-box extent and requires a $0.30$~m gap between the two tallest objects. Fixed-reference left/right questions use the $3\times3$ image grid: the correct target lies in the reference object's row and on the requested side; distractors come from the opposite side or categories absent from the scene. The category alternatives are filtered before option letters are assigned. These conditions apply to the corresponding templates, rather than serving as universal thresholds for all QA.

\paragraph{Historical selection and memory verification.}
The last-seen template selects a category absent at the question frame whose latest occurrence is at least five frame indices earlier. That historical frame also supplies candidates for coappearance and relative-direction questions, with its timestamp stated in the question. Coappearance distractors are absent from the referenced frame. Direction questions require unique category instances and a same-row or same-column grid relation. First-appearance-order questions select three categories whose first sightings occur at least two seconds apart, with the latest first sighting at least five frame indices before the question. We replay the corresponding teacher memory writes and retrieval requests, reconstruct the answer from returned records or relations, and require agreement with the annotation-derived label. Direction examples additionally require the referenced image to be retrievable within the image budget.

\paragraph{Room evidence and retained records.}
Room-area construction uses simulation episodes with a valid floor polygon and unambiguous camera-to-room assignment. Leaving the assigned room resets the observation run. Each polygon edge contributes eight boundary samples; coverage counts samples falling within the horizontal viewing angle of at least one observation in the run. We require eight observations, $90\%$ angular boundary coverage, and a $150^\circ$ heading span. This is a geometric coverage proxy rather than a wall-pixel visibility test. Each retained QA records its template, question time, evidence-frame references, bound objects where applicable, answer, and source-scene provenance. Missing evidence images and failed template checks are excluded before serialization.

\paragraph{Worked current and memory examples.}
Consider an illustrative current nearest-neighbor example at $t=12$~s. A cabinet, bed, chair, and lamp have unique category referents and pass the adjacent-frame visibility check. Their center distances from the cabinet are $1.0$, $1.6$, and $2.2$~m. The bed is retained as the answer because the nearest-neighbor margin is $0.6$~m; the question refers to the cabinet, and its evidence is the frame at $12$~s. If the chair were $1.2$~m away, the combination would fail the $0.40$~m margin. For an illustrative memory example at $t=20$~s, a lamp was last visible at $8$~s and is absent afterward. Its last-seen question has answer $8$~s, with the frame at $8$~s as positive evidence and the intervening prefix establishing that no later sighting occurred. The teacher write covering that frame is replayed, and a lamp-category query must recover the same latest timestamp. A later lamp sighting would change the label rather than preserve the earlier answer.

\subsection{Action-conditioned Data}
\label{app:action_data}
  Action data are constructed through dedicated rendering of one- to three-step discrete camera-action sequences. Each example fixes the starting observation, question, target objects,
  and geometry-derived answer. Every action is executed from the preceding camera state and produces a newly rendered observation under unchanged scene geometry, appearance, and
  imaging settings. Evidence and geometric checks select demonstration sequences; alternative action outcomes are rendered and recorded separately for counterfactual optimization.

\paragraph{Start times and target binding.}
 At candidate question times in simulated observation streams, we select objects with available geometry, a resolvable initial-view referent, and a visible-mask fraction of at least $0.002$ of the image.
 Referents use a unique category or a unique category within a $3\times3$ image region. Up to sixteen objects are retained in descending score order. Each object supplies a size candidate, and each unordered pair supplies center-distance and distance-range candidates. Size is the longest physical bounding-box side, using source physical dimensions when available or transformed mesh dimensions otherwise; pair distance is the Euclidean distance between fixed 3D centers. The same bound objects and answer are retained throughout the path.

  \paragraph{Evidence measurements and endpoint selection.}
  Action construction uses mask-derived visible-pixel fractions for both pre-question observations and newly rendered action views. Corresponding target masks are required to evaluate the historical baseline, ensuring that evidence gains compare the same measurement. Let $p_j(x)$, $f_j(x)$, and $v_j^{\mathrm{occ}}(x)$ denote visible-pixel fraction, framing completeness, and occlusion support, respectively, as defined in Appendix~\ref{app:evidence_measurements}. For the required objects $\mathcal J_q$, we use the view quality $E(x,q)$ of the Stage 2 reward (Appendix~\ref{app:action_optimization}) and its best value over the entire pre-question observation history $O_{\mathrm{pre}}=O_{\leq t_q}$:
  \begin{equation}
  E(x,q)=\min_{j\in\mathcal J_q}\min\{p_j(x)/0.10,1\}\,f_j(x),
  \qquad
  \Phi_{\mathrm{pre}}(q)=\max_{x\in O_{\mathrm{pre}}}E(x,q).
  \end{equation}
  The minimum requires joint visual support for pairwise questions, while the historical maximum prevents improvement from being measured only against a weak starting view. If $\Phi_{\mathrm{pre}}(q)\geq0.2$, the best pre-question view is already adequate and the example answers directly. If $\Phi_{\mathrm{pre}}(q)\leq0.05$, that is, no pre-question view meets the prescribed evidence-quality criterion, an action demonstration is constructed, and an endpoint $x$ is retained only if
  \begin{equation}
  E(x,q)\geq0.2,\qquad
  E(x,q)-\Phi_{\mathrm{pre}}(q)\geq0.10,\qquad
  v_j^{\mathrm{occ}}(x)\geq0.70\quad\forall j\in\mathcal J_q.
  \end{equation}
  Required unavailable measurements fail the check. Questions with no such endpoint within three steps, and those with $0.05<\Phi_{\mathrm{pre}}(q)<0.2$, yield no action demonstration. The test strata (Appendix~\ref{app:test_composition}) and the premature-stop category of failure attribution share this view-quality definition and the base thresholds $0.05$ and $0.2$, and the Stage 2 reward uses the same score. The endpoint-gain and occlusion-support conditions above apply only to training demonstrations; the test strata do not use them.

\paragraph{Discrete action interface and multi-step interaction.}
The action interface consists of five commands: forward, leftward, and rightward translation by $0.5$~m, and left and right rotation by $45^\circ$. Translation follows the current camera's horizontal coordinate frame and preserves orientation; rotation preserves position. The commands target the three causes of insufficient evidence: lateral moves shift the viewpoint around occluders, turns bring truncated targets back into the frame, and forward moves enlarge small targets. The step sizes change the view noticeably in one step while keeping the target nearby, and three attempts bound the interaction to local adjustment. There is no backward command: the stream mostly moves forward, so views from behind the current position usually already exist in the history, where they count toward the best pre-question view and can be retrieved. Long-range navigation and object search are deliberately excluded. The model selects a command, and the runtime assigns its fixed magnitude. Each returned observation is associated with the resulting camera state.   Interactions allow at most three action attempts, each continuing from its predecessor state and producing a specifically rendered observation. Demonstrations terminate at the
  earliest endpoint satisfying the endpoint criteria above. Intermediate views need not satisfy the endpoint thresholds or improve evidence monotonically, allowing
  preparatory actions.

\paragraph{Geometric checks and serialization.}
For each segment, a mesh-based check rejects a surface intersection along the straight camera-center connection. Positions sampled at intervals no greater than $0.025$~m require at least $0.125$~m surface clearance and support within $2$~m below. The serialized example contains the causal prefix, unchanged question, ordered command--image pairs, and final answer; future trajectory images enter the model context only after their corresponding commands. Duplicate candidates sharing the source trajectory, bound target or pair, endpoint, and template are removed. The distance-range template has two phrasings that share the same geometric quantity; identical serialized training sequences are kept once and carry the distance-range label.

  \paragraph{Rendered alternatives for counterfactual optimization.}
  Selected action sequences provide SFT demonstrations. Bandit-based counterfactual optimization additionally requires alternative outcomes for the five discrete commands at each evaluated state,
  rendered under the same scene and imaging settings. Demonstration endpoint thresholds do not filter these alternatives: valid actions with little or negative evidence improvement
  remain available for comparison, and physically invalid actions are recorded separately. Branch validation distinguishes missing rendered outcomes (coverage errors) from physically invalid actions, which preserve the observation and consume an attempt. RL
  rewards use the mask-derived measurements defined in Appendix~\ref{app:action_optimization}.

  \paragraph{Worked three-step example.}
  Consider a size question about a uniquely identified cabinet whose longest physical side is $1.2$~m. At the question time, its best pre-question view has visible-pixel fraction $0.004$ and framing completeness $0.60$, because the image border cuts it, so $\Phi_{\mathrm{pre}}=0.04\times0.60=0.024\leq0.05$ and an action demonstration is constructed. Three successive rendered views yield visible-pixel fractions of $0.006$, $0.015$, and $0.030$ with framing completeness $0.70$, $0.85$, and $0.90$, \ie, evidence scores of $0.042$, $0.128$, and $0.270$. The first two are below $0.2$. The third is adequate and exceeds the pre-question best by $0.246$; with occlusion support $0.80$, it satisfies all endpoint conditions. If all transitions pass geometric checks, we retain the causal prefix,
  fixed question, three discrete command--image pairs, and answer $1.2$~m. A failed transition rejects this sequence; if an earlier view satisfies every condition, the demonstration
  ends there instead.

\subsection{Assembly and Quality Control}
We assemble observations, memory targets, questions, and tool outcomes into causally ordered independent streams of at most 180~s. The training release contains all retained QA examples; the main test and Timing sets are separate evaluation collections. Validation checks timestamps, frame references, referent consistency, answer computations, duplicate semantic questions, tool-image correspondence, and interaction budgets. Examples with missing required evidence or failed constraints are excluded. Public model inputs and private annotation fields are stored separately.

\subsection{Question Catalog and Construction Examples}
\label{app:catalog}
Current QA has 18 templates (14 multiple-choice with 2, 4, or 5 options and 4 numerical) from real and simulated streams, with 302,661 training and 9,147 test QA turns. Memory QA has 10 templates (7 record-answerable and 3 frame-required; 6 multiple-choice with 4 or 5 options and 4 numerical) from real and simulated streams, with 402,662 training and 7,964 test turns. Action QA has 3 templates (1 multiple-choice with 4 options and 2 numerical) from simulated streams only, with 294,677 training and 1,051 test turns.

Tables~\ref{tab:current_templates}, \ref{tab:memory_templates}, and~\ref{tab:action_templates} list the Current, Memory, and Action QA templates, respectively, with representative questions, semantic answers, and construction rules. Current templates use present or recent evidence; Memory templates require historical observations; Action templates specify questions supported by newly acquired views. For multiple-choice examples, we display the semantic answer rather than its option letter. Linked templates show the opening question and specify the subsequent turn in the construction rule.
\begin{longtable}{p{0.20\textwidth}p{0.72\textwidth}}
\caption{Current QA templates and examples.}\label{tab:current_templates}\\
\toprule Template & Example and construction\\\midrule\endfirsthead
\toprule Template & Example and construction\\\midrule\endhead
\bottomrule\endfoot
approached object category & \textbf{Question:} What is the category whose object had the largest decrease in camera-to-3D-center distance among objects visible both now and 5 seconds ago? Choose from the listed options.\newline
\textbf{Answer:} storage unit.\newline
\textbf{Construction:} Compare objects visible at both endpoints. Require a distance decrease of at least 0.20 m and a 0.10 m lead over the runner-up.\\[4pt]
became visible category & \textbf{Question:} Select the option that identifies the category that became visible between 5 seconds ago and now.\newline
\textbf{Answer:} plant.\newline
\textbf{Construction:} Take the difference between endpoint category sets and require one newly visible category.\\[4pt]
camera motion bucket & \textbf{Question:} What is the range of the camera's straight-line endpoint displacement over the last 5 seconds? Choose from the listed options.\newline
\textbf{Answer:} almost stationary (at most 0.2 meters).\newline
\textbf{Construction:} Discretize straight-line camera endpoint displacement, excluding ambiguous interval boundaries.\\[4pt]
camera path length m & \textbf{Question:} Estimate the total length of the camera's trajectory over the last 5 seconds. Answer in meters.\newline
\textbf{Answer:} 0.0.\newline
\textbf{Construction:} Sum Euclidean translations between consecutive camera positions in the specified window.\\[4pt]
camera rotation deg & \textbf{Question:} Provide your numerical estimate of the net difference between the camera's starting and ending orientations over the last 5 seconds. Answer in degrees.\newline
\textbf{Answer:} 120.0.\newline
\textbf{Construction:} Compute net rotation between endpoint camera orientations, expressed in degrees.\\[4pt]
camera to object distance bucket & \textbf{Question:} What is the camera-to-3D-center distance range for the closest visible sofa? Choose from the listed options.\newline
\textbf{Answer:} very near (at most 1.25 meters).\newline
\textbf{Construction:} Bind the nearest visible instance of the named category and discretize its center distance to the camera.\\[4pt]
camera to object distance m & \textbf{Question:} Give a numerical value for the camera-to-3D-center distance to the closest visible sofa. Answer in meters.\newline
\textbf{Answer:} 1.06.\newline
\textbf{Construction:} Bind the nearest visible instance of the named category and compute its center distance to the camera.\\[4pt]
closer than before & \textbf{Question:} Select the option that identifies the change in camera-to-3D-center distance to the table over the last 5 seconds, considering the object visible at both times.\newline
\textbf{Answer:} much closer (decrease of at least 0.75 meters).\newline
\textbf{Construction:} Uniquely match the category referent across the two observations and classify its camera-distance change.\\[4pt]
closest category then distance & \textbf{Question:} Choose the answer corresponding to the category of the currently visible object closest to the camera by 3D center distance.\newline
\textbf{Answer:} sofa.\newline
\textbf{Construction:} First identify the nearest visible category; then query the distance interval of the same object. Count both QA turns.\\[4pt]
closest category then relation & \textbf{Question:} Determine the category of the currently visible object closest to the camera by 3D center distance and select the matching choice.\newline
\textbf{Answer:} sofa.\newline
\textbf{Construction:} First bind the nearest visible object; then identify its nearest neighbor among currently co-visible objects.\\[4pt]
closest neighbor category & \textbf{Question:} Which option describes the listed category whose currently visible object has its 3D center closest to the cabinet?\newline
\textbf{Answer:} bed.\newline
\textbf{Construction:} Require visibility at the question frame and its adjacent frames for offline stability selection, and unique category referents; the nearest two candidates must differ by at least 0.40 m.\\[4pt]
closest object category & \textbf{Question:} Choose the answer corresponding to the category of the currently visible object closest to the camera by 3D center distance.\newline
\textbf{Answer:} sofa.\newline
\textbf{Construction:} Rank currently visible objects by camera-to-center distance; require a nearest-candidate margin of 0.15 m.\\[4pt]
disappeared category & \textbf{Question:} From the options below, indicate the category that disappeared from view between 5 seconds ago and now.\newline
\textbf{Answer:} sofa.\newline
\textbf{Construction:} Take the reverse endpoint category-set difference and require one disappeared category.\\[4pt]
fixed reference relation & \textbf{Question:} From the options below, indicate the listed category whose currently visible object is to the left of the desk in the image at the time of this question.\newline
\textbf{Answer:} double bed.\newline
\textbf{Construction:} Use projected image positions with unique referents; select a same-row left/right relation and place distractors on the opposite side.\\[4pt]
multi candidate height & \textbf{Question:} Which option describes the listed category whose currently visible object has the greatest vertical height (not elevation above the floor)?\newline
\textbf{Answer:} wardrobe.\newline
\textbf{Construction:} Compare world-vertical bounding-box extents of four distinct-category candidates; require a 0.30 m gap between the tallest two.\\[4pt]
object category then later visibility & \textbf{Question:} Determine the category of the currently visible object closest to the camera by 3D center distance and select the matching choice.\newline
\textbf{Answer:} media unit.\newline
\textbf{Construction:} Bind a category/object in the first turn and ask about its visibility only after the later observation arrives.\\[4pt]
object relative to camera & \textbf{Question:} Which option describes the direction of the closest visible bed relative to the camera now?\newline
\textbf{Answer:} far to the right.\newline
\textbf{Construction:} Transform the bound object center into the current camera frame and assign the corresponding direction category.\\[4pt]
room floor area & \textbf{Question:} Calculate the total floor-footprint area of the room just surveyed that the camera is currently inside, including floor beneath furniture from the observations. Answer in square meters.\newline
\textbf{Answer:} 8.43.\newline
\textbf{Construction:} Compute the valid room floor-polygon area. Require at least eight frames, 90 percent boundary-angular coverage, and a 150-degree heading span.\\[4pt]
\end{longtable}
\begin{longtable}{p{0.20\textwidth}p{0.72\textwidth}}
\caption{Memory QA templates and examples.}\label{tab:memory_templates}\\
\toprule Template & Example and construction\\\midrule\endfirsthead
\toprule Template & Example and construction\\\midrule\endhead
\bottomrule\endfoot
first appearance order & \textbf{Question:} Identify the first-observation order of the categories armchair, dining table, storage unit since this episode began by selecting an option.\newline
\textbf{Answer:} storage unit $\rightarrow$ dining table $\rightarrow$ armchair.\newline
\textbf{Construction:} Sort three category-first-observation times, separated by at least two seconds and preceding the current window.\\[4pt]
first seen category then timestamp & \textbf{Question:} Determine the category seen earliest among the listed choices so far in this episode and select the matching choice.\newline
\textbf{Answer:} lighting.\newline
\textbf{Construction:} Identify the earliest category among the options, then query its original first-observation timestamp.\\[4pt]
first seen object category & \textbf{Question:} Among the choices, which one gives the category seen earliest among the listed choices so far in this episode?\newline
\textbf{Answer:} lighting.\newline
\textbf{Construction:} Compare first-observation times of the listed categories; require a unique earliest option before the current window.\\[4pt]
first seen timestamp s & \textbf{Question:} Estimate the first-seen timestamp of the lighting category, measured from the start of this episode. Answer in seconds.\newline
\textbf{Answer:} 0.0.\newline
\textbf{Construction:} Return the earliest observed timestamp for the category over the complete causal prefix.\\[4pt]
historical coappearance category & \textbf{Question:} Choose the answer corresponding to the listed category also visible alongside the lighting at timestamp 3 seconds.\newline
\textbf{Answer:} cabinet.\newline
\textbf{Construction:} Use the explicitly referenced historical frame; the correct category is co-visible with the anchor and distractors are absent.\\[4pt]
historical relative direction & \textbf{Question:} From the options below, indicate the image-plane position of the storage unit relative to the table in the observation at timestamp 44 seconds.\newline
\textbf{Answer:} left.\newline
\textbf{Construction:} Use uniquely bound objects in the historical frame and their same-row or same-column image-plane relationship.\\[4pt]
last seen timestamp & \textbf{Question:} Report the last timestamp at which the lighting category was observed before this question as a number. Answer in seconds.\newline
\textbf{Answer:} 44.0.\newline
\textbf{Construction:} Take the last occurrence before the question; require current absence and an occurrence preceding the pending write window.\\[4pt]
seen count by category & \textbf{Question:} Report the number of distinct table objects observed so far in this episode as a number. Answer with an integer count (objects).\newline
\textbf{Answer:} 4.\newline
\textbf{Construction:} Union visible instance IDs over the prefix; require historical write coverage and that the current view does not contain all counted instances.\\[4pt]
single seen object then closest category & \textbf{Question:} Choose the answer corresponding to the category nearest to that lighting by 3D center distance, among the other objects visible at timestamp 0 seconds alongside the lighting seen earlier and absent now.\newline
\textbf{Answer:} table.\newline
\textbf{Construction:} Bind the only previously observed instance of the target category, absent now; find its nearest co-visible historical neighbor.\\[4pt]
single seen object then closest distance & \textbf{Question:} What is the 3D center distance from that lighting to its nearest neighbor, among the other objects visible at timestamp 0 seconds alongside the lighting seen earlier and absent now? Answer in meters.\newline
\textbf{Answer:} 0.776.\newline
\textbf{Construction:} Use the same historical target binding and return its minimum center distance to another object visible in the referenced frame.\\[4pt]
\end{longtable}
\begin{longtable}{p{0.20\textwidth}p{0.72\textwidth}}
\caption{Action QA templates and examples.}\label{tab:action_templates}\\
\toprule Template & Example and construction\\\midrule\endfirsthead
\toprule Template & Example and construction\\\midrule\endhead
\bottomrule\endfoot
object size m & \textbf{Question:} Estimate the longest physical side length of the 3D bounding box of the drawer chest / corner cabinet visible in the initial view. Answer in meters.\newline
\textbf{Answer:} 1.185.\newline
\textbf{Construction:} Use source physical dimensions or transformed mesh bounding-box dimensions to obtain the longest physical side in meters; hold the bound object fixed across views.\\[4pt]
object to object distance bucket & \textbf{Question:} Identify the 3D center-to-center distance range between the sofa in the middle left of the initial view and the table in the lower left of the initial view by selecting an option.\newline
\textbf{Answer:} near (1 to 1.75 meters).\newline
\textbf{Construction:} Map the bound pair's center distance to an unambiguous interval; require evidence for both objects.\\[4pt]
object to object distance m & \textbf{Question:} Report the 3D center-to-center distance between the sofa in the lower left of the initial view and the chair visible in the initial view as a number. Answer in meters.\newline
\textbf{Answer:} 3.041.\newline
\textbf{Construction:} Compute the Euclidean distance between the bound 3D centers; retain the same pair throughout the action path.\\[4pt]
\end{longtable}

\section{Training Data Composition and Sampling}
\label{app:distribution}
We report source composition, supervision counts, and action-path statistics separately. A QA turn is one question--answer pair; a linked two-turn example contributes two QA turns. Memory writes count completed supervision windows, while camera-action counts refer to executed transitions. Unique questions, demonstration paths, and alternative transitions have separate denominators. Demonstration length counts executed camera actions.

Sampling is organized hierarchically by QA family, semantic template, and source scene. Within each family, we balance templates over eligible examples and preserve scene diversity. Linked question groups are selected together, and memory-writing targets remain attached to their causal windows. Dataset composition and training exposure are reported separately: sampling an example multiple times does not increase the number of distinct QA examples.

The training release contains 1,000,000 QA answer turns: 302,661 Current QA, 402,662 Memory QA, and 294,677 Action QA. All of these turns are training data. MemAct-Bench uses a separate 18,162-turn main QA test set and an additional 2,000-question Timing set (Appendix~\ref{app:timing}). Training and test sets are disjoint at the level of physical spaces; the split unit for each source is given in Appendix~\ref{app:test_composition}. The 1,000,000 training QA turns are serialized into 342,868 independent streams covering 11,724 source-qualified scene identifiers (4,835 simulated layouts and 6,889 real recordings). Each stream spans at most 180~s at 1~FPS, starts from an empty context and memory, and is processed under the stream protocol of Section~\ref{sec:method_loop}: every supervised output is trained on its causal prefix after the inference-time context-management rule is applied (Appendix~\ref{app:training}). Table~\ref{tab:data_composition} breaks down training answer turns by QA family, while Table~\ref{tab:source_composition} separates their sources and shows that Action QA is simulation-only. Table~\ref{tab:action_composition} reports demonstration lengths; one-action paths are most common, with two- and three-action paths providing multi-step supervision. Table~\ref{tab:training_exposure} reports training-set fractions, rather than repeated exposure during optimization, and Table~\ref{tab:template_distribution} gives the finer template-level counts, including linked answer turns.

\subsection{Episode Composition and Temporal Segmentation}
\label{app:episode_composition}
For 3D-FRONT, an episode is an observation sequence associated with a layout, camera trajectory, and appearance configuration. Current and Memory QA use 14,534 simulated episodes from 4,823 layouts. Action QA uses 3,393 layouts for SFT and 901 layouts for Stage 2, 890 of which are shared with SFT. Twelve of the Action layouts do not appear among the 4,823 Current and Memory layouts, so the training release covers 4,835 simulated layouts. The Action test set uses 255 further layouts, none of which appears in any training set (Appendix~\ref{app:test_composition}). Action observations are not counted as independent episodes in Table~\ref{tab:episode_composition}.

Long real-world recordings are divided into shorter temporal segments. The final release uses 8,732 unique segments from 6,889 original recordings; recording counts refer to source recordings, not to the processed streams of at most 180~s that the model receives. A recording is counted when at least one segment is retained, without requiring its full duration to be used. All retained segments reported here belong to the training collection. These recording identities are distinct from episode IDs and serialized training streams. Together, 4,835 simulated layout IDs and 6,889 real recording IDs account for the 11,724 source-qualified scene identifiers reported above. These identifiers count recordings, not physical spaces: several recordings can depict the same space (\eg, repeated ScanNet scans of one room). They are therefore not the unit of the train/test split, which groups all recordings of a physical space on the same side (Appendix~\ref{app:test_composition}).

\begin{table}[t]
\centering\small
\caption{Episode composition. The simulation row covers Current and Memory QA; real-world rows count original recordings and unique retained segments.}
\label{tab:episode_composition}
\begin{tabular}{lrr}
\toprule
Source & Layouts / recordings & Episodes / segments\\
\midrule
3D-FRONT & 4,823 & 14,534\\
\midrule
ScanNet & 1,412 & 1,511\\
ScanNet++ & 796 & 1,457\\
ARKitScenes & 4,681 & 5,764\\
\midrule
Real-world total & 6,889 & 8,732\\
\bottomrule
\end{tabular}
\end{table}

\begin{table}[t]
\centering\small
\caption{Training QA composition by family. Counts denote answer turns.}
\label{tab:data_composition}
\begin{tabular}{lr}
\toprule
Family & Training QA turns\\
\midrule
Current & 302,661\\
Memory & 402,662\\
Action & 294,677\\
Total & 1,000,000\\
\bottomrule
\end{tabular}
\end{table}

\begin{table}[t]
\centering\small
\caption{Training QA composition by source.}
\label{tab:source_composition}
\begin{tabular}{lrrr}
\toprule
Source & Current & Memory & Action\\
\midrule
3D-FRONT & 135,671 & 218,983 & 294,677\\
ScanNet & 31,304 & 30,431 & 0\\
ScanNet++ & 19,160 & 29,201 & 0\\
ARKitScenes & 116,526 & 124,047 & 0\\
\bottomrule
\end{tabular}
\end{table}

\begin{table}[t]
\centering\small
\caption{Training action demonstration lengths after selection.}
\label{tab:action_composition}
\begin{tabular}{lr}
\toprule
Actions & Training QA turns\\
\midrule
1 & 216,418\\
2 & 46,858\\
3 & 31,401\\
\bottomrule
\end{tabular}
\end{table}

\begin{table}[t]
\centering\small
\caption{Training-set composition by QA family and source. Each fraction uses the 1,000,000 training QA turns as its denominator.}
\label{tab:training_exposure}
\begin{tabular}{lrr}
\toprule
Group & QA turns & Fraction\\
\midrule
Current & 302,661 & 30.27\%\\
Memory & 402,662 & 40.27\%\\
Action & 294,677 & 29.47\%\\
ARKitScenes & 240,573 & 24.06\%\\
ScanNet & 61,735 & 6.17\%\\
ScanNet++ & 48,361 & 4.84\%\\
3D-FRONT & 649,331 & 64.93\%\\
\bottomrule
\end{tabular}
\end{table}

\begin{longtable}{p{0.68\textwidth}r}
\caption{Template counts in the training release. Linked templates count all retained answer turns.}\\
\label{tab:template_distribution}\\
\toprule Template & Training QA turns\\\midrule\endfirsthead
\toprule Template & Training QA turns\\\midrule\endhead
\bottomrule\endfoot
\multicolumn{2}{l}{\textbf{Current QA}}\\
approached object category & 18,733\\
became visible category & 18,733\\
camera motion bucket & 18,733\\
camera path length m & 18,733\\
camera rotation deg & 18,733\\
camera to object distance bucket & 18,733\\
camera to object distance m & 18,733\\
closer than before & 18,733\\
closest category then distance & 18,732\\
closest category then relation & 18,732\\
closest neighbor category & 18,739\\
closest object category & 18,732\\
disappeared category & 18,732\\
fixed reference relation & 1,618\\
multi candidate height & 1,316\\
object category then later visibility & 18,732\\
object relative to camera & 18,732\\
room floor area & 18,732\\
\multicolumn{2}{l}{\textbf{Memory QA}}\\
first appearance order & 61,034\\
first seen category then timestamp & 36,870\\
first seen object category & 18,435\\
first seen timestamp s & 18,435\\
historical coappearance category & 60,843\\
historical relative direction & 26,743\\
last seen timestamp & 60,847\\
seen count by category & 24,399\\
single seen object then closest category & 47,528\\
single seen object then closest distance & 47,528\\
\multicolumn{2}{l}{\textbf{Action QA}}\\
object size m & 128,469\\
object to object distance bucket & 70,293\\
object to object distance m & 95,915\\
\end{longtable}

\section{Memory Store Details}
\label{app:router}
The Memory Store is the part of the tool runtime that maintains model-written records and all past frames and retrieves historical evidence through deterministic rules; it has no learned parameters. Its responsibilities are structural validation (Update, Section~\ref{sec:method_memory}), persistent storage, category/time filtering, and budgeted image return. The policy supplies the semantic content of memory writes and formulates retrieval requests.

\subsection{Budgets}
Table~\ref{tab:budgets} lists the budgets of Section~\ref{sec:method_overview}. Retrieval calls, historical images, and camera attempts are per-question tool budgets that reset when a new question arrives. The context budget applies to the complete active context throughout a stream, and the active context and the external memory persist within the stream; a new stream reinitializes both. With the camera disabled (Section~\ref{sec:exp_action}), camera commands are unavailable from the start; this differs from exhausting the attempt budget, and retrieval remains available.
\begin{table}[h]
\centering\small
\caption{Budgets. The first three rows are per-question tool budgets; the context row applies to the active context throughout a stream.}
\label{tab:budgets}
\begin{tabular}{llll}
\toprule
Budget & Initial value & Consumed by & When exhausted \\
\midrule
Retrieval calls & 4 & each \texttt{use\_memory} request & notice; final answer requested \\
Historical images & 4 distinct, $\leq$2 per call & each newly returned frame & frame references returned as text \\
Camera attempts & 3 & each command, including invalid ones & notice; final answer requested \\
Context $C_t$ & $L=16{,}384$ tokens & \begin{tabular}[t]{@{}l@{}}text and visual tokens in the\\active context, including\\question-time interactions\end{tabular} & \begin{tabular}[t]{@{}l@{}}visual tokens of the oldest\\8 frames evicted; text kept\\within the stream\end{tabular} \\
\bottomrule
\end{tabular}
\end{table}

\subsection{Writing and Storage}
A memory write uses the fields \texttt{type}, \texttt{observations}, and \texttt{relations}. Observation entries contain a frame reference and a list of category--count pairs. Update requires contiguous half-open windows $[t_s,t_e)$, exactly one record per supplied frame, normalized lowercase category strings, positive integer counts, and valid relation endpoints. Empty object lists represent frames without visible objects. A valid write is stored with its frame/time mapping and a payload receipt. Repeating an identical write returns \texttt{already\_stored}; a different payload for the same window is rejected as a conflicting retry. Semantic predictions are stored as generated by the model. After a rejection, the model may retry at most twice with the returned error message; if both retries fail, the window is skipped and has no record. MemAct-9B accepts 99.43\% of its write windows on MemAct-Bench within two retries.

\subsection{Queries, Pagination, and Images}
Each question allows at most four memory-retrieval calls. A retrieval request specifies categories, an optional time range, and whether images are required. Category matching returns the union of matching records, with an additional co-visibility indicator for records containing all requested categories; an empty category list selects all categories. Results are ordered by time and paginated at eight records per page. A continuation cursor is bound to the original query and its record snapshot.

When images are requested, the Memory Store selects one representative per contiguous matching run, preferring co-visible records and the timestamp nearest their midpoint. Additional candidates are ordered by the summed counts of requested categories. Each call returns at most two images, with at most four distinct historical images per question. Returned images remain in context and are not resent. New questions reset pagination and image budgets; textual frame references remain available after the image budget is exhausted. These limits concern retrieved history, not action observations or the entire context.

Relations contain a frame, a predicate, and subject/object endpoints specified by category and image-grid region. Query time ranges are half-open. Category filtering selects frame records, and the response includes all relations attached to frames in the returned text page or image list. It does not independently filter relation endpoints by query category.

\paragraph{Illustrative protocol.}
The following payload illustrates a single-frame write at $t=1$~s. A complete write contains one observation entry for every supplied frame in its window. The chair query returns this frame's category counts and the chair--table relation, together with the image if budget permits.
\begin{verbatim}
{"type":"memory_write",
 "observations":[{"frame":"f000001","objects":[
   {"category":"chair","count":1},
   {"category":"table","count":1}]}],
 "relations":[{"frame":"f000001",
   "subject":{"category":"chair","region":"middle_left"},
   "predicate":"left_of_in_image",
   "object":{"category":"table","region":"middle_right"}}]}
{"type":"use_memory","categories":["chair"],
 "time_range":[0,5],"with_frames":true}
\end{verbatim}

\section{Training and Evaluation Details}
\label{app:training}
Training first learns the shared output protocol through supervised fine-tuning and then optimizes action decisions using cumulative continuation returns. Both stages use the same vision--language model and structured outputs. This section specifies optimization, causal rollout construction, and answer scoring.

\subsection{Shared Configuration and Supervised Fine-Tuning}
We use Qwen3.5-4B and Qwen3.5-9B~\citep{qwen2026qwen35} with LoRA~\citep{hu2021lora} applied to the query and value projections, using rank eight, scaling factor sixteen, and zero dropout. Base-model parameters, including the visual encoder, are frozen. We use our dataset together with VSI-590K~\citep{yang2025cambrians} for SFT. Both stages use AdamW~\citep{loshchilov2017adamw} with a learning rate of $10^{-5}$ and gradient clipping at $1.0$. Training uses eight GPUs, one training item per GPU per microbatch, and eight gradient-accumulation steps, giving 64 items per update. SFT items are serialized streams; action-optimization items are decision histories whose candidate rollouts are evaluated internally. Training data follow the stream protocol of Section~\ref{sec:method_loop}: each serialized stream covers at most 180~s at 1~FPS with $640\times480$ RGB frames and starts from an empty context and memory. Within a stream, each supervised output is trained on its own causal prefix after applying the inference-time context-management rule ($L=16{,}384$ tokens over the complete active context, eviction of the visual tokens of the oldest eight frames, text retained within the stream), so training and inference use the same context-management rule.

SFT sequences preserve the order of observations, memory triggers, questions, and tool results. Only assistant outputs contribute to the token loss: memory records, retrieval requests, camera commands, and answers. Prompt and tool-result tokens provide context. The loss is averaged over supervised assistant tokens within each sequence and then over the effective batch. Linked questions retain their original referents and causal context.

\subsection{Bandit-Based Counterfactual Optimization}
\label{app:action_optimization}
Action optimization evaluates camera and stopping decisions using evidence increments and terminal answer quality. Construction thresholds select examples; the following scores define training rewards and common evidence metrics.

\paragraph{View-level evidence.}
For the one or two required objects $\mathcal O_q$, use the visible-pixel fraction $p_o(x)$ and framing completeness $f_o(x)$ defined in Appendix~\ref{app:evidence_measurements}. Define
\begin{align}
v_o(x)&=\min\{p_o(x)/0.10,1\},& s_o(x)&=v_o(x)f_o(x),\\
E_{\mathrm{pixel}}(x,q)&=\frac{1}{|\mathcal O_q|}\sum_{o\in\mathcal O_q}v_o(x),\\
E_{\mathrm{frame}}(x,q)&=\frac{1}{|\mathcal O_q|}\sum_{o\in\mathcal O_q}s_o(x),\\
E_{\mathrm{joint}}(x,q)&=\min_{o\in\mathcal O_q}s_o(x).
\end{align}
Normalization saturates at ten percent image occupancy. Framing discounts cropped targets, and joint evidence is limited by the less-supported object. For single-object questions, framing and joint scores coincide. These reward inputs use mask-derived pixel fractions, not projected-box occupancy.

\paragraph{Historical gain and cumulative returns.}
Let $j$ index camera attempts from zero, as in Section~\ref{sec:training_strategy} (interaction rounds, which also include retrievals, are indexed by $k$), and let $T$ denote the total number of camera attempts before answering; write $[z]_+=\max(z,0)$. The full configuration maintains $\Phi(h,q)=\max_{x\in\mathcal O(h)}E_{\mathrm{joint}}(x,q)$ over pre-question observations and preceding action results, with increment $d_j=\Phi_{j+1}-\Phi_j$, where $\Phi_j$ is $\Phi$ of the history before attempt $j$. The first three ablation configurations instead use $d_j=[E(x_{j+1},q)-E(x_j,q)]_+$ with their respective view score. Local increments can reward recovery of previously acquired evidence; historical gain removes this repeated reward.
\begin{equation}
r_j=0.10d_j-0.03-0.20\mathbf1[\mathrm{invalid}(a_j)],\qquad
G_j=\sum_{\ell=j}^{T-1}r_\ell+R_{\mathrm{QA}}.
\end{equation}
We use a constant cost of $0.03$ per camera attempt, discount $\gamma=1$, and at most three camera-action attempts. Invalid actions preserve the observation, receive no evidence gain, and consume an attempt. Answer-now terminates without camera cost; budget exhaustion requests a final answer. $R_{\mathrm{QA}}$ scores the model's generated answer by exact option correctness or numerical MRA, as defined in Appendix~\ref{app:evaluation}. In the full configuration, evidence increments telescope to $0.10(\Phi_T-\Phi_0)$, while action costs favor shorter paths with equal final evidence and answer quality.

\paragraph{Counterfactual candidate values.}
On rendered branch states, candidates are the five fixed-magnitude commands and \texttt{answer\_now}. All candidates start from an independent copy of the same history and remaining total budget. For each camera command, we average $B=4$ continuations:
\begin{equation}
\widehat Q^\mu(h,a)=\frac{1}{4}\sum_{b=1}^{4}G^{(b)}(h,a).
\end{equation}
Because terminal answers are decoded greedily, \texttt{answer\_now} has a deterministic return $R_{\mathrm{QA}}$, which one direct answer gives; each decision state thus requires 20 continuations and one direct answer.
A frozen snapshot of the current policy is refreshed before each update. Subsequent camera decisions are sampled from its categorical policy at temperature one; terminal answers use greedy decoding. Training includes initial histories and visited nonterminal histories. No new memory retrieval is introduced inside these rollouts.

Candidate scores sum conditional log probabilities of decision-value tokens and are normalized across candidates without length normalization. The action loss is
\begin{equation}
\mathcal L_{\mathrm{action}}=\mathbb E_h\!\left[
-\sum_a\pi_\theta(a\mid h)\widehat Q^\mu(h,a)
+0.05D_{\mathrm{KL}}(\pi_\theta(\cdot\mid h)\Vert\pi_{\mathrm{SFT}}(\cdot\mid h))\right].
\end{equation}
Values are detached during updates and retain their original scale. The loss sums over all candidate actions using the four-rollout value estimates. The SFT reference remains frozen. Private scores and alternative outcomes never enter the decision prefix.

\paragraph{Supervised replay.}
Training items are sampled with probabilities $\rho_{\mathrm{action}}=0.50$, $\rho_{\mathrm{QA}}=0.25$, and $\rho_{\mathrm{mem}}=0.25$. An action item contributes the counterfactual loss for a decision history; QA and memory items contribute assistant-token-averaged SFT losses for answers/tool calls and incremental writes, respectively. Averaging item losses gives the mixture objective
\begin{equation}
\mathcal L_{\mathrm{mix}}=
\mathbb E_{u\sim\rho}\mathbb E_{z\sim\mathcal D_u}[\ell_u(z)].
\end{equation}
The sampling probabilities set item frequencies, and sampled item losses are averaged with equal weights. Rollout branches are evaluated within each action item.

\paragraph{GRPO baseline.}
SFT + GRPO starts from the same SFT checkpoint as SFT + Bandit and uses the same reward, the same replay mixture, and the same update schedule. For each training question, it samples a group of twenty trajectories from the current policy and assigns every decision token of a trajectory the same advantage: the trajectory's return normalized by the mean and standard deviation of the returns in its group.

\paragraph{Ablations and common metrics.}
Reward variants progressively add framing, joint aggregation, and historical gain to pixel visibility, retaining the same cumulative-return definition. Memory-component, record-retention, write-interval, and reward variants are retrained under their respective configurations. GT memory replaces model-written records with the annotation-derived memory targets described in Appendix~\ref{app:construction}. Action-budget comparisons evaluate the same full policy with limits of one, two, three, five, or eight attempts; the policy is trained with at most three attempts and is not retrained for the other limits. The best view within three steps (Table~\ref{tab:action-direct}) is, for each question, the view with the highest $E(x,q)$ among those reachable by at most three commands. All reward variants are evaluated with the same full-history joint score. The independent full-visibility rate (Section~\ref{sec:experimental_setup}) counts an Action question as fully visible when every target is found in the final view by GroundingDINO~\citep{liu2024groundingdino} with SAM2~\citep{ravi2025sam2}, using box and text thresholds of 0.25 and 0.20, with mask IoU $\geq0.50$ against the ground truth and a detected box clear of the image border. Because $\Phi$ is also the training reward, the following metric is only a \emph{training-aligned} diagnostic, reported as $\Phi$-gain in Table~\ref{tab:action_full}:
\begin{align}
\Delta_i&=\Phi_{T_i}-\Phi_0\quad\text{for question }q_i,\\
\Phi\text{-gain}&=\frac{100}{N}\sum_i\Delta_i.
\end{align}
The initial history includes all observations available before acting. It includes all questions, including zero-action answers. Attempts counts the mean number of camera attempts per question, including invalid ones.

\subsection{Streaming Inputs and Main QA}
\label{app:evaluation}
All streaming models process observations chronologically at 1 FPS. Offline models receive up to 32 uniformly sampled frames from the available prefix, using all available frames when fewer are present. Future frames are withheld. Comparison models can also issue camera commands under the same interface and attempt budget, receiving the corresponding observations before answering; on real streams, a camera command returns the original frame. Multiple-choice answers are scored by exact option agreement. For nonzero references $y\neq0$, numeric answers use the mean relative accuracy of VSI-Bench~\citep{yang2024thinking}, with confidence thresholds $\mathcal C=\{0.50,0.55,\ldots,0.95\}$:
\begin{equation}
 \mathrm{MRA}(\hat y,y)=\frac{1}{|\mathcal C|}\sum_{\theta\in\mathcal C}
 \mathbf 1\!\left[\frac{|\hat y-y|}{|y|}<1-\theta\right].
\end{equation}
For zero-valued references, we use exact zero agreement. We report results by QA family, and separate one-, two-, three-, five-, and eight-attempt budgets when evaluating actions (Table~\ref{tab:action_full}). All compared models share the same question times, referents, and available observation prefixes, represented according to their streaming or offline input protocols.

\paragraph{Inference loop.}
Algorithm~\ref{alg:inference} gives the complete inference loop of Section~\ref{sec:method_loop} for one independent stream. A single context-management operation, \textsc{Fit}, is applied whenever content enters the model input: stream frames, memory writes, the question, and each request and result within a question. After the answer, the question, tool-interaction text, and answer stay in $C$, while retrieved frames and acquired views are removed; acquired views are never written to the frame store or the records of the Memory Store. The stream then resumes along its recorded trajectory from the buffered frames. We use \emph{closed-loop} only in this narrow sense: within one question, each retrieved record, frame, or new view informs the next decision, whereas the underlying stream remains the recorded one. Camera actions therefore do not alter the subsequent recorded observation trajectory, which keeps evaluation controlled; identical subsequent observations do not imply an identical subsequent model state, because text retained from earlier questions can influence later outputs. Writing acquired views into memory is left to future work (Section~\ref{sec:limitations}). In training and evaluation, the Camera Executor renders the resulting view in simulation. On a physical platform, the five commands would correspond to camera motions of the same magnitudes, with the camera replacing the renderer; we do not evaluate this setting. The collision checks that mark a command invalid use scene geometry, which a physical platform would obtain from depth or obstacle sensing; the model itself never receives depth.

The camera setting of Section~\ref{sec:exp_action} is a separate switch. With the camera disabled, camera commands are unavailable from the first round, so the exhausted-attempt rule never applies and retrieval remains available within its own budget; with the camera enabled, the model may still answer without any camera command. Exhausting the retrieval calls or, with the camera enabled, the camera attempts triggers a notice and a request for the final answer; once the image budget is exhausted, further retrieval returns frame references as text; invalid commands consume an attempt.
\begin{algorithm}[t]
\caption{MemAct streaming inference for one independent stream. The Memory Store (records $M$ and frame store $F$) and the Camera Executor form the tool runtime and have no learned parameters. $\oplus$ appends content; $\textsc{Fit}(X)$ applies the context-management rule: while $X$ exceeds $L$ tokens, the visual tokens of the oldest eight frames in $X$ are evicted, and text is kept.}
\label{alg:inference}
\small
\begin{algorithmic}[1]
\Require stream $o_1,\dots,o_n$ at 1 FPS ($n\leq180$); policy $\pi_\theta$; context limit $L$; camera setting $\kappa\in\{\text{enabled},\text{disabled}\}$
\State $C\gets$ system prompt;\; $M\gets\emptyset$;\; $F\gets\emptyset$ \Comment{every stream starts afresh}
\For{$t=1,\dots,n$}
  \State $C\gets\textsc{Fit}(C\oplus o_t)$;\; add $o_t$ to $F$
  \If{$t \bmod 5=0$}
    \State $\Delta M\gets\pi_\theta(C,\text{write prompt})$ \Comment{one record per frame; at most two retries if rejected}
    \State $M\gets\operatorname{Update}(M,\Delta M)$;\; $C\gets\textsc{Fit}(C\oplus\Delta M)$
  \EndIf
  \If{a question $q$ arrives at $t$} \Comment{pause; buffer new frames}
    \State $b_0\gets$ 4 calls, 4 images, and, if $\kappa=$ enabled, 3 attempts
    \State $h_0\gets\textsc{Fit}(C\oplus q)$ \Comment{camera commands offered only if $\kappa=$ enabled}
    \For{$k=0,1,\dots$}
      \State $u_k\gets\pi_\theta(h_k,b_k)$
      \If{$u_k$ is an answer}
        \State output $u_k$; \textbf{break}
      \ElsIf{$u_k$ is a \texttt{use\_memory} request}
        \State $z_k\gets\textsc{Query}(M,F,u_k,b_k)$ \Comment{frame references as text once images run out}
      \Else \Comment{camera command}
        \State $z_k\gets\textsc{CameraExecutor}(u_k)$ \Comment{invalid commands consume an attempt; never added to $M,F$}
      \EndIf
      \State $h_{k+1}\gets\textsc{Fit}(h_k\oplus u_k\oplus z_k)$; update $b_{k+1}$
      \If{no retrieval calls remain, or $\kappa=$ enabled and no camera attempts remain}
        \State notify and request a final answer
      \EndIf
    \EndFor
    \State $C\gets h_k\oplus u_k$ with retrieved frames and acquired views removed \Comment{question, tool text, and answer kept}
    \State resume the recorded stream from the buffered frames
  \EndIf
\EndFor
\end{algorithmic}
\end{algorithm}

\paragraph{Streaming context.}
\label{app:streaming_context}
Training and evaluation use independent streams of at most 180~s, sampled at 1~FPS with $640\times480$ RGB frames; a new stream reinitializes the context, the records, and the frame store. The budget $L=16{,}384$ tokens applies to the complete active context: the system prompt and other retained text, visual tokens, and question-time tool interactions. When the context would exceed $L$, the visual tokens of the oldest eight frames are evicted and the retained context is re-prefilled; the system prompt, questions and answers, memory records, and tool-interaction text are retained within the stream. These values are our implementation; the idea of removing old visual tokens to bound the KV cache follows StreamingVLM~\citep{xu2025streamingvlm}, and the idea of reducing only visual tokens while carrying text forward in the KV cache follows Stream3D-VLM~\citep{yu2026stream3d}. The budget bounds the active context and its KV cache, not the external frame store and Memory Store, which keep every frame and record of the stream, so evicted frames remain retrievable. Training applies the same rule to the causal prefix of every supervised output.

Under the reported stream lengths and input configuration, the most crowded processed contexts still retained 22 visual frames after eviction, and the retained text remained compatible with the total context budget in both training and inference. This count is an observation in the reported configuration rather than a fixed hyperparameter, and early in a stream fewer frames have arrived. It characterizes the finite-stream setting evaluated in this work; retaining all textual history over arbitrarily long streams would require an additional text-memory management policy.

\paragraph{Offline QA prompts.}
Offline QA prompts present chronologically ordered frames with timestamps, followed by the question and its labeled options or numerical unit. When submitting a final answer, models return exactly one JSON object, \texttt{\{"type":"answer","answer":value\}}. For multiple-choice questions, \texttt{value} must be a single uppercase option letter rather than the option text. For numerical questions, it must be a JSON number in the unit specified by the question; object counts must be integers. Linked questions include preceding questions and the model's own predictions as context.

\paragraph{Main QA aggregation and latency.}
The main evaluation contains 9,094 simulated and 9,068 real QA turns. Overall ACC pools the multiple-choice turns and Overall MRA the numerical turns, weighting each task and domain by its applicable question count. Action contributes simulated questions only; no zero score is imputed for the real domain. Linked questions are scored per answer turn. Answers are parsed as in VSI-Bench. Latency is measured with three warm-up calls and pooled over all main QA turns; generation latency excludes memory operations and action execution, and runtime latency includes them; neither includes rendering. The per-write latency is the corresponding mean over memory writes. These are means and do not bound the latency of individual operations.

\subsection{Timing Evaluation}
\label{app:timing}
Timing is a supplementary evaluation, separate from the main QA of MemAct-Bench. It contains 2,000 questions from five templates (1,787 multiple-choice and 213 numerical; 1,000 simulated and 1,000 real), presented before their evidence appears. It evaluates when to answer; memory and camera tools are disabled so that the timing decision is isolated from retrieval and acting. Frames arrive chronologically at 1 FPS, retaining observed history and withholding future frames. At each checkpoint the model outputs \texttt{wait} or \texttt{answer}; the first answer is final.

The first answerable time $t_i^*$ is derived from visibility and evidence references: the unique earliest option-category appearance, the queried category's first appearance, the earliest prefix leaving one consistent appearance-order option, or the arrival of the referenced historical frame for coappearance and direction questions. Tied first appearances are excluded, and derived answers must match ground truth. The model does not receive $t_i^*$; no example is answerable at the initial frame.

For first-answer time $t_i^a$, let $z_i=\mathbf1[t_i^*\leq t_i^a\leq t_i^*+5\,\mathrm{s}]$, or zero if unanswered. The inclusive window uses video time, not inference latency. An answer at the deadline is allowed; waiting at that checkpoint is unanswered. We report
\begin{align}
T&=\frac{100}{|\mathcal Q|}\sum_{i\in\mathcal Q}z_i,\\
JA&=\frac{100}{|\mathcal Q_c|}\sum_{i\in\mathcal Q_c}z_i\mathbf1[\hat y_i=y_i],\\
JM&=\frac{100}{|\mathcal Q_n|}\sum_{i\in\mathcal Q_n}z_i m_i.
\end{align}
Here $m_i\in[0,1]$ is numerical MRA; $|\mathcal Q|=2{,}000$, $|\mathcal Q_c|=1{,}787$, and $|\mathcal Q_n|=213$. Denominators include all applicable questions, with early, late, and unanswered cases contributing zero. Timing is separate from Overall QA.

\paragraph{Results.}
Table~\ref{tab:timing} reports the three metrics. MemAct-9B reaches 85.45 T, 56.86 JA, and 51.92 JM. Fine-tuning Qwen3.5-9B on our QA data without memory or action supervision raises T from 42.00 to 63.60 (same-backbone control; the retrieval baseline reaches 63.50), and the complete MemAct training scheme raises T/JA/JM from 63.60/38.39/33.05 to 85.45/56.86/51.92. No variant isolates memory supervision from action supervision on Timing, so we do not attribute this gain to either component.
\begin{table}[h]
\centering\small
\caption{Timing results. T: on-time rate; JA: joint timing-and-answer accuracy; JM: timing-gated MRA.}
\label{tab:timing}
\begin{tabular}{lccc}
\toprule
Model & T & JA & JM\\
\midrule
GPT-6 Astra & 81.05 & 54.62 & 50.42 \\
Gemini-3.8-Flash & 67.60 & 39.90 & 36.06 \\
\midrule
Qwen3.5-4B & 31.35 & 14.72 & 11.22 \\
Qwen3.5-9B & 42.00 & 20.59 & 16.10 \\
Qwen3.8-27B & 47.55 & 23.84 & 19.77 \\
InternVL3.5-8B & 33.95 & 14.27 & 12.02 \\
InternVL3.5-38B & 38.70 & 17.35 & 15.26 \\
\midrule
VLM-3R & 44.10 & 21.82 & 18.45 \\
Cambrian-S & 43.35 & 21.15 & 17.51 \\
Cambrian-S-FT & 74.30 & 48.29 & 39.15 \\
\midrule
StreamingVLM & 62.85 & 38.56 & 31.74 \\
Stream3D-VLM & 61.60 & 36.65 & 30.23 \\
\midrule
Same-backbone control & 63.60 & 38.39 & 33.05 \\
Retrieval baseline & 63.50 & 39.00 & 32.63 \\
\midrule
MemAct-4B & 85.35 & 53.16 & 44.93 \\
MemAct-9B & 85.45 & 56.86 & 51.92 \\
MemAct (Qwen2.5-VL-7B) & 85.40 & 54.67 & 46.81 \\
\bottomrule
\end{tabular}
\end{table}

\section{Test-Set Construction and Composition}
\label{app:test_composition}
\paragraph{Split and answer balancing.}
Splits group recordings by physical space: the ScanNet space (the \texttt{sceneXXXX} prefix shared by repeated scans), the ScanNet++ scene, the ARKitScenes visit, and the 3D-FRONT layout; samples whose physical space cannot be determined are dropped. Within each template, test answers are balanced in a fixed hash order that does not depend on any model output: the most frequent answer content may exceed chance by at most 5 points, binary templates are balanced within 2 points, and numerical templates are rebalanced over per-template quantile bins until the training-median constant scores at most 20 MRA. No test space appears in VSI-590K or in the external benchmarks of Section~\ref{sec:exp_external}, and the Action test set uses 255 3D-FRONT layouts that appear in no training set (Appendix~\ref{app:distribution}).

\paragraph{Evidence-location annotations.}
For a Memory question, the evidence frames are those that determine the answer under the template's rule (\eg, the last frame showing the category for last-seen questions). Retained/evicted labels are computed once under a fixed reference protocol, the eviction rule of Section~\ref{sec:method_loop} replayed for the question, and are shared by all models: the evidence counts as retained (in context) only if every required evidence frame is still in $C_{t_q}$, and as evicted if at least one has left it. Question times are sampled so that about 56\% of record-answerable and 70\% of frame-required questions are labeled evicted (62.0\% overall). An evicted label does not mean that all relevant information is gone, since records and other text may remain in context. The R/F type is likewise a reference classification by question template and record format and does not guarantee that a model's own records are complete or accurate. The evidence age is $t_q$ minus the time of the earliest required evidence frame; ages are binned at 10, 30, 60, and 120~s. For an Action question, let $x_0$ be the initial view and $\Phi_{\mathrm{pre}}(q)=\max_{x\in O_{\mathrm{pre}}}E(x,q)$ the best view quality over the entire pre-question history (Appendix~\ref{app:construction}). The question \emph{needs no move} if $E(x_0,q)\geq0.2$; it \emph{needs completion} if no pre-question view meets the prescribed quality criterion, $\Phi_{\mathrm{pre}}(q)\leq0.05$, and some view reachable by a command sequence of at most three steps has $E(x,q)\geq0.2$; it is excluded otherwise. The test strata use only these quality thresholds, not the additional endpoint-gain and occlusion-support conditions of training demonstrations. The Action test set contains 1,051 questions: 282 need no move and 769 (73.2\%; at least 70\% by construction) need completion, of which 470 are one-step and 299 multi-step. All evidence-location annotations and strata were fixed before any model was evaluated.

\paragraph{Failure attribution.}
\label{app:attribution}
Every evaluation run logs each memory write and whether it was accepted, each retrieval request and its returned records and frames, each camera command and resulting view, and the final answer. Because writes and retrievals are explicit structured outputs, fixed rules assign each error to a stage. Each wrong Memory answer receives the first label whose condition holds: (M4) \emph{in-context error}, when the required evidence frames are still in context and the model issues no retrieval; (M3) \emph{wrong although retrieved}, when a retrieval for the question returned the required evidence (records suffice for record-answerable templates, whereas frame-required templates need the evidence frames); (M1) \emph{write error}, when the write covering the evidence frames is missing or wrong; or (M2) \emph{not retrieved}, when correct records exist but the required evidence was not returned (since records also remain in the text context, M2 does not mean that the information was inaccessible to the model). An Action error is classified as (A1) \emph{premature stop}, when the model answers with $\Phi<0.2$ before exhausting its attempts; (A2) \emph{ineffective exploration}, when it uses all attempts without reaching $\Phi\geq0.2$; or (A3) \emph{wrong despite meeting the predefined view-quality threshold}, when $\Phi\geq0.2$ at answering time; meeting this threshold does not imply that all evidence needed for the answer was obtained. Invalid commands and unnecessary actions on no-move-needed questions are flagged separately. Attribution is automatic; a manual check of 150 attributed errors agrees with the automatic labels in 148 cases (Section~\ref{sec:exp_failure}), which shows that the rules are applied consistently but does not establish a unique cause for each error.

\paragraph{Composition.}
Table~\ref{tab:test_composition} lists the test questions, episodes, and scenes or layouts per template, the share of Memory questions with evicted evidence, the share of Action questions that need completion, and whether each Memory template is record-answerable (R) or frame-required (F). Frame-required templates need instance identity across frames (distinct counts) or 3D geometry at a past time (nearest neighbors), which the records do not contain. MemAct-Bench has 11,335 multiple-choice and 6,827 numerical QA turns.
\begin{longtable}{p{0.40\textwidth}rrrrr}
\caption{MemAct-Bench test composition per template. Evicted / compl.: share (\%) of Memory questions with evicted evidence, or of Action questions that need completion. For Action templates, Episodes counts question start states, one per question.}\\
\label{tab:test_composition}\\
\toprule Template & Test QA & Episodes & Scenes & Evicted / compl. & Type\\\midrule\endfirsthead
\toprule Template & Test QA & Episodes & Scenes & Evicted / compl. & Type\\\midrule\endhead
\bottomrule\endfoot
\multicolumn{6}{l}{\textbf{Current QA}}\\
approached object category & 452 & 197 & 196 & -- & -- \\
became visible category & 599 & 212 & 211 & -- & -- \\
camera motion bucket & 724 & 236 & 235 & -- & -- \\
camera path length m & 712 & 238 & 237 & -- & -- \\
camera rotation deg & 726 & 238 & 237 & -- & -- \\
camera to object distance bucket & 261 & 162 & 161 & -- & -- \\
camera to object distance m & 673 & 224 & 223 & -- & -- \\
closer than before & 414 & 179 & 178 & -- & -- \\
closest category then distance & 534 & 173 & 172 & -- & -- \\
closest category then relation & 1,224 & 219 & 218 & -- & -- \\
closest neighbor category & 194 & 99 & 98 & -- & -- \\
closest object category & 674 & 227 & 226 & -- & -- \\
disappeared category & 592 & 213 & 212 & -- & -- \\
fixed reference relation & 137 & 83 & 81 & -- & -- \\
multi candidate height & 119 & 78 & 78 & -- & -- \\
object category then later visibility & 636 & 166 & 165 & -- & -- \\
object relative to camera & 376 & 193 & 192 & -- & -- \\
room floor area & 100 & 100 & 96 & -- & -- \\
\multicolumn{6}{l}{\textbf{Memory QA}}\\
first appearance order & 166 & 163 & 157 & 56.0 & R \\
first seen category then timestamp & 974 & 209 & 208 & 56.1 & R \\
first seen object category & 445 & 209 & 209 & 56.0 & R \\
first seen timestamp s & 525 & 216 & 215 & 56.0 & R \\
historical coappearance category & 1,058 & 326 & 298 & 56.0 & R \\
historical relative direction & 287 & 135 & 129 & 55.7 & R \\
last seen timestamp & 1,084 & 339 & 309 & 55.9 & R \\
seen count by category & 683 & 184 & 183 & 70.0 & F \\
single seen object then closest category & 1,814 & 203 & 202 & 70.0 & F \\
single seen object then closest distance & 928 & 203 & 202 & 70.0 & F \\
\multicolumn{6}{l}{\textbf{Action QA}}\\
object size m & 687 & 687 & 255 & 74.8 & -- \\
object to object distance bucket & 142 & 142 & 69 & 84.5 & -- \\
object to object distance m & 222 & 222 & 128 & 60.8 & -- \\
\end{longtable}

\section{Baselines and Prompts}
\label{app:baselines}
\paragraph{Baseline settings.}
Table~\ref{tab:main_full} gives each model's backbone and whether it was trained on our data, always together with VSI-590K; none of the compared models other than MemAct was trained with memory or action supervision, since training a backbone with that supervision and the tool protocol yields MemAct on that backbone (Section~\ref{sec:experimental_setup}). Streaming models run at 1~FPS with $L=16{,}384$; offline models receive 32 uniformly sampled frames; closed-source models are queried through their APIs with temperature zero, the same frames, and the same JSON answer format. For backbone-bound methods, we align the frame rate, the context budget, the training QA data, and the tool protocol. Two settings cannot be aligned: Cambrian-S, Cambrian-S-FT, and VLM-3R are evaluated with 32 uniformly sampled frames, because their released inference takes offline inputs, and Stream3D-VLM keeps its StreamVGGT geometry branch (518-pixel crops), so it receives geometry features that the other models do not.

\paragraph{Tool-protocol prompt.}
Zero-shot models receive the same memory and camera tools through a text protocol, so that the comparison does not depend on fine-tuning for tool use. The prompt asks for exactly one JSON object per turn, lists the five camera commands with their names and magnitudes and the \texttt{use\_memory} request format, and states the per-question budgets.
Table~\ref{tab:tool_use} reports how often each model's outputs violate the protocol and how often it acts on Action questions. Zero-shot models follow the protocol in at least 93.8\% of cases but act on only 9.9--15.4\% of Action questions, so their small gains from the action interface reflect rarely acting rather than failing the protocol.
\begin{table}[h]
\centering\small
\caption{Protocol compliance and action use on Action questions. Format failures: outputs that do not follow the tool protocol; Acts: share of questions on which the model issues at least one camera command, over all Action questions and over those that need completion. Rows with three seeds report the mean.}
\label{tab:tool_use}
\begin{tabular}{lccc}
\toprule
Model & Format failures (\%) & Acts, all (\%) & Acts, needs compl. (\%)\\
\midrule
GPT-6 Astra & 0.95 & 15.4 & 15.9 \\
Gemini-3.8-Flash & 1.05 & 11.2 & 10.5 \\
Qwen3.5-4B & 6.18 & 11.4 & 11.6 \\
Qwen3.5-9B & 4.09 & 9.9 & 9.8 \\
Qwen3.8-27B & 2.19 & 10.0 & 9.8 \\
InternVL3.5-8B & 5.33 & 11.2 & 11.2 \\
InternVL3.5-38B & 2.95 & 11.4 & 11.2 \\
VLM-3R & 3.24 & 10.3 & 10.5 \\
Cambrian-S & 2.85 & 12.9 & 11.3 \\
\midrule
Cambrian-S-FT & 0.95 & 11.5 & 11.7 \\
StreamingVLM & 0.19 & 12.2 & 12.5 \\
Stream3D-VLM & 0.48 & 11.8 & 11.7 \\
Same-backbone control & 0.67 & 13.7 & 13.5 \\
Retrieval baseline & 0.19 & 11.6 & 12.2 \\
\midrule
MemAct-9B & 0.44 & 79.8 & 99.6 \\
\bottomrule
\end{tabular}
\end{table}

\section{External Benchmarks}
\label{app:external}
All external benchmarks follow their official protocols, including inputs, action formats, and scoring. On AVS-HM3D we use the official VG-AVS evaluation code with Habitat-Sim~\citep{savva2019habitat} on HM3D~\citep{ramakrishnan2021hm3d}: the model outputs one action in the benchmark's format (heading rotation, forward distance, and final rotation), and the official verifier answers from the resulting view, scored by LLM-Match; MemAct produces this format through the official prompt without training on it. On ESI-Bench we evaluate all 3,081 tasks and follow the official evaluation standard, including its action steps and metric. For the streaming evaluation of VSI-Bench and ReVSI, videos are played at 1~FPS with $L=16{,}384$; with memory disabled, the same model runs without memory writes or retrieval, so only the frames retained in its context are available. This switch differs from the question-time retrieval ablation of Section~\ref{sec:exp_memory}, which keeps memory writing and disables only new retrieval for the current question.

\begin{table}[h]
\centering
\caption{\small Transfer results. AVS-HM3D (LLM-Match) and ESI-Bench under their official evaluation standards; on AVS-HM3D, MemAct-9B selects the view and the official verifier answers. On VSI-Bench and ReVSI, MemAct-9B streams at 1~FPS with memory disabled (same model, no memory writes or retrieval) or enabled, while published models use offline inputs and are listed for reference. --: not reported.}
\label{tab:transfer}
\begingroup
\small
\setlength{\tabcolsep}{2pt}
\begin{tabular}{lcccc}
\toprule
Model & AVS-HM3D & ESI-Bench & VSI-Bench & ReVSI \\
\midrule
Query view, no action~\citep{koo2025vgavs} & 58.65 & -- & -- & -- \\
GPT-5~\citep{koo2025vgavs} & 64.91 & -- & -- & -- \\
VG-AVS (SFT+RL)~\citep{koo2025vgavs} & 70.70 & -- & -- & -- \\
Target view, upper bound~\citep{koo2025vgavs} & 78.09 & -- & -- & -- \\
Gemini-3.1-Pro, active~\citep{hong2026esi} & -- & 56.90 & -- & -- \\
GPT-5.2 & -- & -- & 54.49 & 50.90 \\
Gemini-3-Pro & -- & -- & 55.10 & 60.90 \\
Qwen3.5-9B & -- & -- & 57.90 & 51.60 \\
VLM-3R-7B & -- & -- & 60.90 & 50.10 \\
Cambrian-S-7B & -- & -- & 67.50 & 49.10 \\
\midrule
MemAct-9B, memory disabled & -- & -- & 60.03 & 48.31 \\
MemAct-9B & 68.39 & 45.65 & 63.44 & 52.37 \\
\bottomrule
\end{tabular}
\endgroup
\end{table}

\paragraph{Active perception results.}
AVS-HM3D~\citep{koo2025vgavs} poses 208 open-ended EXPRESS-Bench~\citep{Jiang2025iccv} questions from views of real HM3D scans in which the target is partly occluded or poorly framed, and ESI-Bench~\citep{hong2026esi} evaluates the perception--action loop on 3,081 tasks in simulated homes; we evaluate on the full set of each, follow their official evaluation standards, and do not train on either. On AVS-HM3D, where MemAct-9B controls the view and the official verifier answers, MemAct-9B reaches 68.39, 9.74 points above answering from the query view (58.65) and 3.48 above GPT-5 (Table~\ref{tab:transfer}), so its view selection transfers to real scans and a continuous action format; this setting does not test the full memory and QA system or deployment on a physical robot. On ESI-Bench, whose tasks also require locomotion and manipulation beyond the local view completion that MemAct is trained for, it reaches 45.65, below Gemini-3.1-Pro with active perception (56.90), which marks a capability boundary between this broader active embodiment and local view completion.

\paragraph{Video spatial benchmark results.}
On VSI-Bench~\citep{yang2024thinking} and ReVSI~\citep{zhang2026revsi}, published results~\citep{li2026orarl,zhang2026revsi,dai2026sagent,wang2026guide,taichu2026results,spatialaxiom2026results} use offline inputs, so we list them for reference rather than as a same-protocol comparison. Streaming with memory, MemAct-9B scores 63.44 and 52.37, below Cambrian-S-7B on VSI-Bench, whose training set VSI-590K is also part of ours, and below Gemini-3-Pro on ReVSI. For the same model, enabling memory adds 3.41 and 4.06 points; this switch turns memory writing and retrieval on or off together, unlike the question-time ablation of Table~\ref{tab:retrieval-ablation} (image benchmarks: Table~\ref{tab:image_spatial}).

\section{Additional Results}
\label{app:full_results}
\paragraph{Backbones, latency, and reference baselines.}
Table~\ref{tab:main_full} repeats the results of Table~\ref{tab:main} with each model's backbone and latency and adds the reference baselines. The modal-content baseline reaches 29.58 Action ACC, close to the 25.00 of uniform choice; the training-median constant stays below 15 MRA in every family; and the blind baseline exceeds uniform choice by at most 2.99 ACC points. No zero-shot model falls below the reference baselines, although on numerical Action questions Qwen3.5-4B (15.09) and InternVL3.5-8B (15.48) stay within 2.2 MRA points of the constant. For MemAct-9B, the mean runtime latency is 0.56~s per question, pooled over all main QA turns, and 0.28~s per memory write under the corresponding definition; neither includes rendering (Appendix~\ref{app:evaluation}).
\begin{table}[h]
\centering
\caption{\small MemAct-Bench results of Table~\ref{tab:main} with backbones, latency, and reference baselines. ACC and MRA for multiple-choice and numerical questions; latency in seconds per question (generation / runtime, pooled means excluding rendering; Appendix~\ref{app:evaluation}). \emph{Data}: fine-tuning on our data together with VSI-590K, where QA = QA turns only and Full = QA, memory, and action supervision. Rows with three seeds report the mean.}
\label{tab:main_full}
\vspace{-3mm}
\begingroup
\fontsize{8.5}{9}\selectfont
\renewcommand{\arraystretch}{0.92}
\setlength{\tabcolsep}{3pt}
\setlength{\aboverulesep}{1pt}
\setlength{\belowrulesep}{1pt}
\resizebox{\textwidth}{!}{%
\begin{tabular}{@{}l@{\hspace{6pt}}l@{\hspace{6pt}}c@{\hspace{9pt}}rr@{\hspace{10pt}}rr@{\hspace{10pt}}rr@{\hspace{10pt}}rr@{\hspace{9pt}}c@{}}
\toprule
Model / baseline & Backbone & Data & \multicolumn{2}{c}{Current} & \multicolumn{2}{c}{Memory} & \multicolumn{2}{c}{Action} & \multicolumn{2}{c}{Overall QA} & Latency (s) \\
\cmidrule(lr){4-5}\cmidrule(lr){6-7}\cmidrule(lr){8-9}\cmidrule(lr){10-11}\cmidrule(l){12-12}
 & & & ACC & MRA & ACC & MRA & ACC & MRA & ACC & MRA & gen. / runtime \\
\midrule
\rowcolor{MergedBand}\multicolumn{12}{l}{\strut\color{MergedInk}\itshape Reference baselines} \\
Uniform options & -- & -- & 22.90 & -- & 21.77 & -- & 25.00 & -- & 22.50 & -- & -- \\
Modal answer content & -- & -- & 19.67 & -- & 21.33 & -- & 29.58 & -- & 20.41 & -- & -- \\
Modal option letter & -- & -- & 24.86 & -- & 23.68 & -- & 28.17 & -- & 24.46 & -- & -- \\
Numerical constant & -- & -- & -- & 13.46 & -- & 14.40 & -- & 13.33 & -- & 13.95 & -- \\
Blind (text only) & Qwen3.5-9B & QA & 25.89 & 14.06 & 22.79 & 16.32 & 19.72 & 14.58 & 24.65 & 15.35 & -- \\
\rowcolor{MergedBand}\multicolumn{12}{l}{\strut\color{MergedInk}\itshape Closed-source models} \\
GPT-6 Astra & closed & \xmark & 68.07 & 50.36 & 68.48 & 61.50 & 49.30 & 46.53 & 67.98 & 55.90 & -- \\
Gemini-3.8-Flash & closed & \xmark & 58.48 & 40.16 & 60.96 & 47.95 & 39.44 & 36.62 & 59.17 & 43.92 & -- \\
\rowcolor{MergedBand}\multicolumn{12}{l}{\strut\color{MergedInk}\itshape Open-source models} \\
Qwen3.5-4B & Qwen3.5-4B & \xmark & 40.61 & 17.65 & 54.66 & 34.11 & 29.58 & 15.09 & 45.75 & 26.25 & 1.56 / 1.71 \\
Qwen3.5-9B & Qwen3.5-9B & \xmark & 44.45 & 21.83 & 60.49 & 41.41 & 34.51 & 18.21 & 50.35 & 31.98 & 2.48 / 2.68 \\
Qwen3.8-27B & Qwen3.8-27B & \xmark & 46.32 & 23.29 & 62.74 & 44.55 & 38.03 & 21.07 & 52.39 & 34.54 & 5.42 / 5.81 \\
InternVL3.5-8B & InternVL3.5-8B & \xmark & 39.65 & 18.11 & 56.05 & 35.69 & 31.69 & 15.48 & 45.71 & 27.30 & 2.50 / 2.71 \\
InternVL3.5-38B & InternVL3.5-38B & \xmark & 42.24 & 20.31 & 57.58 & 39.06 & 35.92 & 17.35 & 47.92 & 30.10 & 5.45 / 5.83 \\
\rowcolor{MergedBand}\multicolumn{12}{l}{\strut\color{MergedInk}\itshape Spatial models} \\
VLM-3R & LLaVA-Video-7B-Qwen2 & \xmark & 44.65 & 23.61 & 63.40 & 44.33 & 35.21 & 19.78 & 51.57 & 34.35 & 2.62 / 2.84 \\
Cambrian-S & Qwen2.5-7B & \xmark & 44.05 & 22.57 & 61.22 & 43.51 & 35.21 & 19.77 & 50.38 & 33.57 & 2.45 / 2.65 \\
Cambrian-S-FT & Qwen2.5-7B & QA & 66.36 & 37.31 & 67.35 & 52.08 & 46.48 & 35.67 & 66.48 & 45.11 & 2.52 / 2.73 \\
\rowcolor{MergedBand}\multicolumn{12}{l}{\strut\color{MergedInk}\itshape Streaming models} \\
StreamingVLM & Qwen2.5-VL-7B & QA & 62.27 & 35.79 & 61.99 & 50.50 & 30.99 & 30.23 & 61.77 & 43.04 & 0.38 / 0.46 \\
Stream3D-VLM & Qwen2.5-VL-7B + StreamVGGT-1B & QA & 59.76 & 32.46 & 61.29 & 46.73 & 27.46 & 27.35 & 59.93 & 39.53 & 0.41 / 0.49 \\
\rowcolor{MergedBand}\multicolumn{12}{l}{\strut\color{MergedInk}\itshape Controls (Qwen3.5-9B backbone)} \\
Same-backbone control & Qwen3.5-9B & QA & 66.68 & 49.61 & 71.04 & 59.94 & 42.49 & 45.13 & 68.01 & 54.63 & -- / -- \\
Retrieval baseline & Qwen3.5-9B & QA & 66.67 & 48.91 & 72.21 & 60.17 & 40.85 & 43.37 & 68.43 & 54.29 & -- / -- \\
\rowcolor{MergedBand}\multicolumn{12}{l}{\strut\color{MergedInk}\itshape Ours} \\
MemAct-4B & Qwen3.5-4B & Full & 61.06 & 36.06 & 65.19 & 55.98 & 50.70 & 49.08 & 62.48 & 48.61 & 0.38 / 0.52 \\
MemAct-9B & Qwen3.5-9B & Full & 68.03 & 52.33 & 74.02 & 64.02 & 52.35 & 52.63 & 70.08 & 58.72 & 0.42 / 0.56 \\
MemAct (Qwen2.5-VL-7B) & Qwen2.5-VL-7B & Full & 65.50 & 40.45 & 67.35 & 56.88 & 51.41 & 50.13 & 66.02 & 50.66 & 0.41 / 0.54 \\
\bottomrule
\end{tabular}%
}
\endgroup
\vspace{-3mm}
\end{table}

\paragraph{Question-time retrieval ablation.}
\label{app:retrieval_ablation}
Table~\ref{tab:retrieval-ablation} follows this protocol. (1)~Each of the three MemAct-9B checkpoints is evaluated with retrieval enabled and disabled, and the two settings are paired per question. (2)~For every question, both branches start from the same pre-question state, with the same retained textual and visual history. (3)~The disabled branch only prevents the current question from issuing new retrieval requests; records, previously retrieved text, and earlier answers already in context remain. (4)~Memory writing and all other settings are unchanged, including the camera permission on Memory QA. (5)~The enabled branch continues the original stream, whereas the disabled branch serves only as a diagnostic for the current question, and its output is not written back to the subsequent stream. (6)~Both branches use the same context budget; we do not claim that their subsequent token counts, tool calls, or computational cost are equal. Question groups follow the reference annotations of Appendix~\ref{app:test_composition}: R/F is the benchmark classification by question template and record format, and retained/evicted labels come from the fixed reference protocol, where evicted means that at least one required visual frame has been evicted, not that all relevant information is gone. The evicted groups contain 4,938 of the 7,964 questions (62.0\%), the share given by the reference labels. $N$ counts unique questions rather than evaluations summed over checkpoints; scores are means over the three checkpoints, and differences come from the paired comparison. The 95\% confidence intervals use a paired cluster bootstrap with task scenes as clusters, which keeps each enabled/disabled pair together. The gains on record-answerable evicted (+3.59) and frame-required retained questions (+1.36) may reflect more targeted presentation of records or historical images, and a frame that is still in context is not necessarily located and used; the experiment does not separate these mechanisms. It also does not separate retrieved text from retrieved images or measure the full contribution of textual memory, and, because the disabled branch is a diagnostic for the current question rather than a system that never retrieves from the start of the stream, it is not a decomposition of MemAct's gain over the same-backbone control.

\paragraph{Memory ablations.}
\emph{GT memory} replaces model-written records with the annotation-derived targets. Each \emph{w/o} variant removes one field from the records and is retrained. \emph{Record retention} deletes records and frames older than the retention window from the Memory Store while the active-context rule stays unchanged, and is retrained with the same retention; unlike the question-time retrieval ablation, it removes stored evidence. \emph{Write-interval} variants are retrained with the corresponding interval. These are retrained component variants: the full model is the mean over three seeds, and each variant uses one seed. Table~\ref{tab:memory-ablation} reports the results. Counts matter mainly for numerical questions and relations mainly for multiple-choice ones: removing counts lowers MRA by 1.19 while leaving ACC within 0.30, and removing relations lowers ACC by 1.29 while leaving MRA within 0.38. Among the tested intervals, five seconds performs on par with shorter intervals while writing less often.
\begin{table}[h]
\centering
\caption{\small MemAct-9B Memory QA ablations: (A) memory content, (B) retention of memory records, (C) write interval. ACC / MRA on the Memory test set. The full model (Full memory, Full history, 5 s) is the mean over three seeds; variants use one seed.}
\label{tab:memory-ablation}
\vspace{-3mm}
\begingroup
\fontsize{8}{9}\selectfont
\renewcommand{\arraystretch}{0.90}
\setlength{\tabcolsep}{3pt}
\setlength{\aboverulesep}{1pt}
\setlength{\belowrulesep}{1pt}
\begin{minipage}[t]{0.32\textwidth}
\vspace{0pt}
\begin{tabular*}{\linewidth}{@{\extracolsep{\fill}}lrr@{}}
\toprule
\rowcolor{MemoryBand}\multicolumn{3}{l}{\strut\color{MemoryInk}\itshape A. Memory components} \\
Setting & ACC & MRA \\
\midrule
Full memory & 74.02 & 64.02 \\
GT memory & 77.31 & 67.64 \\
w/o Frames & 69.77 & 60.82 \\
w/o Count & 74.32 & 62.83 \\
w/o Relations & 72.73 & 64.40 \\
\multicolumn{3}{c}{} \\
\bottomrule
\end{tabular*}
\end{minipage}%
\hfill%
\begin{minipage}[t]{0.32\textwidth}
\vspace{0pt}
\begin{tabular*}{\linewidth}{@{\extracolsep{\fill}}lrr@{}}
\toprule
\rowcolor{MemoryBand}\multicolumn{3}{l}{\strut\color{MemoryInk}\itshape B. Record retention} \\
Retention & ACC & MRA \\
\midrule
10 s & 67.96 & 58.71 \\
20 s & 69.60 & 60.10 \\
30 s & 70.78 & 61.10 \\
40 s & 71.81 & 62.21 \\
50 s & 73.31 & 63.07 \\
Full history & 74.02 & 64.02 \\
\bottomrule
\end{tabular*}
\end{minipage}%
\hfill%
\begin{minipage}[t]{0.32\textwidth}
\vspace{0pt}
\begin{tabular*}{\linewidth}{@{\extracolsep{\fill}}lrr@{}}
\toprule
\rowcolor{MemoryBand}\multicolumn{3}{l}{\strut\color{MemoryInk}\itshape C. Write interval} \\
Interval & ACC & MRA \\
\midrule
1 s & 73.22 & 63.25 \\
2 s & 74.11 & 63.60 \\
3 s & 73.53 & 63.46 \\
5 s & 74.02 & 64.02 \\
10 s & 72.52 & 61.87 \\
\multicolumn{3}{c}{} \\
\bottomrule
\end{tabular*}
\end{minipage}%
\endgroup
\vspace{-3mm}
\end{table}

\paragraph{Action ablations.}
Table~\ref{tab:action_full} reports all Action metrics for the Stage 2 objectives (A), the reward composition (B), and the action budget (C). Reward ablations share the terminal answer reward, action costs, and cumulative returns. Adding framing completeness improves Score by 1.95 points ([1.47, 2.44]) over pixel visibility alone, requiring joint support for both targets adds 1.30 ([0.76, 1.84]), and measuring gains against the best historical view rather than the current one adds 2.83 ([2.31, 3.36]); the full-visibility rate rises from 63.50\% to 68.00\%, 71.20\%, and 77.45\% along the way. Each increment belongs to its step in this cumulative order and is not an order-independent component effect. In part C, the same policy, trained with at most three attempts, is evaluated under different limits: raising the limit from one to three attempts raises Score from 49.12 to 52.59 (+3.47, [2.96, 3.98]) and multi-step Score from 29.96 to 40.19 (+10.23), whereas limits of five and eight add only 0.21 ([0.03, 0.40]) and 0.23 ([0.06, 0.41]); because the test set contains only questions whose adequate view is reachable within three steps, this limited benefit should not be extrapolated to local active-view tasks in general. Under the three-attempt limit, 25.89\% of no-move-needed questions still receive actions.
\begin{table}[h]
\centering
\caption{\small MemAct-9B Action QA ablations: (A) training strategy, (B) reward composition, (C) action budget (training used at most three). ACC, MRA, and Score on the full Action set; No move / 1-step / Multi: Score per stratum (Section~\ref{sec:test_construction}); Full vis.: full-visibility rate (independent); $\Phi$-gain: training-aligned evidence gain; Attempts: mean camera attempts per question, including invalid ones. Every row starts from MemAct-9B's pre-question states and is answered by the final MemAct-9B, so rows differ only in the views their policies acquire; Table~\ref{tab:answerer} repeats the comparisons with a frozen SFT answerer. Part A extends Table~\ref{tab:action-ablation}. All rows report the mean over three seeds.}
\label{tab:action_full}
\vspace{-2mm}
\begingroup
\fontsize{9}{9.5}\selectfont
\setlength{\tabcolsep}{5pt}
\renewcommand{\arraystretch}{0.90}
\setlength{\aboverulesep}{1pt}
\setlength{\belowrulesep}{1pt}
\begin{tabular*}{\textwidth}{@{\extracolsep{\fill}}lccccccccc@{}}
\toprule
Setting & ACC & MRA & Score & No move & 1-step & Multi & Full vis. & $\Phi$-gain & Attempts \\
\midrule
\rowcolor{ActionBand}\multicolumn{10}{l}{\strut\color{ActionInk}A. Training strategy} \\
Camera disabled & 38.97 & 46.05 & 45.10 & 57.60 & 47.96 & 28.81 & 32.64 & 0.00 & 0.00 \\
SFT only & 44.60 & 48.61 & 48.07 & 57.13 & 52.25 & 32.94 & 66.03 & 25.58 & 2.05 \\
SFT + GRPO & 46.24 & 49.32 & 48.91 & 56.90 & 53.35 & 34.37 & 66.41 & 26.76 & 1.96 \\
SFT + Bandit (full) & 52.35 & 52.63 & 52.59 & 57.39 & 57.60 & 40.19 & 77.45 & 33.05 & 1.89 \\
\addlinespace[1pt]
\rowcolor{ActionBand}\multicolumn{10}{l}{\strut\color{ActionInk}B. Reward composition (cumulative)} \\
Pixel visibility & 41.31 & 47.33 & 46.51 & 56.78 & 50.70 & 30.25 & 63.50 & 24.05 & 2.12 \\
+ Full framing & 46.01 & 48.84 & 48.46 & 57.14 & 53.01 & 33.12 & 68.00 & 26.65 & 2.00 \\
+ Two-target evidence & 45.31 & 50.46 & 49.76 & 57.23 & 54.23 & 35.70 & 71.20 & 29.07 & 1.91 \\
+ Historical gain (full) & 52.35 & 52.63 & 52.59 & 57.39 & 57.60 & 40.19 & 77.45 & 33.05 & 1.89 \\
\addlinespace[1pt]
\rowcolor{ActionBand}\multicolumn{10}{l}{\strut\color{ActionInk}C. Action budget (same policy, different limits)} \\
At most 1 attempt & 47.65 & 49.35 & 49.12 & 57.48 & 56.30 & 29.96 & 64.10 & 21.16 & 0.80 \\
At most 2 attempts & 50.23 & 51.64 & 51.45 & 57.43 & 57.54 & 36.22 & 72.98 & 28.77 & 1.46 \\
At most 3 attempts (full) & 52.35 & 52.63 & 52.59 & 57.39 & 57.60 & 40.19 & 77.45 & 33.05 & 1.89 \\
At most 5 attempts & 53.29 & 52.72 & 52.80 & 57.25 & 57.54 & 41.15 & 77.55 & 33.17 & 2.22 \\
At most 8 attempts & 53.29 & 52.74 & 52.82 & 57.36 & 57.52 & 41.13 & 77.61 & 33.18 & 2.25 \\
\bottomrule
\end{tabular*}
\endgroup
\vspace{-4mm}
\end{table}

\paragraph{Seed variation and confidence intervals.}
Table~\ref{tab:seed_sd} reports the mean and standard deviation over three seeds for MemAct-9B and the same-backbone control. Table~\ref{tab:paired_ci} reports paired differences with 95\% confidence intervals from an episode-level cluster bootstrap with 10{,}000 resamples; Action intervals are clustered by layout (255 layouts; 69 for the multiple-choice template). Across the rows of Table~\ref{tab:action_full}, the seed standard deviation is at most 0.76 for Score, 3.33 for ACC (142 questions), and 1.43 for the full-visibility rate.
\begin{table}[h]
\centering\footnotesize
\setlength{\tabcolsep}{3pt}
\caption{Mean $\pm$ standard deviation over three seeds on MemAct-Bench.}
\label{tab:seed_sd}
\begin{tabular}{lcccccccc}
\toprule
 & \multicolumn{2}{c}{Current} & \multicolumn{2}{c}{Memory} & \multicolumn{2}{c}{Action} & \multicolumn{2}{c}{Overall}\\
Model & ACC & MRA & ACC & MRA & ACC & MRA & ACC & MRA\\
\midrule
MemAct-9B & 68.03 $\pm$ 0.23 & 52.33 $\pm$ 0.42 & 74.02 $\pm$ 0.25 & 64.02 $\pm$ 0.16 & 52.35 $\pm$ 2.47 & 52.63 $\pm$ 0.33 & 70.08 $\pm$ 0.19 & 58.72 $\pm$ 0.26 \\
Same-backbone control & 66.68 $\pm$ 0.22 & 49.61 $\pm$ 0.39 & 71.04 $\pm$ 0.19 & 59.94 $\pm$ 0.33 & 42.49 $\pm$ 0.41 & 45.13 $\pm$ 0.13 & 68.01 $\pm$ 0.20 & 54.63 $\pm$ 0.31 \\
\bottomrule
\end{tabular}
\end{table}
\begin{table}[h]
\centering\small
\caption{Paired differences (points) with 95\% cluster-bootstrap confidence intervals. Top: MemAct-Bench comparisons (Tables~\ref{tab:main} and~\ref{tab:memory-grouped}). Bottom: Action comparisons of Table~\ref{tab:action_full}, all answered by the final MemAct-9B, as Score on all Action questions and on those that need completion. Differences use unrounded means and may differ by 0.01 from differences of the rounded table entries.}
\label{tab:paired_ci}
\begin{tabular}{llcc}
\toprule
Comparison & Group & ACC & MRA\\
\midrule
MemAct-9B $-$ same-backbone control & Current & 1.35 [0.64, 2.05] & 2.72 [2.27, 3.18] \\
MemAct-9B $-$ same-backbone control & Memory & 2.98 [2.07, 3.89] & 4.08 [3.75, 4.43] \\
MemAct-9B $-$ same-backbone control & Action & 9.86 [2.55, 17.14] & 7.49 [5.24, 9.72] \\
MemAct-9B $-$ same-backbone control & Overall & 2.07 [1.54, 2.61] & 4.09 [3.71, 4.49] \\
MemAct-9B $-$ GPT-6 Astra & Overall & 2.10 [1.26, 2.92] & 2.82 [2.29, 3.35] \\
MemAct-9B $-$ retrieval baseline & Memory & 1.81 [0.99, 2.64] & 3.85 [3.55, 4.15] \\
MemAct-9B $-$ retrieval baseline & Overall & 1.66 [1.12, 2.18] & 4.43 [4.01, 4.85] \\
\midrule
Comparison & & All Action & Needs completion\\
\midrule
Bandit $-$ camera disabled & & 7.49 [6.78, 8.23] & 10.32 [9.38, 11.27] \\
Bandit $-$ SFT only & & 4.52 [3.93, 5.13] & 6.09 [5.34, 6.86] \\
Bandit $-$ GRPO & & 3.68 [3.12, 4.25] & 4.85 [4.13, 5.59] \\
Bandit $-$ scripted scan & & 7.63 [6.90, 8.36] & 9.19 [8.25, 10.11] \\
Bandit $-$ random commands & & 8.75 [8.10, 9.43] & 10.20 [9.33, 11.08] \\
+ Full framing $-$ pixel visibility & & 1.95 [1.47, 2.44] & 2.53 [1.91, 3.17] \\
+ Two-target $-$ + full framing & & 1.30 [0.76, 1.84] & 1.74 [1.03, 2.46] \\
+ Historical gain $-$ + two-target & & 2.83 [2.31, 3.36] & 3.81 [3.12, 4.51] \\
3 attempts $-$ 1 attempt & & 3.47 [2.96, 3.98] & 4.77 [4.11, 5.44] \\
3 attempts $-$ 2 attempts & & 1.14 [0.79, 1.50] & 1.58 [1.12, 2.04] \\
5 attempts $-$ 3 attempts & & 0.21 [0.03, 0.40] & 0.34 [0.11, 0.58] \\
8 attempts $-$ 3 attempts & & 0.23 [0.06, 0.41] & 0.32 [0.09, 0.56] \\
\bottomrule
\end{tabular}
\end{table}

\paragraph{Answerer robustness of the Action comparisons.}
All rows of Table~\ref{tab:action-ablation} (top) are answered by the final MemAct-9B, which is itself the Bandit-trained model. To check that the comparisons do not favor the policy whose model answers, we let a frozen SFT model, the Stage 1 checkpoint from which every Stage 2 variant starts, answer from the views each policy acquires (Table~\ref{tab:answerer}; the first two rows also appear in Table~\ref{tab:action-ablation}, bottom). Every comparison keeps its sign and its confidence interval excludes zero, so the policy gain remains positive with a shared frozen answerer; the gaps are smaller, so the magnitude of the gain depends on the answerer, and the final-answerer gaps are not purely answerer-independent policy effects.
\begin{table}[h]
\centering\small
\caption{Action comparisons (Score, points, 95\% CI) under two answerers: the final MemAct-9B (as in Table~\ref{tab:action-ablation}) and a frozen SFT model.}
\label{tab:answerer}
\begin{tabular}{lcccc}
\toprule
 & \multicolumn{2}{c}{MemAct-9B answerer} & \multicolumn{2}{c}{Frozen SFT answerer}\\
\cmidrule(lr){2-3}\cmidrule(l){4-5}
Comparison & All & Needs compl. & All & Needs compl.\\
\midrule
Bandit $-$ SFT only & 4.52 [3.93, 5.13] & 6.09 [5.34, 6.86] & 2.07 [1.78, 2.37] & 2.20 [1.83, 2.59] \\
Bandit $-$ GRPO & 3.68 [3.12, 4.25] & 4.85 [4.13, 5.59] & 1.75 [1.43, 2.08] & 1.90 [1.52, 2.30] \\
Bandit $-$ scripted scan & 7.63 [6.90, 8.36] & 9.19 [8.25, 10.11] & 2.85 [2.49, 3.21] & 3.22 [2.74, 3.69] \\
Bandit $-$ random commands & 8.75 [8.10, 9.43] & 10.20 [9.33, 11.08] & 3.27 [2.95, 3.60] & 3.50 [3.10, 3.92] \\
\bottomrule
\end{tabular}
\end{table}

\paragraph{Image spatial benchmarks.}
We also evaluate MemAct-9B on the image spatial benchmarks MMSI-Bench~\citep{yang2025mmsi} and ViewSpatial~\citep{li2025viewspatial}, following each benchmark's protocol, and compare with published results\citep{li2026orarl,zhang2026revsi,dai2026sagent,wang2026guide,taichu2026results,spatialaxiom2026results} in Table~\ref{tab:image_spatial}. MemAct-9B reaches 37.80 on MMSI-Bench and 52.21 on ViewSpatial. It obtains the highest ViewSpatial score among the listed models and exceeds both spatial models, VLM-3R-7B and Cambrian-S-7B, on MMSI-Bench, so its spatial training also transfers to image-based spatial reasoning.
\begin{table}[h]
\centering\small
\caption{Image spatial benchmarks. Published results are taken from the cited papers.}
\label{tab:image_spatial}
\begin{tabular}{lcc}
\toprule
Model & MMSI-Bench & ViewSpatial\\
\midrule
GPT-5.2 & 41.30 & 47.30\\
Gemini-3-Pro & 45.20 & 50.40\\
Qwen3.5-9B & 36.50 & 48.20\\
VLM-3R-7B & 27.90 & 40.50\\
Cambrian-S-7B & 27.10 & 41.30\\
\midrule
MemAct-9B & 37.80 & 52.21\\
\bottomrule
\end{tabular}
\end{table}

\paragraph{Memory accuracy by evidence age.}
Table~\ref{tab:memory_age} groups Memory QA by the time between the earliest required evidence frame and the question. All models degrade as the evidence ages, and the lead of MemAct-9B over the same-backbone control grows from 0.81 points at most 10~s after the evidence to 5.21 points beyond 120~s.
\begin{table}[h]
\centering\small
\caption{Memory QA Score by evidence age at question time. Oracle retrieval is a diagnostic that gives MemAct-9B the annotated evidence frames.}
\label{tab:memory_age}
\begin{tabular}{lccccc}
\toprule
Model & $\leq$10 s & 10--30 s & 30--60 s & 60--120 s & $>$120 s\\
\midrule
Share of questions & 8.9\% & 18.7\% & 24.1\% & 26.2\% & 22.1\%\\
StreamingVLM & 81.70 & 71.41 & 58.95 & 48.17 & 41.58\\
Same-backbone control & 85.74 & 77.62 & 67.65 & 59.45 & 53.61\\
Retrieval baseline & 85.18 & 77.44 & 68.37 & 60.59 & 55.16\\
Oracle retrieval (diagnostic) & 89.16 & 83.33 & 76.88 & 70.97 & 66.92\\
MemAct-9B & 86.55 & 79.31 & 71.07 & 63.75 & 58.82\\
\bottomrule
\end{tabular}
\end{table}

\paragraph{Per-template results.}
Table~\ref{tab:per_template} reports ACC for multiple-choice templates and MRA for numerical templates.
\begin{longtable}{p{0.34\textwidth}rrrrrr}
\caption{Per-template results on MemAct-Bench (ACC for multiple-choice templates, MRA for numerical ones). Same-bb.: same-backbone control; Astra: GPT-6 Astra; Modal: modal answer content (multiple-choice only); Blind: text-only baseline. The linked first-seen-category-then-timestamp template is reported per turn.}\\
\label{tab:per_template}\\
\toprule Template & MemAct-9B & Same-bb. & StreamingVLM & Astra & Modal & Blind\\\midrule\endfirsthead
\toprule Template & MemAct-9B & Same-bb. & StreamingVLM & Astra & Modal & Blind\\\midrule\endhead
\bottomrule\endfoot
\multicolumn{7}{l}{\textbf{Current QA}}\\
approached object category & 73.38 & 72.12 & 67.04 & 72.12 & 14.38 & 19.25 \\
became visible category & 68.06 & 65.66 & 61.27 & 66.94 & 11.02 & 20.70 \\
camera motion bucket & 76.47 & 75.41 & 52.62 & 75.69 & 29.97 & 25.69 \\
camera path length m & 47.73 & 45.81 & 27.49 & 47.32 & -- & 13.89 \\
camera rotation deg & 45.49 & 42.61 & 27.77 & 43.61 & -- & 14.56 \\
camera to object distance bucket & 57.85 & 53.90 & 53.26 & 59.77 & 29.89 & 30.27 \\
camera to object distance m & 66.95 & 63.53 & 53.36 & 62.67 & -- & 13.77 \\
closer than before & 53.95 & 52.01 & 39.13 & 56.28 & 29.95 & 29.23 \\
closest category then distance & 67.17 & 65.86 & 69.48 & 70.41 & 14.98 & 25.47 \\
closest category then relation & 70.92 & 71.21 & 69.28 & 68.46 & 16.75 & 24.10 \\
closest neighbor category & 68.21 & 66.49 & 65.98 & 68.04 & 26.29 & 22.16 \\
closest object category & 74.13 & 73.49 & 73.44 & 74.04 & 14.39 & 24.93 \\
disappeared category & 79.67 & 78.66 & 69.76 & 80.57 & 11.99 & 23.65 \\
fixed reference relation & 70.80 & 68.61 & 67.15 & 72.26 & 15.33 & 27.74 \\
multi candidate height & 64.71 & 61.62 & 61.34 & 65.55 & 25.21 & 32.77 \\
object category then later visibility & 63.89 & 61.16 & 62.74 & 64.31 & 25.94 & 38.99 \\
object relative to camera & 37.32 & 35.55 & 39.36 & 39.63 & 25.00 & 24.47 \\
room floor area & 36.33 & 33.83 & 34.90 & 38.20 & -- & 13.50 \\
\multicolumn{7}{l}{\textbf{Memory QA}}\\
first appearance order & 62.65 & 59.04 & 54.22 & 57.23 & 1.81 & 29.52 \\
first seen category then timestamp (ACC) & 85.63 & 82.55 & 74.33 & 82.96 & 23.00 & 22.38 \\
first seen category then timestamp (MRA) & 76.65 & 71.23 & 60.10 & 72.65 & -- & 13.55 \\
first seen object category & 78.20 & 75.06 & 69.66 & 75.51 & 20.22 & 23.15 \\
first seen timestamp s & 73.12 & 67.46 & 57.87 & 67.77 & -- & 11.89 \\
historical coappearance category & 66.95 & 63.52 & 55.10 & 61.91 & 19.38 & 27.13 \\
historical relative direction & 78.63 & 68.64 & 64.11 & 71.43 & 29.97 & 23.69 \\
last seen timestamp & 68.43 & 64.07 & 64.07 & 67.99 & -- & 15.12 \\
seen count by category & 53.60 & 49.87 & 40.83 & 51.71 & -- & 28.29 \\
single seen object then closest category & 74.31 & 72.84 & 61.19 & 67.25 & 22.71 & 19.51 \\
single seen object then closest distance & 54.78 & 52.35 & 32.55 & 51.71 & -- & 12.87 \\
\multicolumn{7}{l}{\textbf{Action QA}}\\
object size m & 49.71 & 41.92 & 29.90 & 42.15 & -- & 14.59 \\
object to object distance bucket & 52.35 & 42.49 & 30.99 & 49.30 & 29.58 & 19.72 \\
object to object distance m & 61.64 & 55.09 & 31.26 & 60.09 & -- & 14.55 \\
\end{longtable}

\section{Qualitative Examples}
\label{app:visualization}
\begin{figure}[h]
\centering
\includegraphics[width=\linewidth]{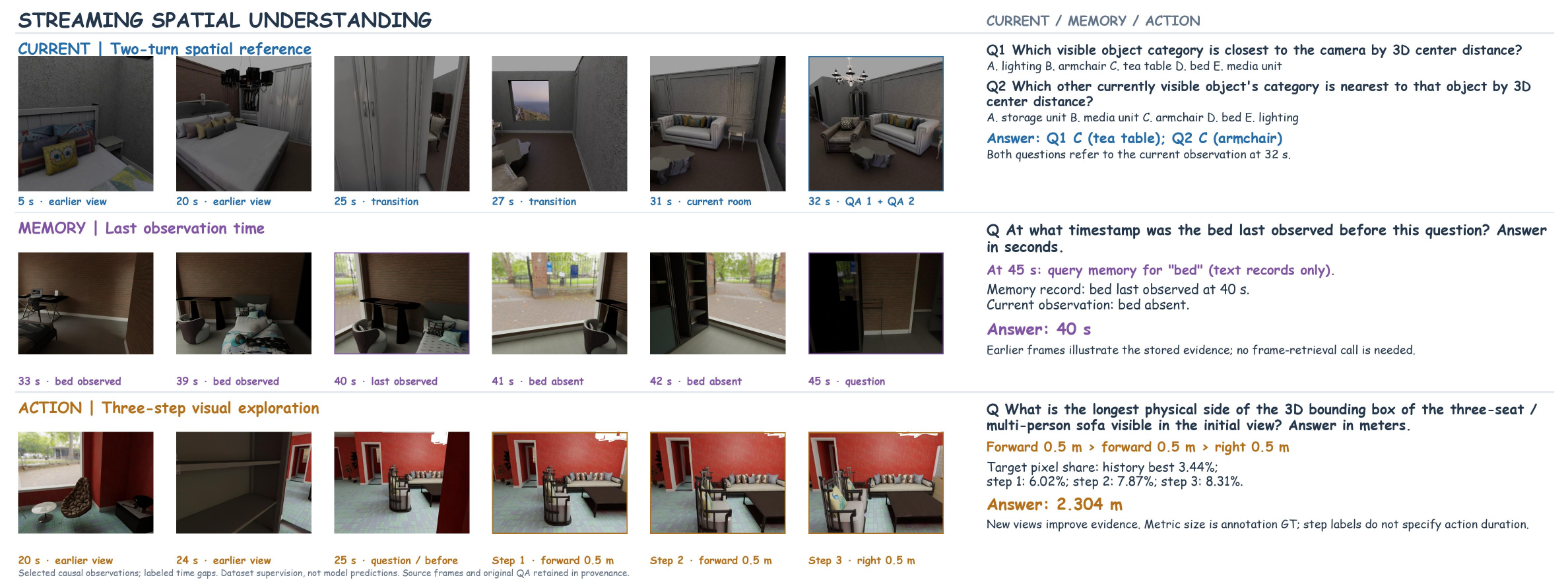}
\vspace{-3mm}
\caption{\small Examples of Current, Memory, and Action QA. Current QA uses the latest observation; the Memory example is record-answerable, recovering the last observation time from memory records; the Action example completes a partly visible target with three camera commands.}
\label{fig:trace_samples}
\vspace{-5mm}
\end{figure}

We provide visual examples of the evidence used by each QA family and of the incremental memory records that support historical retrieval. Figure~\ref{fig:trace_samples} gives one example per family.

\begin{figure}
    \centering
    \includegraphics[width=1.0\linewidth,height=0.85\textheight,keepaspectratio]{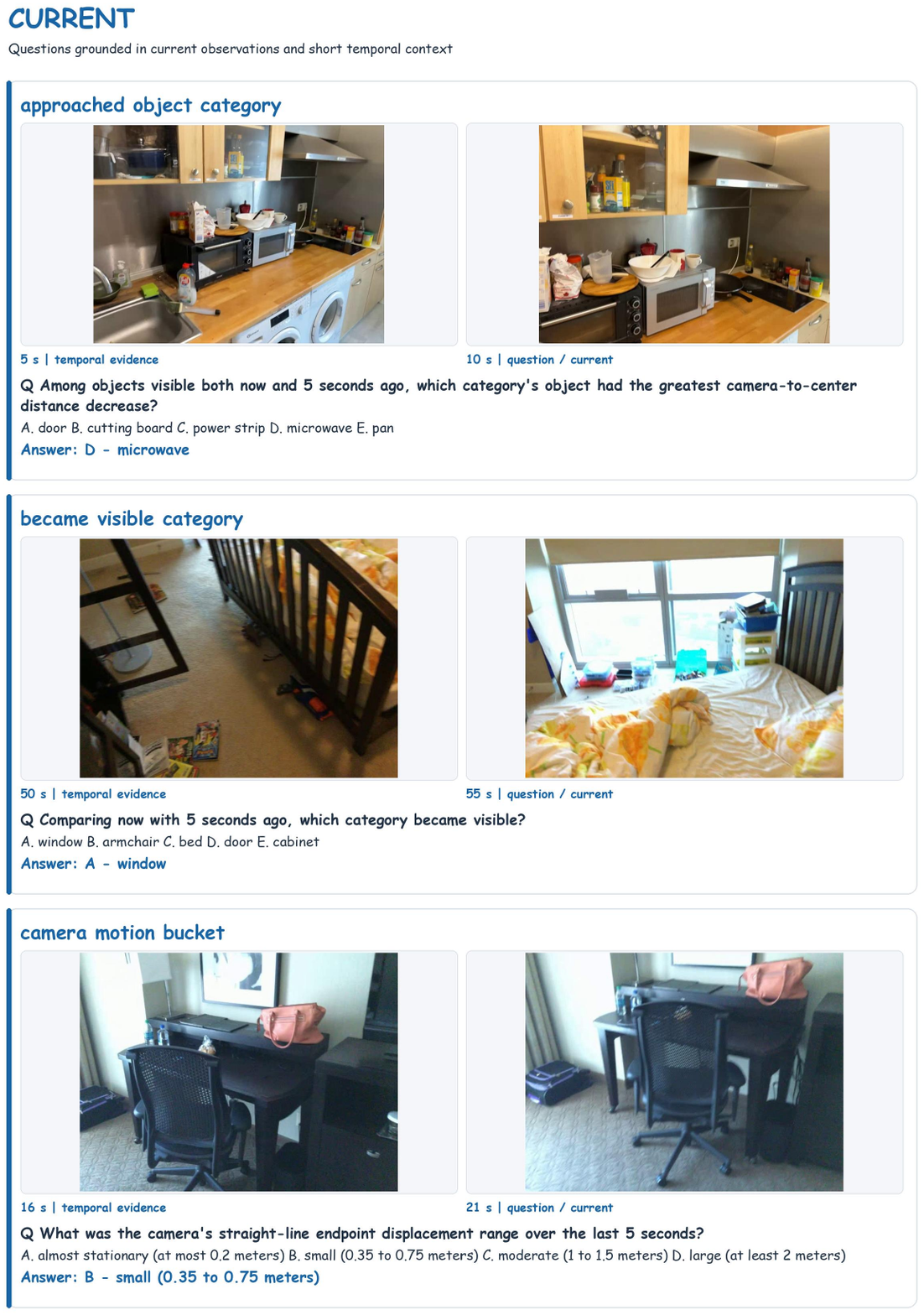}
    \caption{Current QA sample 1}
    \label{fig:current_sample_1}
\end{figure}

\begin{figure}
    \centering
    \includegraphics[width=1.0\linewidth,height=0.85\textheight,keepaspectratio]{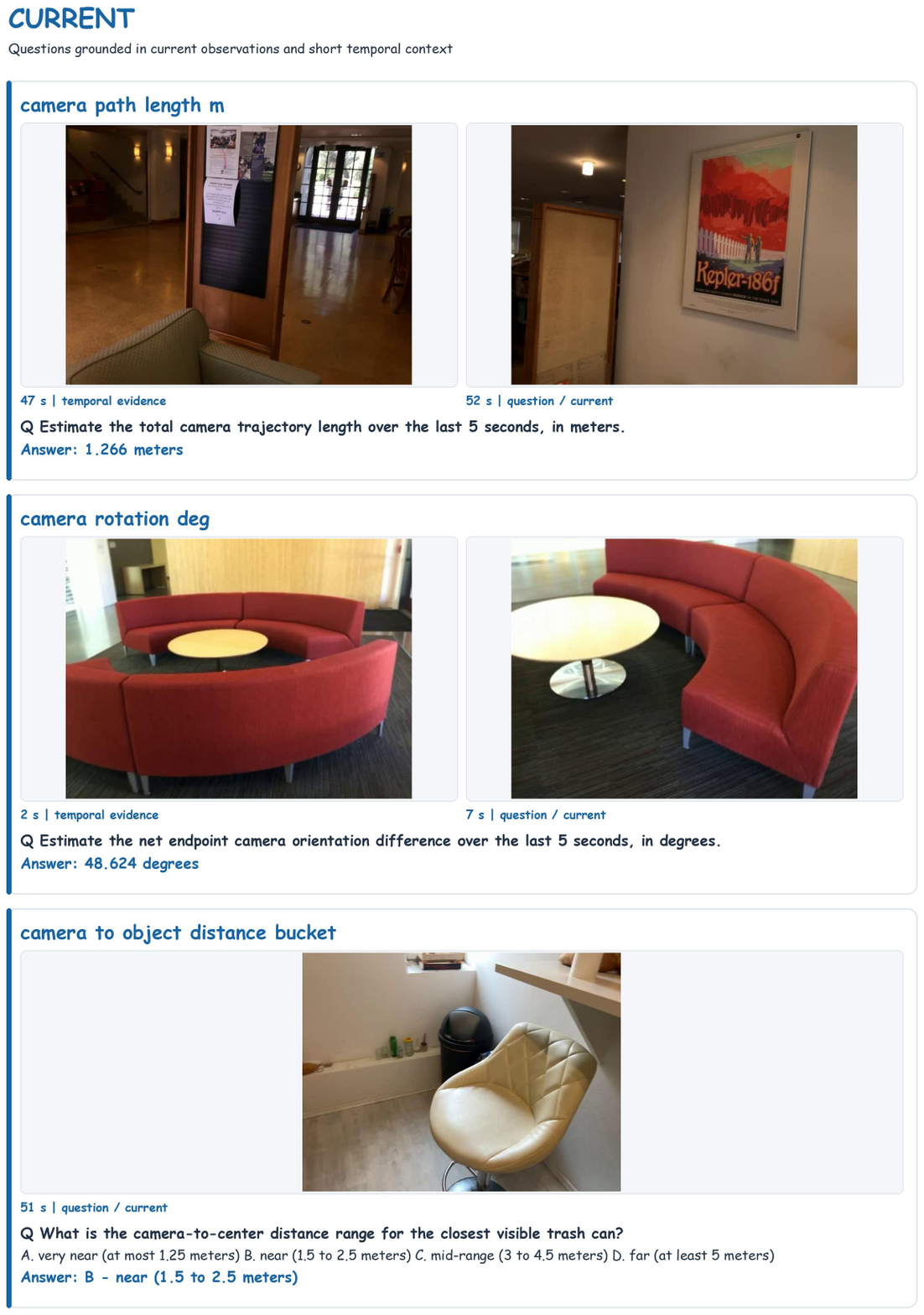}
    \caption{Current QA sample 2}
    \label{fig:current_sample_2}
\end{figure}

\begin{figure}
    \centering
    \includegraphics[width=1.0\linewidth,height=0.85\textheight,keepaspectratio]{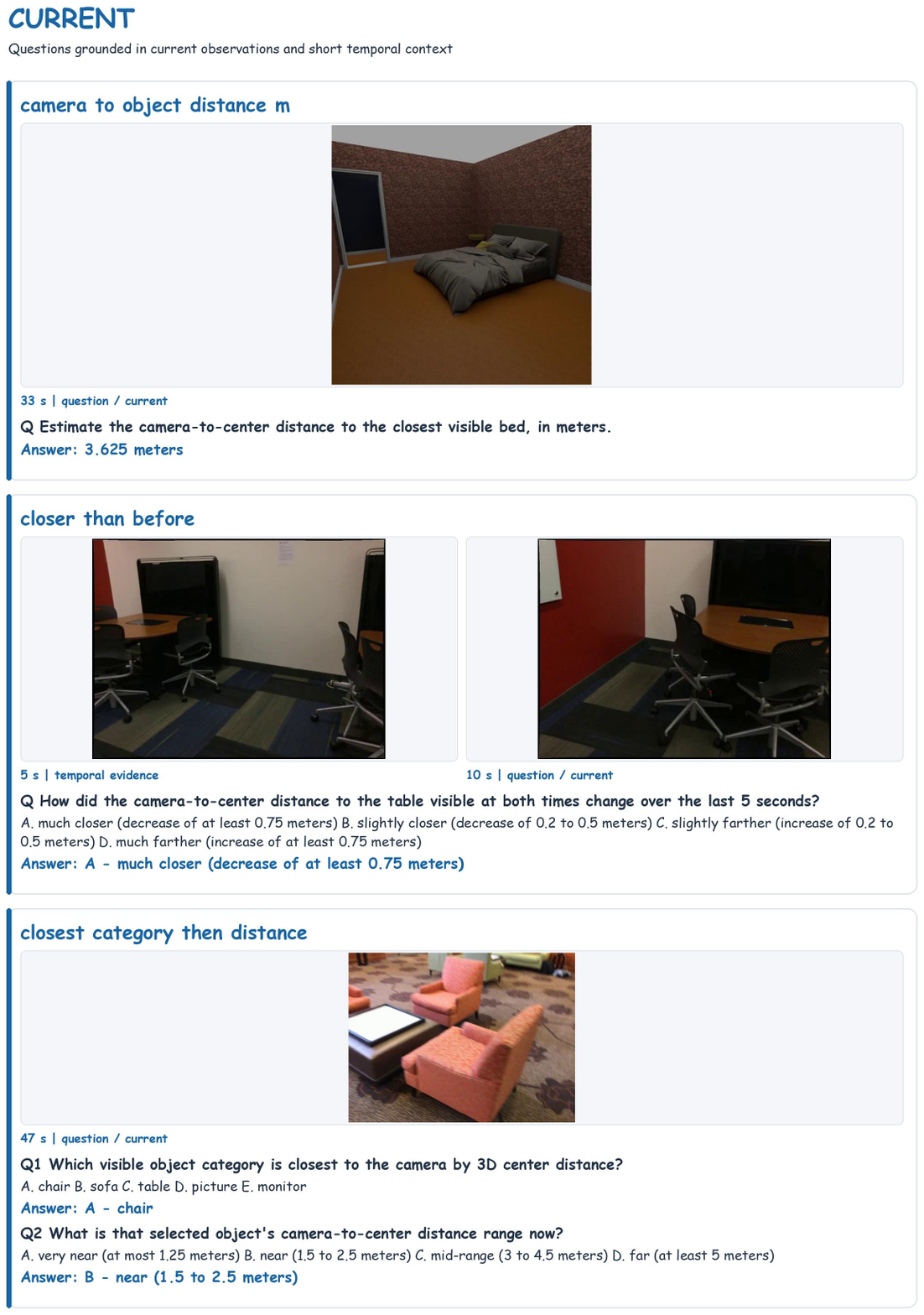}
    \caption{Current QA sample 3}
    \label{fig:current_sample_3}
\end{figure}

\begin{figure}
    \centering
    \includegraphics[width=1.0\linewidth,height=0.85\textheight,keepaspectratio]{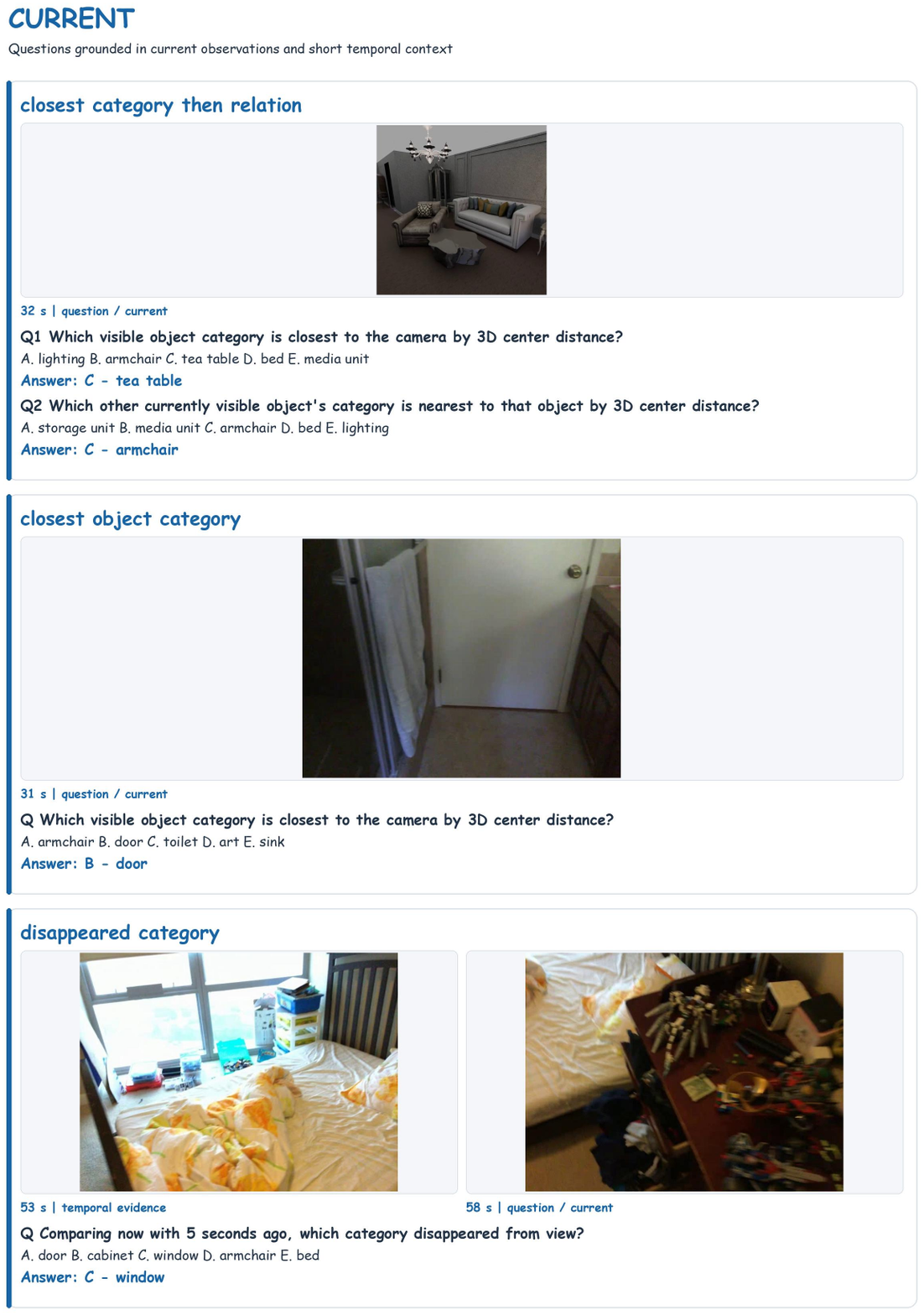}
    \caption{Current QA sample 4}
    \label{fig:current_sample_4}
\end{figure}

\begin{figure}
    \centering
    \includegraphics[width=1.0\linewidth,height=0.85\textheight,keepaspectratio]{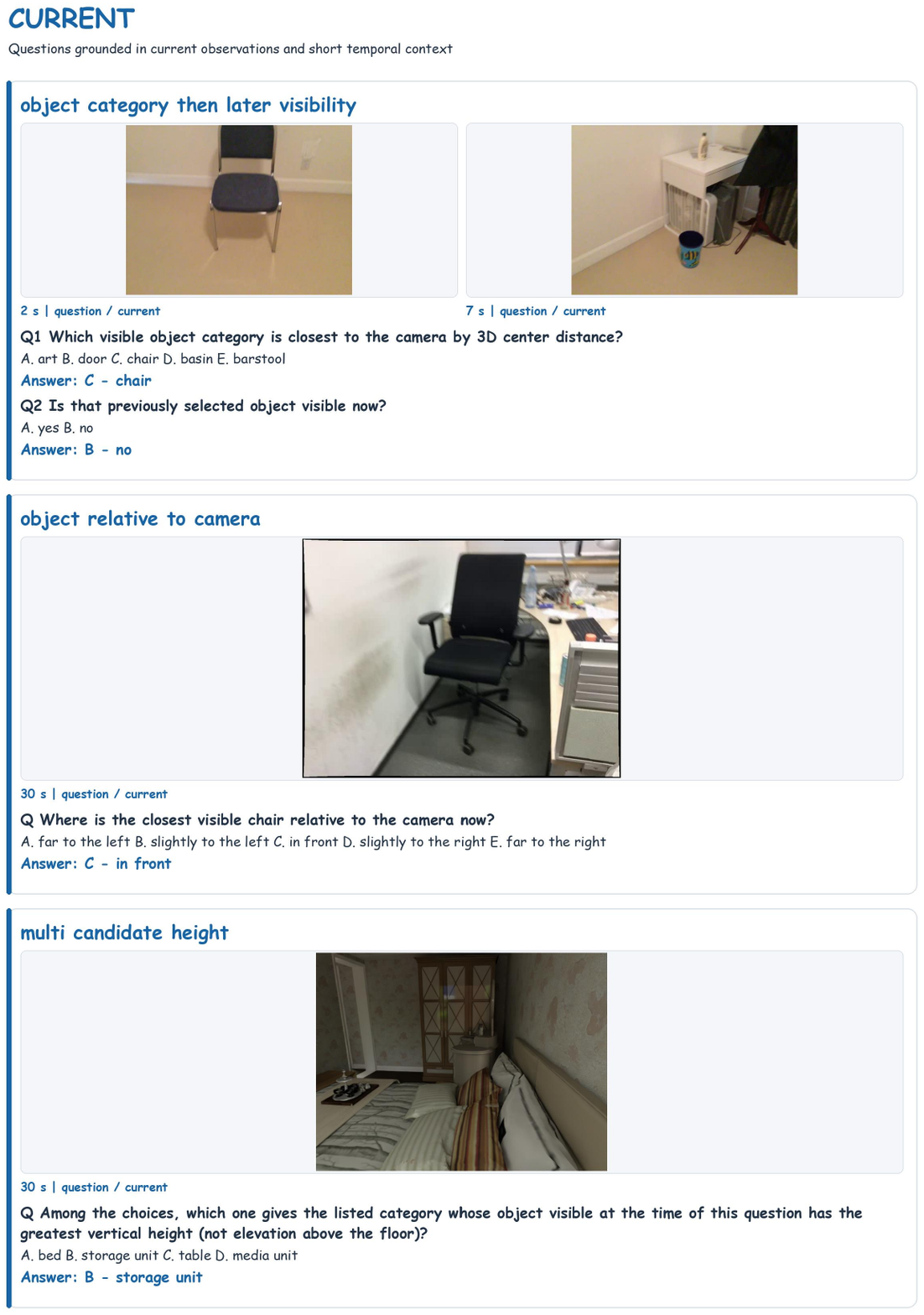}
    \caption{Current QA sample 5}
    \label{fig:current_sample_5}
\end{figure}

\begin{figure}
    \centering
    \includegraphics[width=1.0\linewidth,height=0.85\textheight,keepaspectratio]{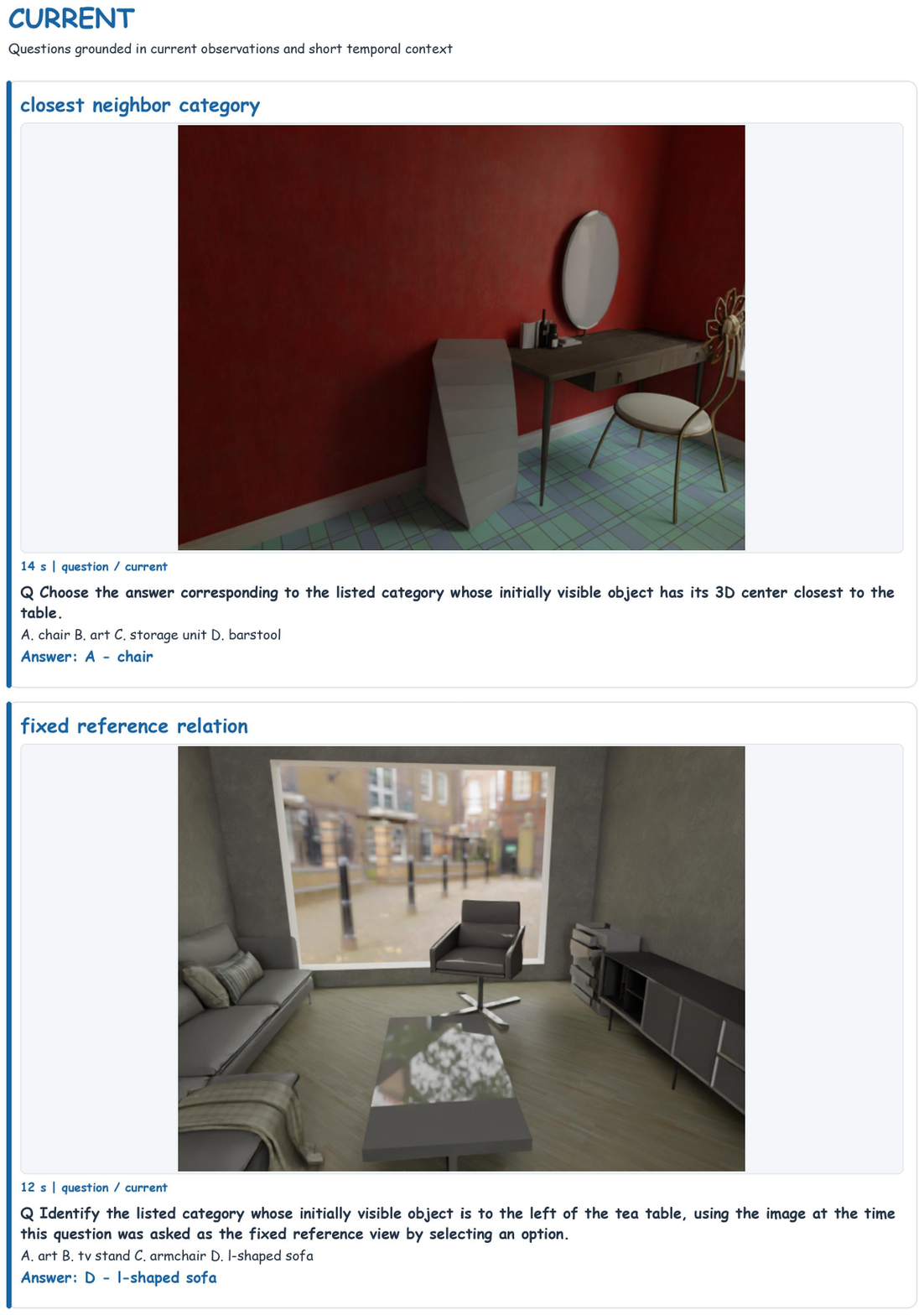}
    \caption{Current QA sample 6}
    \label{fig:current_sample_6}
\end{figure}

\paragraph{Current observations and recent changes.}
Figure~\ref{fig:current_sample_1} illustrates approaching objects, newly visible categories, and camera displacement using observations five seconds apart.
Figure~\ref{fig:current_sample_2} distinguishes trajectory length, camera rotation, and camera-to-object distance, while Figure~\ref{fig:current_sample_3} combines metric distance, changes in distance, and a linked category--distance query.
Figure~\ref{fig:current_sample_4} shows closest-object identification, a follow-up relation query, and disappearance from view.
Figure~\ref{fig:current_sample_5} illustrates tracking a selected referent across turns, camera-relative direction, and height comparison; Figure~\ref{fig:current_sample_6} shows nearest-neighbor and fixed-reference spatial relations.
Together, these examples distinguish geometric reasoning in the current view from short-window temporal comparisons.

\begin{figure}
    \centering
    \includegraphics[width=1.0\linewidth,height=0.85\textheight,keepaspectratio]{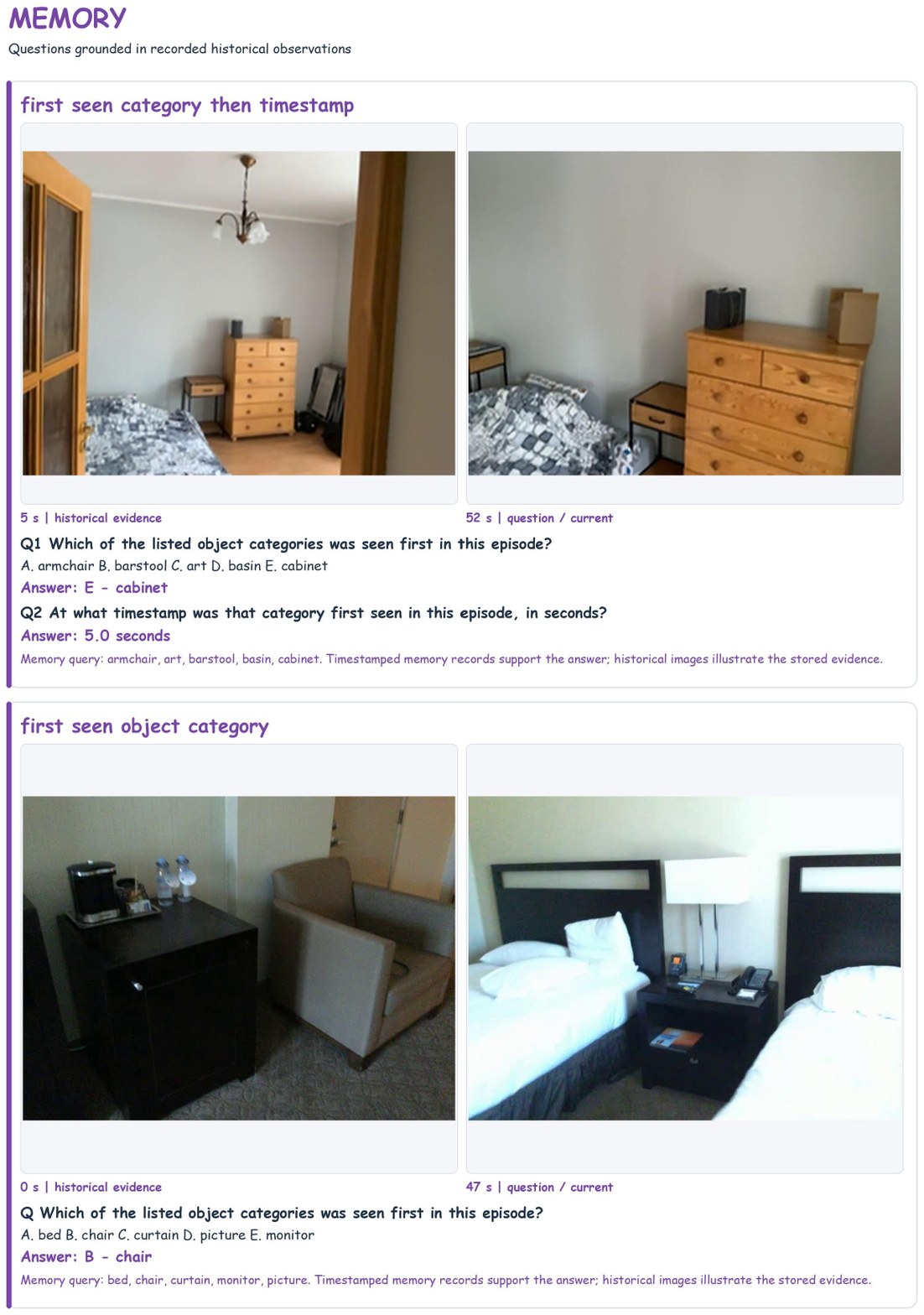}
    \caption{Memory QA sample 1}
    \label{fig:memory_sample_1}
\end{figure}

\begin{figure}
    \centering
    \includegraphics[width=1.0\linewidth,height=0.85\textheight,keepaspectratio]{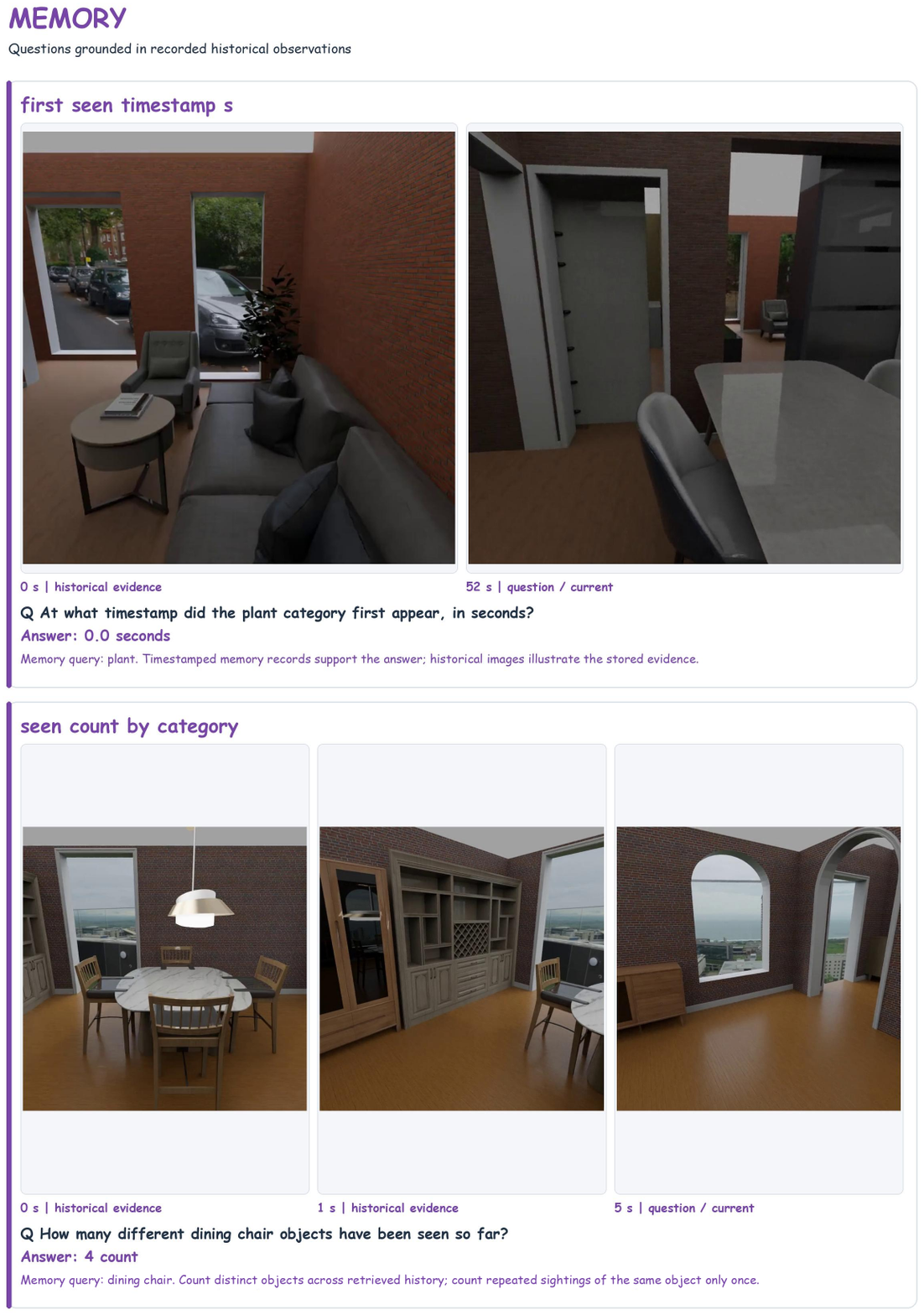}
    \caption{Memory QA sample 2}
    \label{fig:memory_sample_2}
\end{figure}

\begin{figure}
    \centering
    \includegraphics[width=1.0\linewidth,height=0.85\textheight,keepaspectratio]{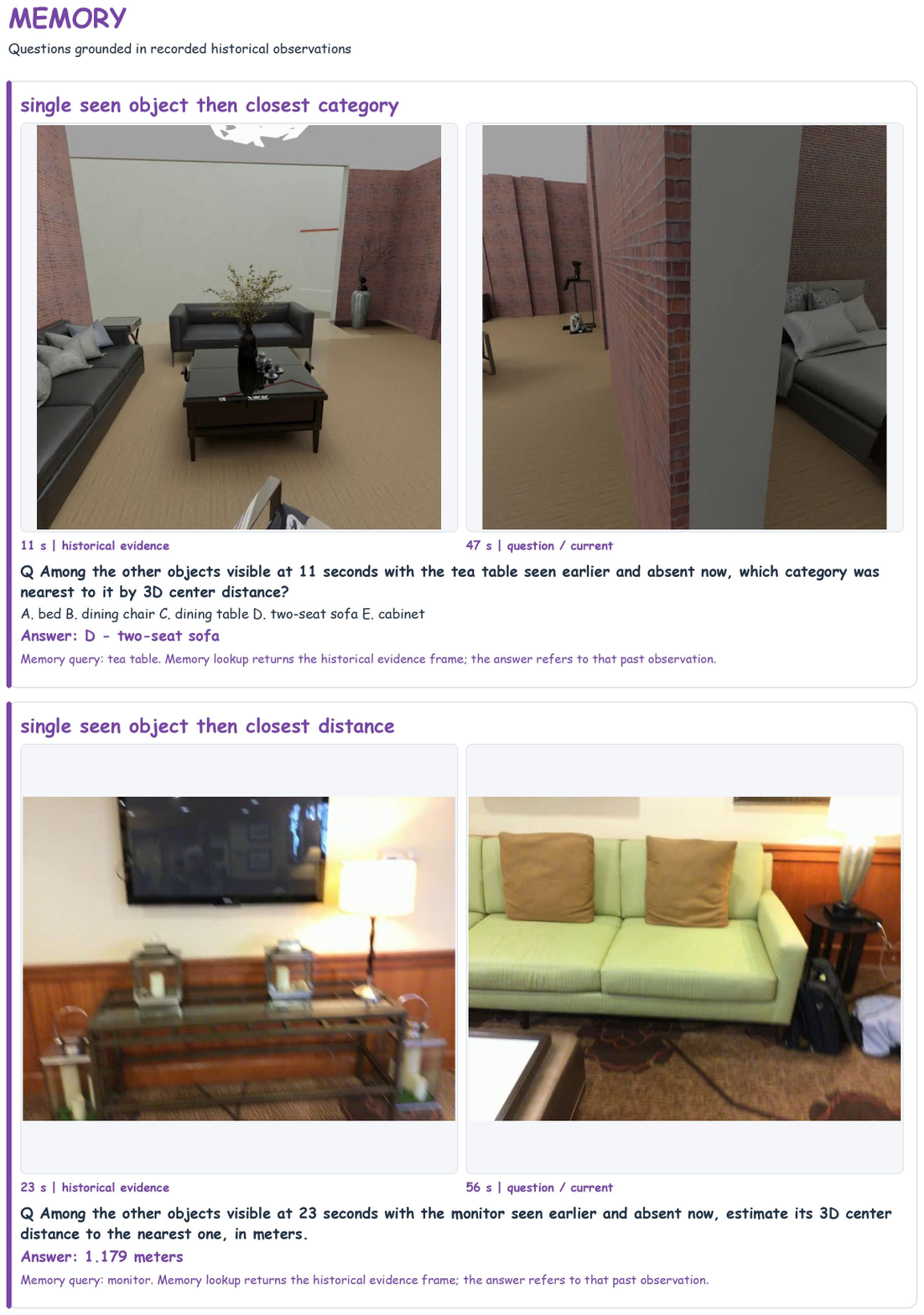}
    \caption{Memory QA sample 3}
    \label{fig:memory_sample_3}
\end{figure}

\begin{figure}
    \centering
    \includegraphics[width=1.0\linewidth,height=0.85\textheight,keepaspectratio]{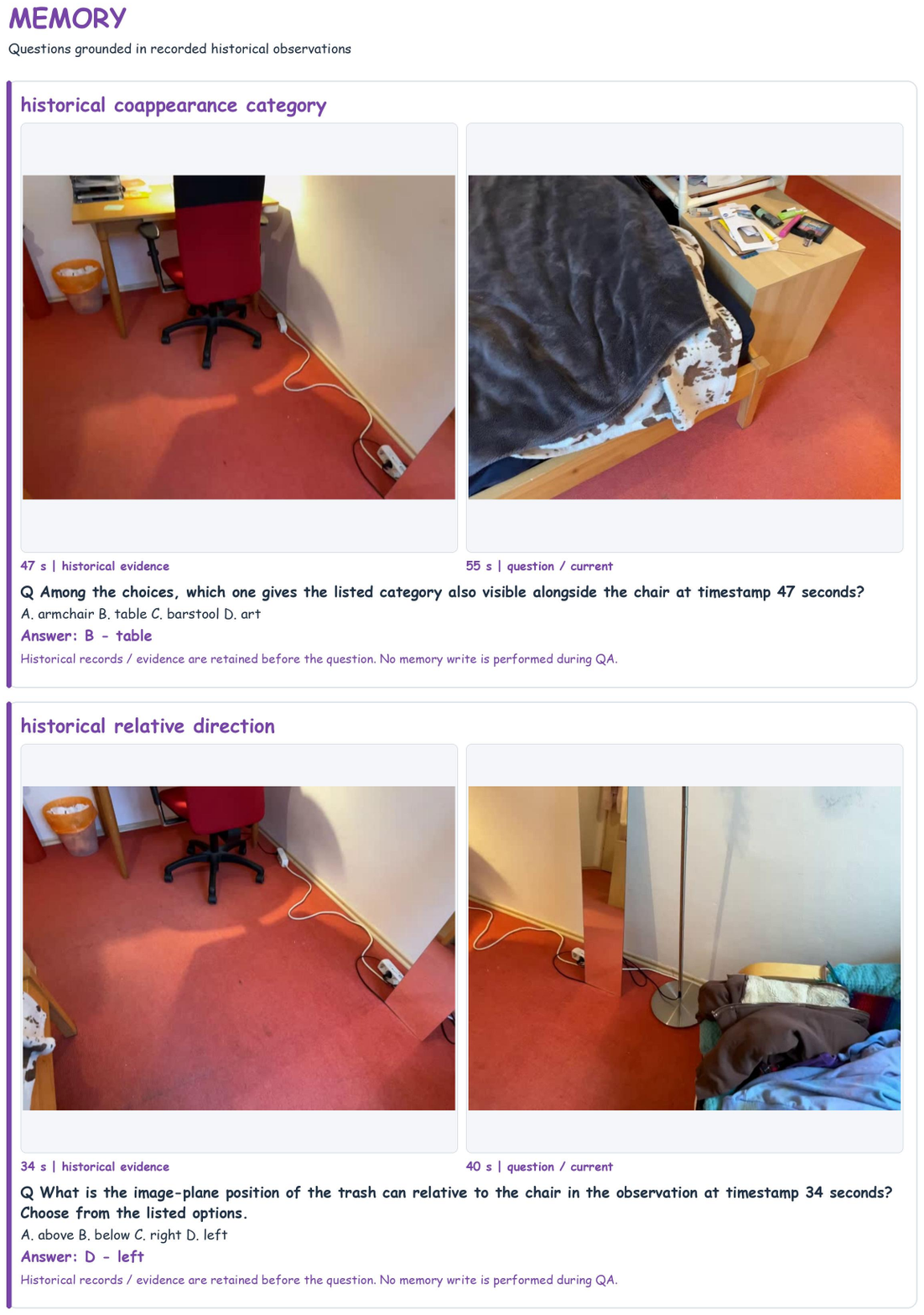}
    \caption{Memory QA sample 4}
    \label{fig:memory_sample_4}
\end{figure}

\begin{figure}
    \centering
    \includegraphics[width=1.0\linewidth,height=0.85\textheight,keepaspectratio]{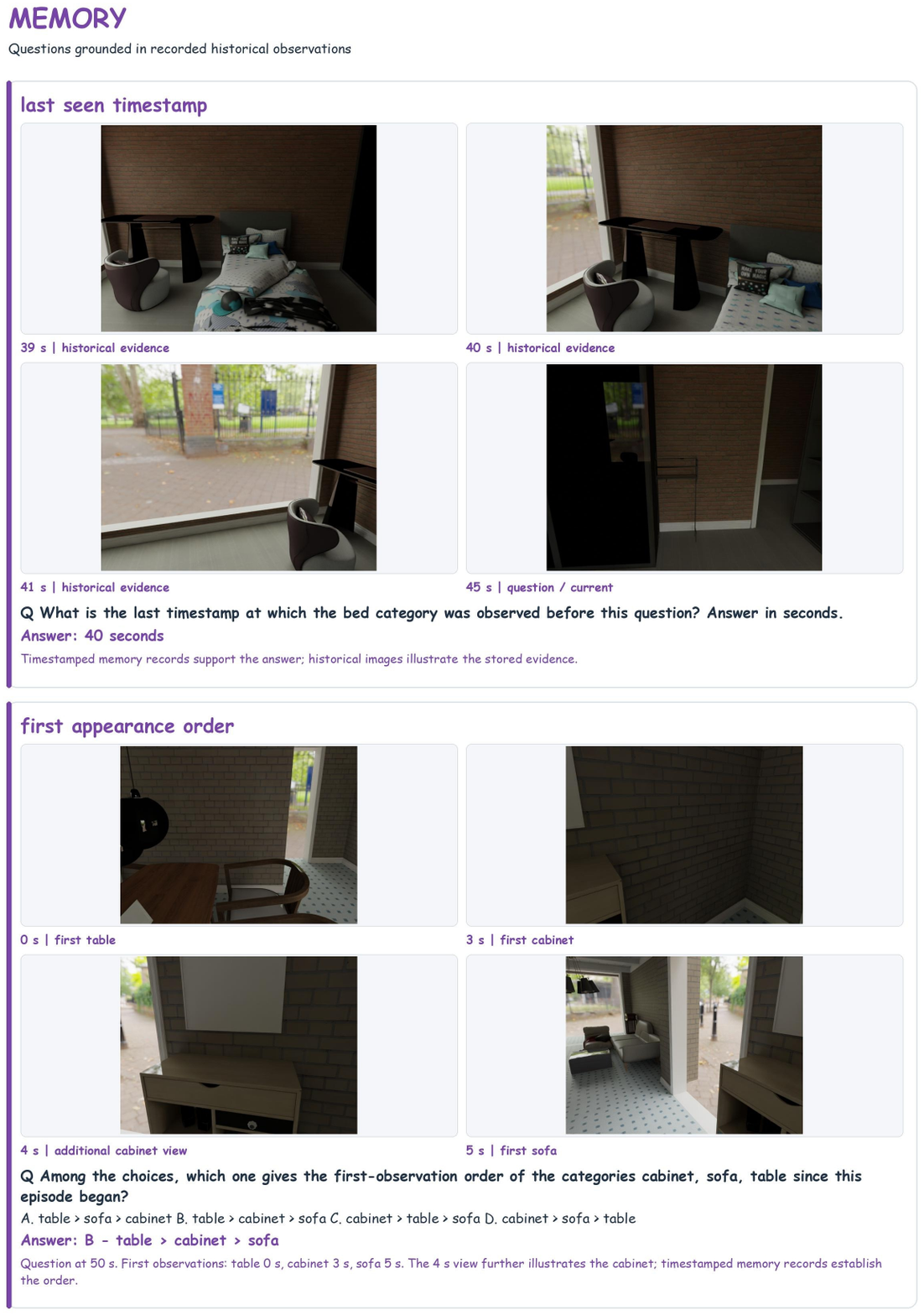}
    \caption{Memory QA sample 5}
    \label{fig:memory_sample_5}
\end{figure}

\begin{figure}
    \centering
    \includegraphics[width=1.0\linewidth,height=0.85\textheight,keepaspectratio]{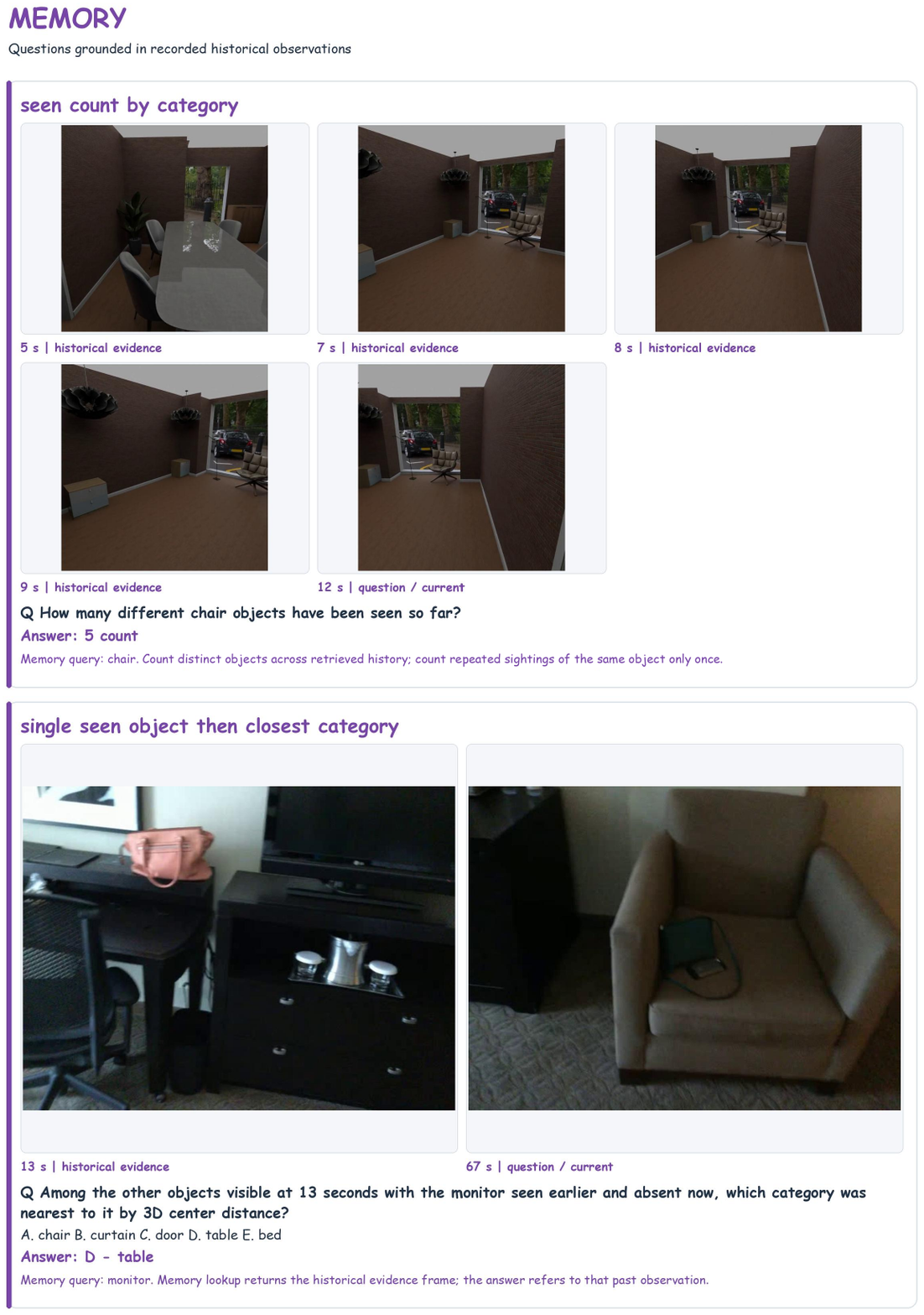}
    \caption{Memory QA sample 6}
    \label{fig:memory_sample_6}
\end{figure}

\paragraph{Historical evidence.}
Figure~\ref{fig:memory_sample_1} illustrates retrieving the first observed category and its timestamp, and Figure~\ref{fig:memory_sample_2} contrasts first-appearance lookup with counting distinct previously observed instances.
Figure~\ref{fig:memory_sample_3} uses an earlier observation to recover the nearest category or distance for a referent no longer visible.
Figure~\ref{fig:memory_sample_4} shows historical coappearance and image-plane direction; Figure~\ref{fig:memory_sample_5} uses timestamps to recover the last observation and first-appearance order.
Figure~\ref{fig:memory_sample_6} provides further distinct-instance counting and historical nearest-neighbor examples, illustrating why the current view alone may be insufficient.

\begin{figure}
    \centering
    \includegraphics[width=1.0\linewidth,height=0.85\textheight,keepaspectratio]{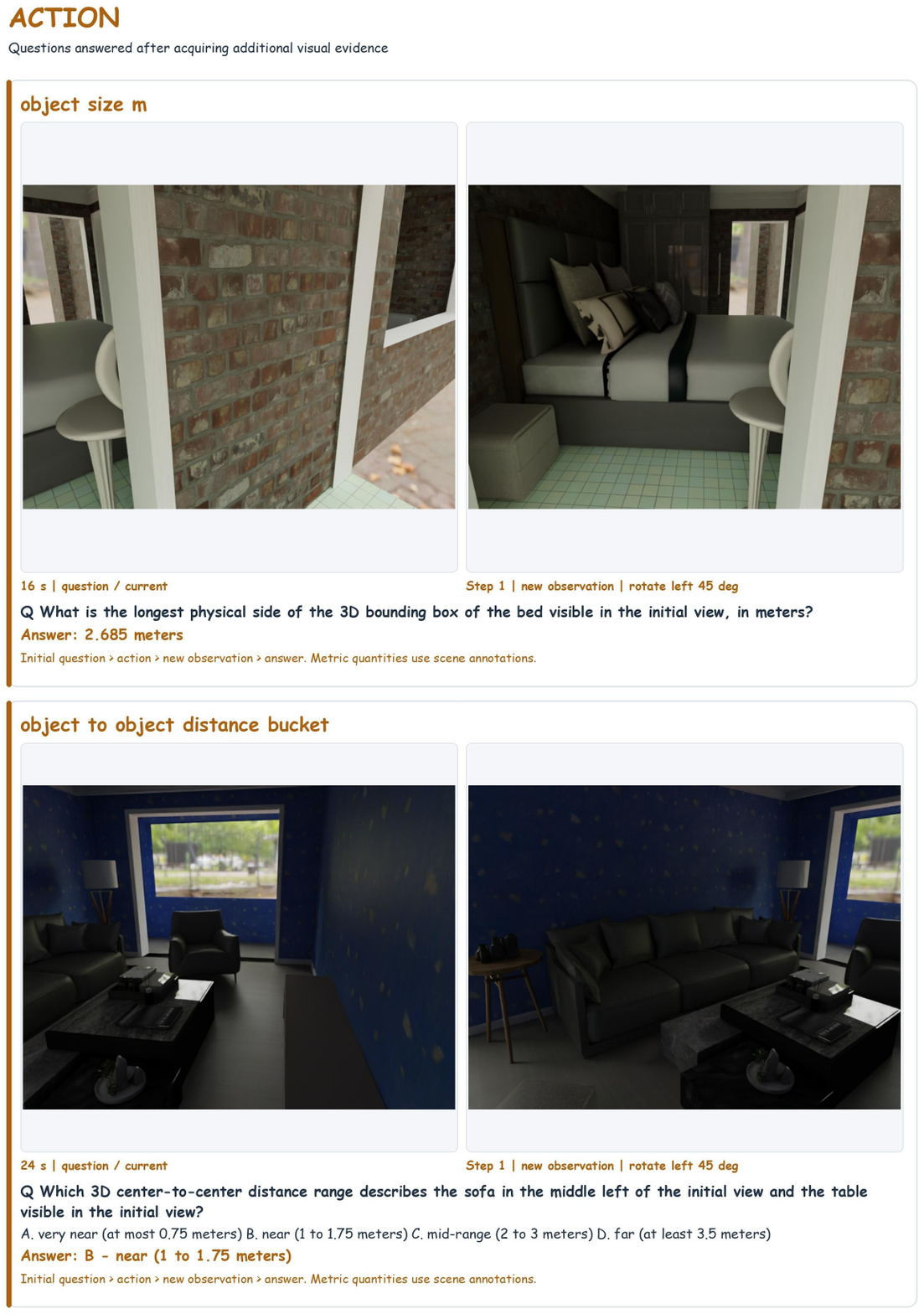}
    \caption{Action QA sample 1}
    \label{fig:action_sample_1}
\end{figure}

\begin{figure}
    \centering
    \includegraphics[width=1.0\linewidth,height=0.85\textheight,keepaspectratio]{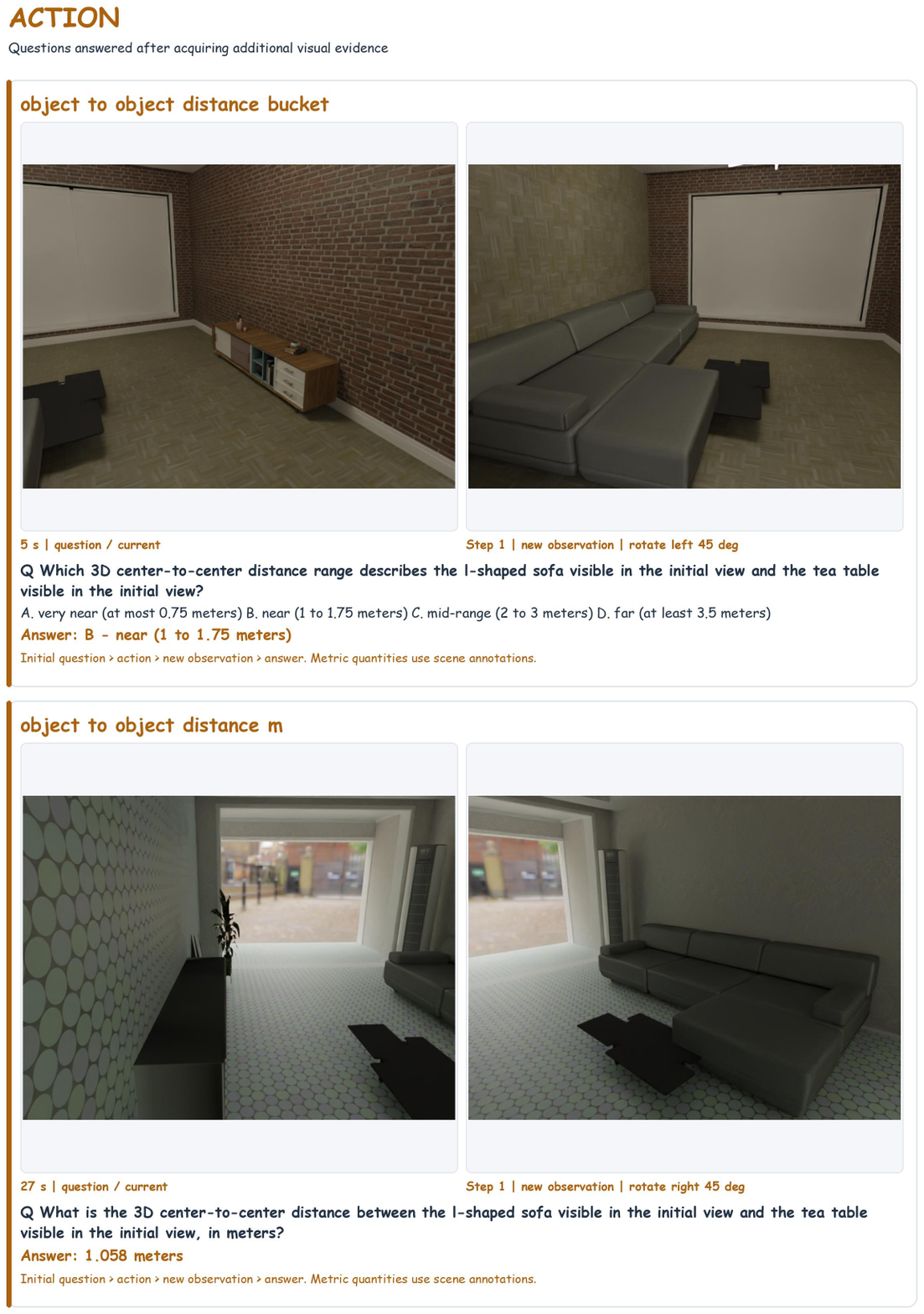}
    \caption{Action QA sample 2}
    \label{fig:action_sample_2}
\end{figure}

\begin{figure}
    \centering
    \includegraphics[width=1.0\linewidth,height=0.85\textheight,keepaspectratio]{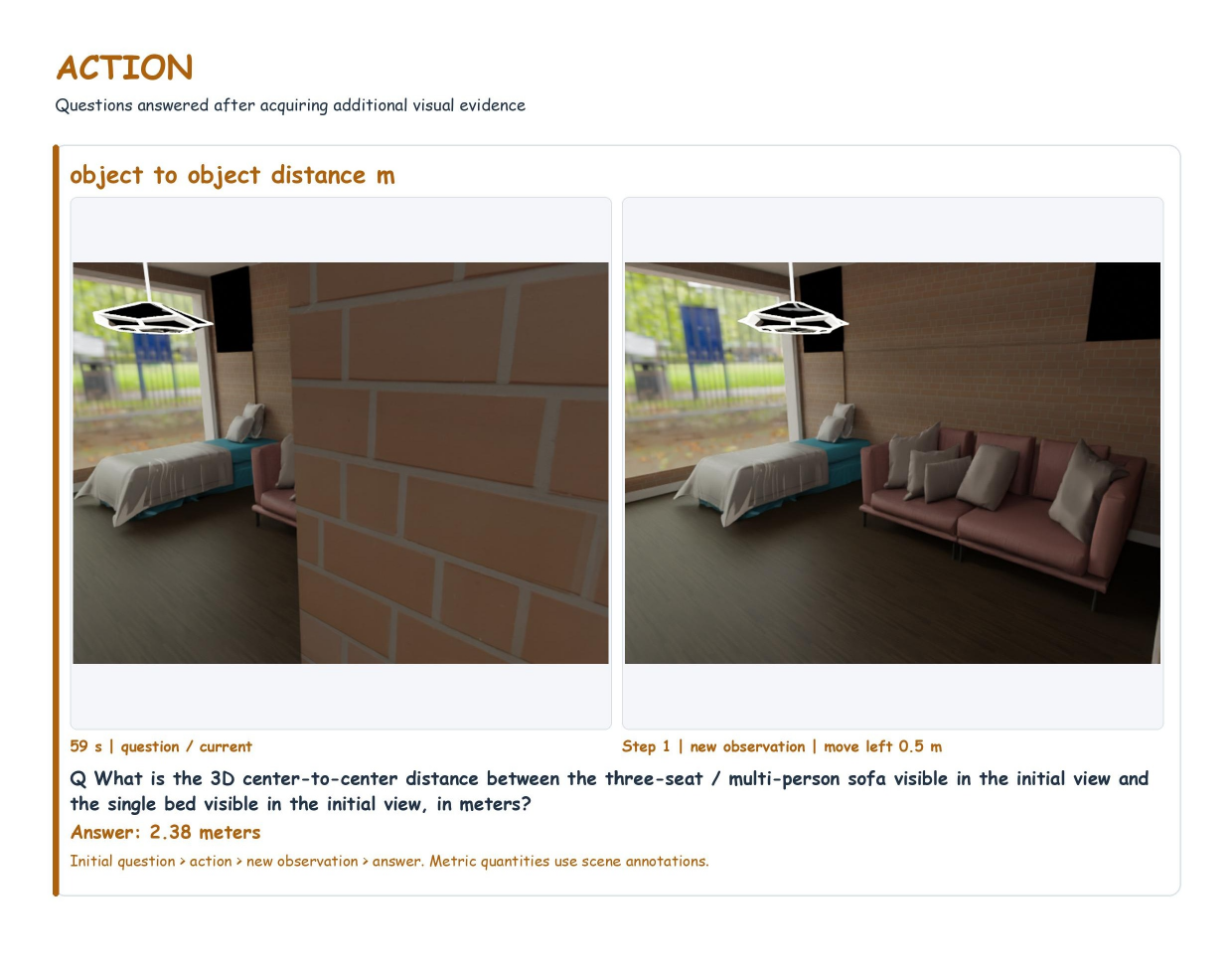}
    \caption{Action QA sample 3}
    \label{fig:action_sample_3}
\end{figure}

\begin{figure}
    \centering
    \includegraphics[width=1.0\linewidth,height=0.85\textheight,keepaspectratio]{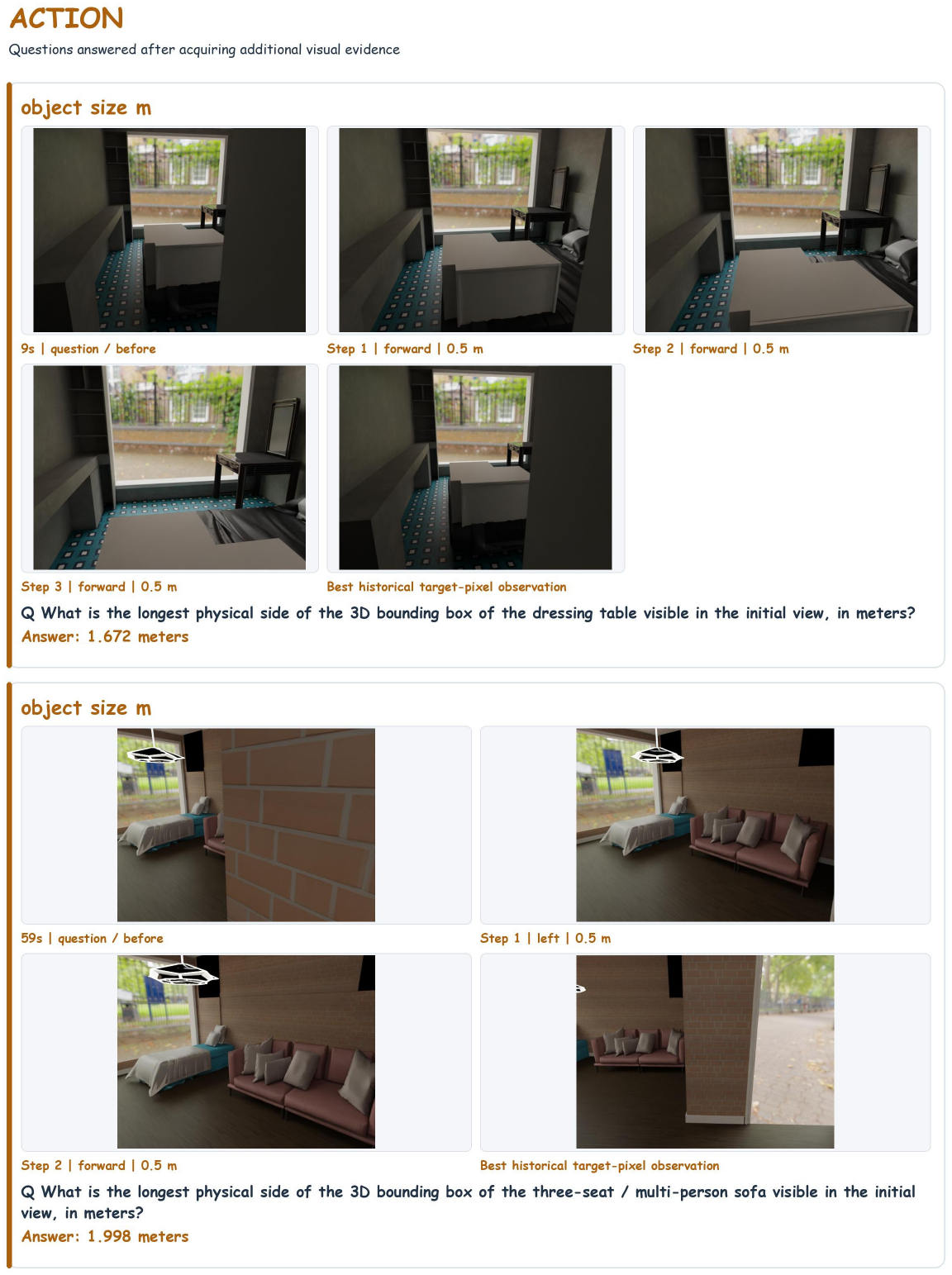}
    \caption{Qualitative examples of local view acquisition for object-size estimation (MemAct-9B rollouts). Each example shows the observation at question time, the view returned after each camera command, and the pre-question view with the largest target-pixel share. The camera commands provide additional views of the queried object to support size estimation.}
    \label{fig:action_sample_4}
\end{figure}

\paragraph{Acquiring additional views.}
Figure~\ref{fig:action_sample_1} shows turns that expose a target for size or distance-range questions, and Figure~\ref{fig:action_sample_2} illustrates categorical and numerical inter-object distance queries after a turn.
Figure~\ref{fig:action_sample_3} shows a lateral movement that provides an additional view for an inter-object distance question.
Figure~\ref{fig:action_sample_4} shows MemAct-9B rollouts in which multi-step camera commands provide additional views of the queried object before size estimation, together with the pre-question view with the largest target-pixel share.
These examples visualize the evidence acquired by actions.

\begin{figure}
    \centering
    \includegraphics[width=1.0\linewidth,height=0.85\textheight,keepaspectratio]{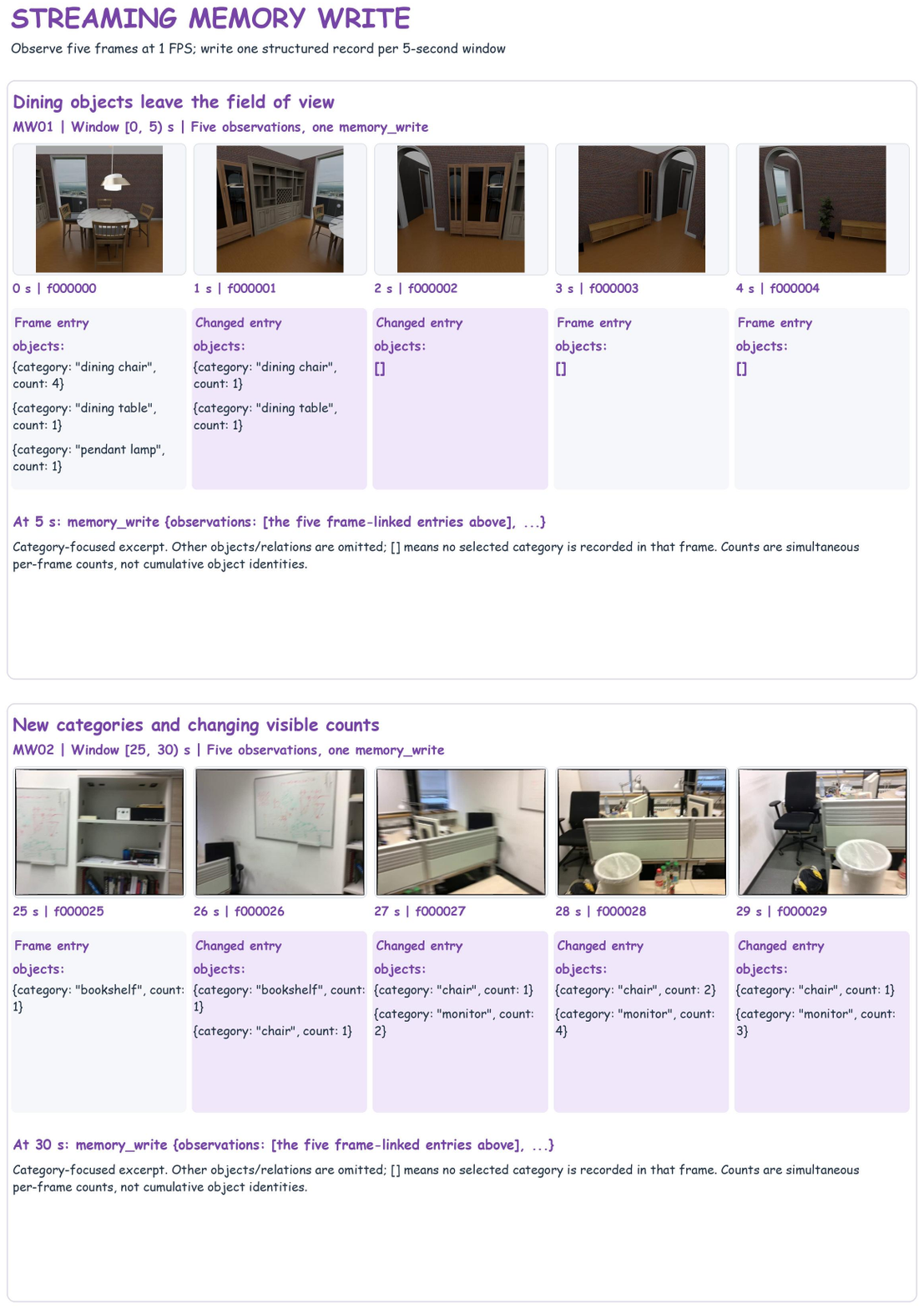}
    \caption{Memory write sample 1. Each five-second window is written by one \texttt{memory\_write}, which contains one record per frame.}
    \label{fig:memory_write_sample_1}
\end{figure}

\begin{figure}
    \centering
    \includegraphics[width=1.0\linewidth,height=0.85\textheight,keepaspectratio]{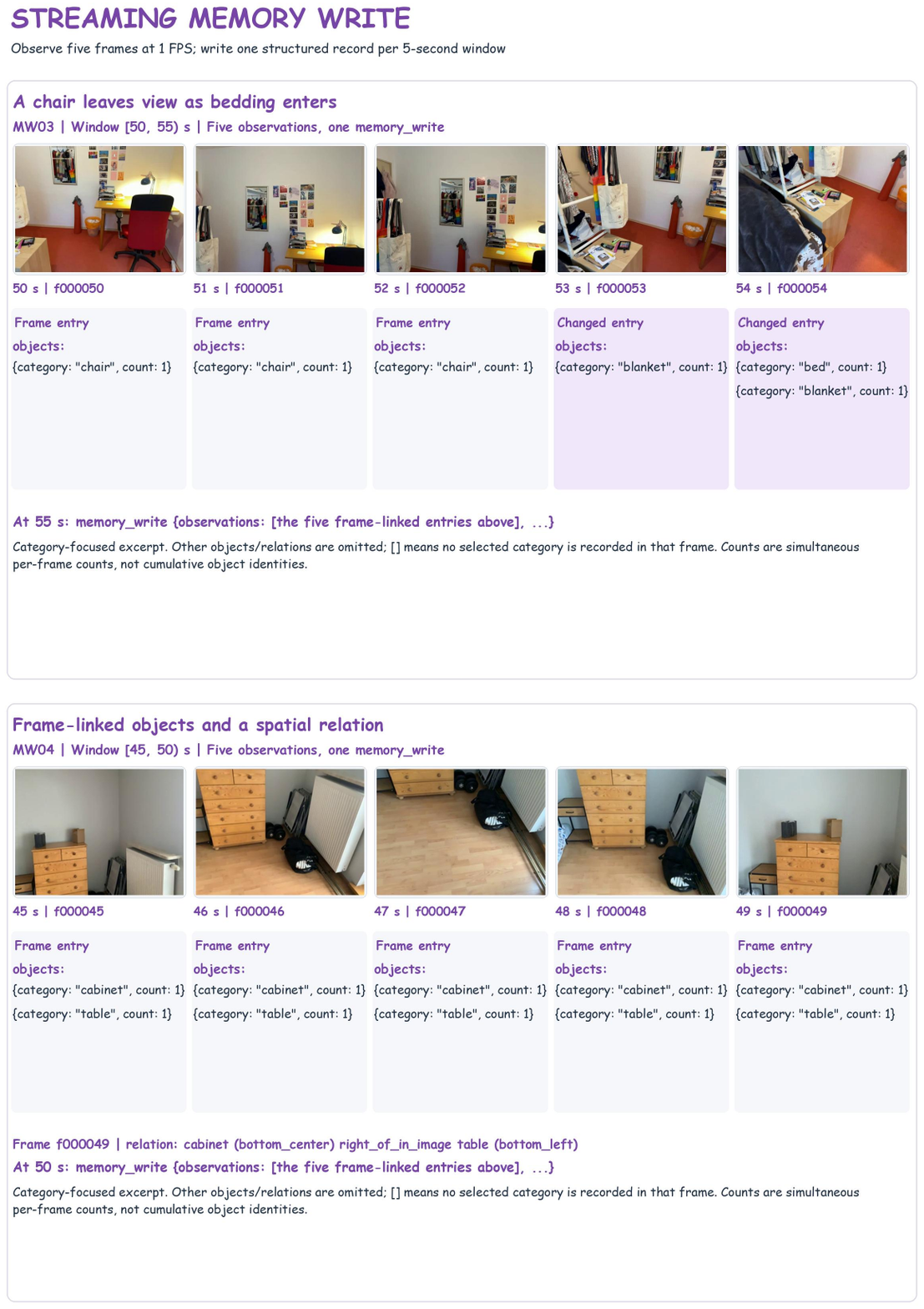}
    \caption{Memory write sample 2. Each five-second window is written by one \texttt{memory\_write}, which contains one record per frame.}
    \label{fig:memory_write_sample_2}
\end{figure}

\paragraph{Incremental memory writes.}
Figure~\ref{fig:memory_write_sample_1} shows five-second write windows in which objects leave view or new categories and counts appear.
Figure~\ref{fig:memory_write_sample_2} illustrates changing categories and a frame-linked spatial relation.
Both figures retain individual frame entries within each write window: per-frame visible counts are not cumulative distinct-instance counts.

\end{document}